\documentclass[pdflatex,sn-mathphys-num]{sn-jnl}

\usepackage{graphicx}%
\usepackage{multirow}%
\usepackage{amsmath,amssymb,amsfonts}%
\usepackage{amsthm}%
\usepackage{mathrsfs}%
\usepackage[title]{appendix}%
\usepackage{xcolor}%
\usepackage{textcomp}%
\usepackage{manyfoot}%
\usepackage{booktabs}%
\usepackage{algorithm}%
\usepackage{algorithmicx}%
\usepackage{algpseudocode}%
\usepackage{listings}%
\usepackage{subcaption}
\usepackage{hhline}
\usepackage{subcaption}
\usepackage{array}

\theoremstyle{thmstyleone}%
\theoremstyle{thmstyletwo}%

\theoremstyle{thmstylethree}%

\begin{document}

\title{ENIGMA-360: An Ego-Exo Dataset for Human Behavior Understanding in Industrial Scenarios}

\author*[1,2]{\fnm{Francesco} \sur{Ragusa}}\email{francesco.ragusa@unict.it}
\author[1,2]{\fnm{Rosario} \sur{Leonardi}}
\author[2]{\fnm{Michele} \sur{Mazzamuto}}
\author[2]{\fnm{Daniele} \sur{Di Mauro}}
\author[1]{\fnm{Camillo} \sur{Quattrocchi}}
\author[1]{\fnm{Alessandro} \sur{Passanisi}}
\author[1]{\fnm{Irene} \sur{D'Ambra}} 
\author[1,2]{\fnm{Antonino} \sur{Furnari}}
\author[1,2]{\fnm{Giovanni Maria} \sur{Farinella}}

\affil*[1]{\orgdiv{LIVE@IPLAB, Department of Mathematics and Computer Science}, \orgname{University of Catania}, \orgaddress{\city{Catania}, \country{Italy}}}
\affil[2]{\orgname{Next Vision s.r.l. - Spinoff of the University of Catania}, \orgaddress{\city{Catania}, \country{Italy}}}


\abstract{Understanding human behavior from  complementary egocentric (ego) and exocentric (exo) points of view enables the development of systems that can support workers in industrial environments and enhance their safety. However, progress in this area is hindered by the lack of datasets capturing both views in realistic industrial scenarios. To address this gap, we propose ENIGMA-360, a new ego-exo dataset acquired in a real industrial scenario. The dataset is composed of 180 egocentric and 180 exocentric procedural videos temporally synchronized offering complementary information of the same scene. The 360 videos have been labeled with temporal and spatial annotations, enabling the study of different aspects of human behavior in industrial domain. We provide baseline experiments for 4 foundational tasks for human behavior understanding: 1) Temporal Action Segmentation, 2) Keystep Recognition, 3) Egocentric Human-Object Interaction Detection, and 4) Video-Grounded Cross-View VQA, showing the limits of state-of-the-art approaches on this challenging scenario. These results highlight the need for new models capable of robust ego-exo understanding in real‑world environments. We publicly release the dataset and its annotations at \url{https://fpv-iplab.github.io/ENIGMA-360/}.}

\keywords{Ego-exo dataset, Industrial domain, Egocentric and Exocentric views, Human Behavior Understanding}



\maketitle

\section{Introduction}
\label{intro}

\begin{figure}[ht]
    \centering
    \includegraphics[width=\linewidth]{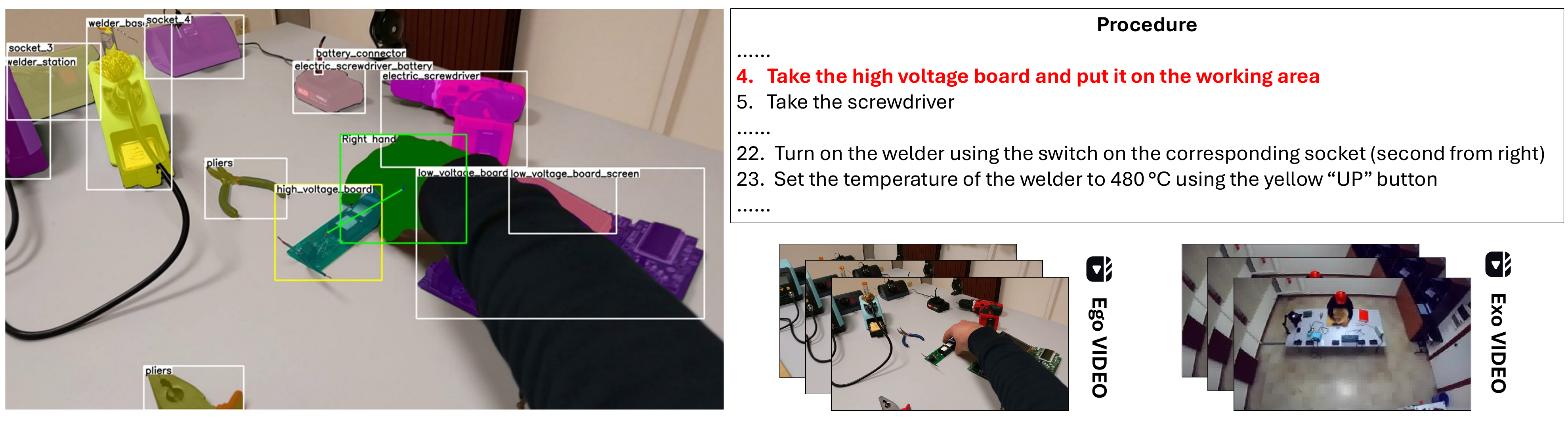}
    \caption{Overview of the ENIGMA-360 dataset. On the left, we show the spatial annotations available. On the top-right, we report the designed procedure for acquisition purpose. On the bottom-right we show synchronized egocentric and exocentric video frames illustrating the multi-view setup.}
    \label{fig:labels}
\end{figure}

Understanding the human behavior and its interactions with the physical world is a fundamental aspect for building AI agents capable to support humans in daily life. Procedures often involve complex, multi-step activities that require diverse skills and the use of different objects.
For instance, in industrial environments, tasks such as machinery maintenance demand a sequence of coordinated actions. Following standard protocols, a worker may begin by wearing personal protective equipment (i.e., PPEs) such as helmets or gloves, proceed to interact with tools (e.g., take the iron solder, connect the battery to the electric screwdriver), perform functional tests (e.g., press buttons on the electric panel), and conclude by reporting the intervention (e.g., take the pen, write the report). These scenarios highlight the need for intelligent systems that can temporally segment complex procedures, identify the occurrence and ordering of individual keysteps, and detect detailed human–object interactions, enabling real-time verification of task execution quality and safety compliance.

To effectively assist humans in complex real-world environments, an ideal intelligent system should be able to understand human behavior from multimodal observations. These observations can be captured by wearable smart cameras (e.g., smart glasses)~\cite{Plizzari2024AnOutlook,Colombo3316782.3322754,mazzamuto2023wearable}, which provide detailed insights into human-object interactions and performed actions, as well as by fixed cameras that capture the broader environmental context~\cite{Grauman_2024_CVPR}. These complementary viewpoints (ego-exo) offer a holistic understanding of the same scene, enabling intelligent assistants to provide comprehensive, 360-degree support (see Figure \ref{fig:labels}-bottom-right).
For example, during the execution of a maintenance procedure, the intelligent system should continuously monitor the sequence of worker actions, understand the order in which steps are performed, and track the progression of the procedure in real time. Additionally, the system should verify whether each step has been completed correctly, provide guidance on how to perform it properly, and alert the worker to potential safety risks (e.g., \textit{Before touching the soldering iron, wear protective gloves!}). However, despite this potential, existing datasets rarely provide synchronized ego-exo recordings in real industrial environments.

In general, the design of novel architectures able to understand human behavior has been driven by the availability of public datasets spanning diverse domains~\cite{Grauman_2024_CVPR, Shan2020UnderstandingHH, EgoProceLECCV2022, Liu_2022_CVPR}. These include everyday environments such as kitchens~\cite{damen2018epickitchen, lu2021egocentric, zhukov2019crosstask}, daily activities~\cite{Pirsiavash2012DetectingAO, Lee_UTE_2012}, and simulated industrial settings~\cite{sener2022assembly101, ragusa2023meccano, schoonbeek2024industreal, ikea}. However, existing industrial-like datasets typically rely on proxy scenarios such as toy assemblies or simplified environments with textureless components \cite{ragusa2023meccano, sener2022assembly101, schoonbeek2024industreal}, due to the inherent challenges of capturing data in real industrial contexts, such as privacy regulations, safety constraints, and the need to protect proprietary processes. As a result, they fail to capture the visual complexity, tool variability, and procedure realism required for developing deployable industrial-assistance models.

To address these limitations, we present ENIGMA-360, a new multi-view dataset composed of 180 egocentric and 180 exocentric videos acquired in a real industrial laboratory. The video pairs are temporally synchronized, with each pair capturing the same scene from two distinct perspectives. This dual-view setup enables the extraction of complementary information: the egocentric view offers fine-grained details of hand-object interactions, while the exocentric view provides broader contextual understanding of the environment and task dynamics. To collect the dataset, we set up an industrial laboratory equipped with real industrial tools and components, such as a power supply unit, soldering iron, and electrical boards. In collaboration with industrial experts, we designed two realistic maintenance procedures to ensure the dataset reflects realistic industrial workflows. To eliminate the need for physical manuals~\cite{ragusa2023meccano, sener2022assembly101, schoonbeek2024industreal}, we developed a wearable application that provides step-by-step instructions to participants during data collection. 

Unlike existing large-scale egocentric datasets, such as Ego4D\cite{Grauman2022Ego4DAT} or Ego-Exo4D\cite{Grauman_2024_CVPR}, which primarily capture unconstrained daily activities or opportunistic recordings, ENIGMA-360 is explicitly designed around highly structured, instruction-driven procedural workflows defined in collaboration with domain experts. Rather than merely expanding scale, our dataset serves as a high-fidelity, multi-task test-bed where strict step-level dependencies, professional objects, and multi-view synchronization present unique challenges.

ENIGMA-360 is richly annotated to support a wide range of research tasks from both egocentric and exocentric perspectives. Specifically, it includes temporal annotations marking the start and end times of keysteps, as well as spatial annotations identifying hand-object interactions within keyframes. In addition to the manual annotations, we provide a comprehensive set of additional resources to support extended research tasks. Specifically, we release segmentation masks for hands and manipulated objects~\cite{sam_hq}, and provide DINOV2 video features~\cite{oquab2023dinov2}. 

Crucially, ENIGMA-360 establishes a professional-grade digital twin context by providing synchronized 3D scans of the laboratory and all industrial objects present in the dataset. To encourage research on synthetic-to-real transfer and scalable model training, we make these 3D models fully available. This combination of synchronized multi-view videos, detailed expert annotations, and high-fidelity 3D assets is completely unique among industrial datasets and absent in daily-activity benchmarks.

To highlight the usefulness of the proposed multi-view dataset, we performed baseline experiments related to 4 tasks focused on human behavior understanding: 1) Temporal Action Segmentation, 2) Keystep Recognition, 3) Egocentric Human-Object Interaction Detection, and 4) Video-Grounded Cross-View VQA. The performance gap observed across methods highlights the substantial domain gap between industrial and everyday activities and confirms that ENIGMA-360 poses new challenges for human behavior understanding.

This work extends our previous research~\cite{ragusa2023enigma51} by incorporating 129 additional egocentric videos and 180 exocentric videos, enriching the dataset with new temporal and spatial annotations, and introducing a benchmark tailored to the analysis of human behavior in industrial environments. We temporally synchronize each ego-exo pair to enable joint modeling across viewpoints, and we provide fine-grained temporal labels of keysteps along with spatial annotations capturing human–object interactions. These include bounding boxes around the user’s hands and manipulated objects, as well as explicit hand–object association links. In addition, we expand the benchmark by evaluating state‑of‑the‑art methods under this more challenging setting. This extended study contributes a new suite of experiments spanning three additional tasks and offers further insights into the complexity and unique characteristics of industrial scenarios.

In sum the contributions of this work are as follows: 1) we present ENIGMA-360, a novel ego-exo dataset captured in a real industrial laboratory, consisting of 180 egocentric and 180 exocentric videos; 2) We provide detailed manual annotations to study different aspects of human behavior in industrial domains; 3) We study 4 different tasks, showing the limitations of current state-of-the-art models in this challenging domain; 4) We release a set of supplementary resources, to push researchers to explore additional tasks on the proposed dataset. The ENIGMA-360 dataset and its additional resources are publicly available at: \url{https://fpv-iplab.github.io/ENIGMA-360/}.

\section{Related Work}\label{rel_work}

Our work is related to previous research lines which are revised in the following sections.

\subsection{Datasets for Human Behavior Understanding}

\subsubsection{Egocentric Datasets}

Early egocentric datasets such as ADL \cite{Pirsiavash2012DetectingAO} and THU-READ \cite{thu-read_Tang_2017} established action recognition benchmarks in daily-life settings, while EGTEA Gaze+ \cite{li2020eye} and the EPIC-KITCHENS family \cite{damen2018epickitchen, Damen2022RESCALING} scaled this line up to large, richly annotated kitchen activities, later enabling fine-grained hand-object interaction and segmentation tasks \cite{gun-71_dataset_Rogez_2015, VISOR2022, Liu_2022_CVPR, fan2023arctic}. VOST \cite{tokmakov2023breaking} extends this fine-grained focus to objects that undergo dramatic appearance changes under manipulation, while EgoProceL \cite{EgoProceLECCV2022} aggregates 16 procedural tasks (including toy-bike, tent, and PC assembly/disassembly, partly sourced from CMU-MMAC, EGTEA Gaze+, MECCANO, and EPIC-Tent) to support procedure-learning research across domains. Crucially, however, the procedures in EgoProceL, and in daily-activity datasets more broadly, are largely unconstrained: participants perform tasks with substantial freedom in ordering, pacing, and execution style, and only a small fraction of EgoProceL (e.g., PC assembly) approaches an industrial domain, without expert-defined workflows or mandatory step ordering. Ego4D \cite{Grauman2022Ego4DAT} and CaptainCook4D \cite{peddi2024captaincook4ddatasetunderstandingerrors} further increased video volume and annotation diversity, including procedural-error detection, while HoloAssist \cite{HoloAssist2023} introduced human-human collaborative supervision. Despite this progress, all of these datasets are confined to a single (egocentric) viewpoint and, with the partial exception of the industrial-like tasks, to daily-life or cooking domains, which limits their relevance to industrial monitoring, where the broader environmental context captured by fixed cameras is as informative as hand-level detail.
Closer to our target domain, MECCANO \cite{ragusa2023meccano} and IndustReal \cite{schoonbeek2024industreal} address industrial-like procedures, but both are captured only from the egocentric viewpoint and rely on simplified, texture-less toy assemblies, a design choice motivated by the practical difficulty of recording in real industrial facilities. 
ENIGMA-360 instead couples this notion of structured, expert-defined procedure with acquisition in an operational industrial laboratory, using realistic tools and components (oscilloscope, welding station, electrical boards under repair) and a synchronized exocentric view, a combination not offered by any existing single-view industrial dataset.

\subsubsection{Ego-Exo Datasets}

A second line of work augments the egocentric perspective with additional viewpoints to jointly capture fine-grained manipulation and scene-level context. The 100 Days of Hands (100DOH) dataset \cite{Shan2020UnderstandingHH} takes a complementary approach, aggregating large-scale, multi-source hand-object interaction footage (100K frames, 131 days of recordings) annotated with contact state and active-object bounding boxes, though without temporal synchronization across dedicated viewpoints. LEMMA \cite{jia2020lemma} and H2O \cite{Kwon_2021_ICCV} provide synchronized multi-view recordings of daily activities and bimanual object manipulation, respectively, but at small scale and without industrial content. IKEA-ASM \cite{ikea} and Assembly101 \cite{sener2022assembly101} move toward procedural, assembly-style tasks with multi-view (up to 12 static cameras) acquisition, yet, like MECCANO \cite{ragusa2023meccano} and IndustReal \cite{schoonbeek2024industreal}, are built around simplified or texture-less furniture/toy parts rather than authentic industrial equipment.
FineBio \cite{yagi2025finebio} adopts a comparable multi-view protocol (6 synchronized views, 14.5 hours) but targets biological laboratory experiments, a domain with fundamentally different tool semantics, ergonomics, and safety implications than industrial maintenance and repair. Ego-Exo4D \cite{Grauman_2024_CVPR} represents the largest and most diverse multi-domain effort to date, spanning sports, cooking, dance, and, mechanical repair. However, the bike-repair subset accounts for only 66 of its 916 sequences (14.94 of 233.82 hours) and is captured on bicycle components under comparatively opportunistic recording conditions, rather than around fully specified workflows co-designed with domain experts. The dataset also lacks the fixed-machinery interactions (electrical panels, oscilloscopes, welding stations) and 3D digital-twin assets that ENIGMA-360 provides. Most closely related to our work, IndEgo \cite{Chavan2025IndEgo} recently introduced a large-scale industrial collection (197 hours, 3,460 sequences) with 2 synchronized views, but targets primarily mistake detection and visual question answering, and does not provide the co-located, multi-task annotation suite (temporal segmentation, keystep recognition, HOI detection and segmentation on the same synchronized streams) that we release with ENIGMA-360.
Beyond raw scale or view count, what most distinguishes ENIGMA-360 is the nature of its procedures. Unlike unconstrained activities where actions are largely independent events, ENIGMA-360 is built around two industrial maintenance procedures co-designed with domain experts, in which actions are steps of a predefined sequence with explicit dependencies. This structured, instruction-driven setting, combined with synchronized ego-exo views, high-fidelity 3D digital twins of the laboratory and objects, and co-located annotations for temporal action segmentation, keystep recognition, and HOI detection on the same video streams, makes ENIGMA-360, to our knowledge, an unmatched test-bed for studying how procedural constraints and multi-view consistency affect human behavior understanding models. Our benchmark results confirm that this setting shows a substantial performance gap even for methods competitive on existing, less-structured benchmarks, indicating that rigorous procedural structure introduces challenges not fully captured by prior datasets.

\begin{table}[t] 
\centering 
\resizebox{\textwidth}{!}{ 
\begin{tabular}{|llcccccccccc|m{5.5cm}|} 
\hline 
\textbf{Dataset} & \textbf{Year} & \textbf{Video?} & \textbf{Multi-View?} & \textbf{Settings} & \textbf{Views} & \textbf{Hours} & \textbf{Avg. duration (m)} & \textbf{Sequences} & \textbf{Subjects} & \textbf{Task} \\ \hline 
ADL\cite{Pirsiavash2012DetectingAO} & 2012 & \checkmark & X & Daily activities & 1 & 10 & & 20 & 20 & Action recognition, object interaction \\ 
EGTEA Gaze+\cite{li2020eye} & 2017 & \checkmark & X & Cooking activities & 1 & 28 & & 86 & 32 & Action recognition, action anticipation, gaze prediction \\ 
THU-READ\cite{thu-read_Tang_2017} & 2019 & \checkmark & X & Daily activities & 1 & 224 & & 1920 & 8 & Action recognition \\ 
GUN-71\cite{gun-71_dataset_Rogez_2015} & 2015 & X & X & Daily activities & 1 & N/A & & N/A & 8 & Grasp recognition, hand pose estimation \\ 
EPIC-KITCHENS\cite{VISOR2022} & 2022 & \checkmark & X & Kitchen activities & 1 & 100 & & 700 & 45 & HOI segmentation \\ 
HOI4D\cite{Liu_2022_CVPR} & 2022 & \checkmark & X & Objects manipulation & 1 & 22 & & 4000 & N/A & 3D hand pose, object pose tracking, action segmentation \\ 
ARCTIC\cite{fan2023arctic} & 2023 & \checkmark & X & Object manipulation & 1 & 2 & & 339 & 10 & 3D hand-object reconstruction, interaction field est. \\ 
VOST\cite{tokmakov2023breaking} & 2023 & \checkmark & X & Daily + Industrial & 1 & 4 & & 713 & N/A & Video object segmentation, object tracking \\ 
EgoProcel\cite{EgoProceLECCV2022} & 2022 & \checkmark & X & Multi-domain & 1 & 62 & & N/A & 130 & Action recognition, procedure learning \\ 
Ego4D\cite{Grauman2022Ego4DAT} & 2022 & \checkmark & X & Multi-domain & 1 & 3670 & & 9650 & 923 & \begin{tabular}{@{}l@{}}Episodic memory, PNR localization, state change detection, \\ obj. state classification, locomotion/hand prediction, \\ interaction anticipation, Audio-Visual diarization\end{tabular} \\ 
CaptainCook4D\cite{peddi2024captaincook4ddatasetunderstandingerrors} & 2024 & \checkmark & X & Daily activities & 1 & 94.5 & & 384 & N/A & Action recognition, procedure learning, mistake detection \\ 
MECCANO\cite{ragusa2023meccano} & 2023 & \checkmark & X & Industrial-like & 1 & 7 & & 20 & 20 & Action recognition, anticipation, HOI detection, active object det. \\ 
HoloAssist\cite{HoloAssist2023} & 2023 & \checkmark & X & Assistive & 1 & 166 & & 2221 & 350 & Mistake detection, intervention prediction, hand forecasting \\ 
IndustReal\cite{schoonbeek2024industreal} & 2024 & \checkmark & X & Assembly & 1 & 5.8 & & 84 & 27 & Proc. step recognition, action recognition, assembly state det. \\ \hline 
IKEA ASM\cite{ikea} & 2020 & \checkmark & \checkmark & Assembly & 3 & 35.27 & 1.89 & 371 & 48 & Action recognition, instance segmentation, object tracking, pose est. \\ 
100 Days Hands\cite{Shan2020UnderstandingHH} & 2020 & X & \checkmark & Daily activities & N/A & 3144 & N/A & 27000 & 1350+ & Hand-object interaction det., hand detection, hand reconstruction \\ 
LEMMA\cite{jia2020lemma} & 2020 & \checkmark & \checkmark & Daily activities & 4 & 10.1 & 2.0 & 324 & N/A & Action recognition, human-object interaction detection \\ 
H2O\cite{Kwon_2021_ICCV} & 2021 & \checkmark & \checkmark & Objects manipulation & 5 & N/A & N/A & N/A & N/A & 3D hand-object pose estimation, 3D hand pose estimation \\ 
FineBio\cite{yagi2025finebio} & 2025 & \checkmark & \checkmark & Biological exp. & 6 & 14.5 & 3.9 & 107 & 18 & Temporal action/step segmentation, object detection, HOI detection. \\ 
Assembly101\cite{sener2022assembly101} & 2022 & \checkmark & \checkmark & Assembly & 12 & 42.75 & 7.1 & 362 & 53 & Action recognition, anticipation, temporal action segmentation \\ 
Ego-Exo4D\cite{Grauman_2024_CVPR} & 2024 & \checkmark & \checkmark & Multi-domain & 5-6 & 233.82 & 15.30 & 916 & 740 & Key-step recognition, proficiency estimation, pose estimation \\ 
Ego-Exo4D-Bike (subset) & 2024 & \checkmark & \checkmark & Industrial & 5-6 & 14.94 & 13.54 & 66 & 32 & Key-step recognition, proficiency estimation, pose estimation \\ 
IndEgo\cite{Chavan2025IndEgo} & 2025 & \checkmark & \checkmark & Industrial & 2 & 197 & 3.42 & 3460 & 20 & Mistake detection, VQA, action recognition \\ 
\textbf{ENIGMA-360} & 2025 & \checkmark & \checkmark & Industrial & 2 & 55.77 & 18.59 & 180 & 33 & \begin{tabular}{@{}l@{}}Temporal action segmentation, Action/key-step recognition, HOI segmentation, \\ ST anticipation, NL understanding\end{tabular} \\ \hline 
\end{tabular} 
} 
\caption{Overview of egocentric and multi-view datasets that allow the study of human behavior.} 
\label{tab:ego-dataset} 
\end{table}

\subsection{Temporal Action Segmentation}
Temporal Action Segmentation (TAS) requires jointly classifying and temporally localizing actions in untrimmed video. Foundational approaches rely on temporal convolutional encoder-decoders, from single-stage models \cite{lea2017temporal} to multi-stage refinement \cite{li2020ms}, and on recurrent formulations that align discriminative sub-action representations over long sequences \cite{8585084}. Building on these architectures, C2F-TCN \cite{singhania2021coarse} introduces an implicit multi-resolution ensemble to mitigate sequence fragmentation, MS-TCN++ \cite{li2020ms} combines dilated convolutions at multiple temporal scales, and more recent work incorporates auxiliary structure through semi-supervised training \cite{singhania2022iterative} or temporal-logic constraints \cite{xu2022don}. Transformer-based models have since become the dominant paradigm. ASFormer \cite{chinayi_ASformer} injects local-connectivity priors into a hierarchical encoder-decoder, LTContext \cite{ltc2023bahrami} leverages sparse attention to capture full-sequence context efficiently, and FACT \cite{lu2024fact} performs bidirectional frame-action cross-attention, currently representing the state of the art on standard benchmarks; diffusion-based refinement \cite{liu2023diffusion} offers an alternative, iterative formulation. Distinct from evaluating TAS in a single view, \cite{quattrocchi2023synchronization} exploits synchronized ego-exo pairs for unsupervised cross-view domain adaptation, the setting most closely related to ours.
Notably, none of these methods has been evaluated on real industrial procedures with expert-defined step dependencies and instruction-guided execution. Existing TAS benchmarks are drawn from daily-life or cooking activities, or from mixed-domain collections in which industrial-like content is a marginal, opportunistically recorded subset. We benchmark six representative architectures spanning this landscape on ENIGMA-360, both within each view and across views (EGO→EXO, EXO→EGO), to evaluate whether segmentation performance is driven by the structured and procedural nature of our data. Our results reveal a substantial performance degradation across all evaluated baselines when cross-view generalization is required. While general-purpose frameworks (e.g., FACT \cite{lu2024fact}) evaluate temporal action segmentation on mixed, daily-life datasets, they do not assess models under rigid industrial and professional constraints.

\subsection{Keystep Recognition}
Recognizing keysteps in procedural videos~\cite{ashutosh2024video,bansal2022my,dvornik2022flow,dvornik2023stepformer} is essential for procedural planning~\cite{bi2021procedure,chang2020procedure} and understanding task structures~\cite{narasimhan2023learning,zhou2018towards} such as graphs \cite{Grauman_2024_CVPR, seminara2024differentiable}.
Weakly supervised approaches align keysteps to video content via multimodal embedding similarity \cite{miech2020end}, learned joint embeddings that bring corresponding keysteps closer together \cite{bansal2022my}, or Dynamic Time Warping \cite{cao2020few, chang2019d3tw} and its flow-graph-based \cite{dvornik2022flow} or self-supervised \cite{dvornik2023stepformer} extensions; procedure-planning methods further exploit learned task structure for downstream reasoning \cite{bi2021procedure, chang2020procedure}. Because keysteps are fine-grained and exhibit low inter-class variability, relevant architectures overlap substantially with fine-grained action recognition. LFB \cite{wu2019long} models long-term context via a feature bank, FineGym \cite{shao2020finegym} shows that coarse backbones fail to capture the subtle spatio-temporal semantics of fine-grained actions, and this limitation has motivated dedicated temporal modeling in TimeSformer \cite{gberta_2021_ICML} and TSM \cite{lin2019tsm}, as well as query-based formulations such as TQN \cite{zhang2021temporal}.
Unlike prior keystep-recognition work, which largely assumes a single (typically egocentric) viewpoint, ENIGMA-360 provides an identical keystep taxonomy and synchronized annotations for both views. We adopt TimeSformer \cite{gberta_2021_ICML} as baseline and evaluate it independently on the egocentric and exocentric subsets, isolating the specific contribution of the acquisition viewpoint to performance in a structured, instruction-driven industrial procedure.

\subsection{Egocentric Human-Object Interaction detection}
The task of detecting Human-Object Interactions (HOI) has been widely explored in the context of third-person vision~\cite{Gupta2015VisualSR, gkioxari2018detecting, gao2018ican, qi2018learning, Liao2020PPDMPP, zhang2022upt, wu2022mining}. These methods are designed to detect humans and describe their interactions with surrounding objects, typically by predicting structured triplets (e.g., \textit{person, cutting with, knife}) involving human and object bounding boxes along with interaction labels~\cite{gkioxari2018detecting}. Significant advances in HOI detection have been enabled by the availability of large-scale annotated datasets such as HICO-DET~\cite{Chao2015HICOAB} and V-COCO~\cite{Gupta2015VisualSR}, as well as standardized evaluation protocols that enable consistent benchmarking and facilitate methodological progress. However, while third-person HOI detection has seen significant development, egocentric HOI detection remains comparatively underexplored. One of the fundamental differences is that, whereas third-person vision relies on full-body cues to guide interaction detection, egocentric vision depends primarily on the appearance and motion of the hands. Building on this observation, \cite{Shan2020UnderstandingHH} introduced a new task formulation for Hand-Object Interaction understanding, which jointly addresses the localization of hands and manipulated objects through bounding boxes, along with the estimation of hand contact states. This formulation has been adopted and further developed in several works~\cite{ragusa2023meccano, leonardi2024synthdata, Fu2021SequentialDF}, often serving as the foundation for HOI detection pipelines in egocentric settings. 
Recently, this domain has seen a surge in standardization efforts, with new dedicated benchmarks and methods introduced to formalize egocentric HOI evaluation~\cite{DENG2026130216}.

In contrast, other studies~\cite{VISOR2022, zhang2022fine, banerjee2024hot3d, leonardi2024synthetic} have proposed a Hand-Object Segmentation (HOS) approach, framing HOI detection as a pixel-level segmentation task rather than relying on bounding box-based localization. Simultaneously, advanced unified architectures have emerged to integrate interaction analyzing, reasoning, and pixel-level grounding into cohesive frameworks~\cite{su2025annexe}.

In this work, we follow the HOS formulation proposed in~\cite{VISOR2022} and carry out a comparative analysis with several methods in egocentric settings.
Beyond isolated interactions, the broader goal of this research trajectory is the development of intelligent agents. As thoroughly discussed in recent surveys~\cite{li2026buildingegocentricproceduralai}, building robust egocentric procedural AI assistants requires benchmarks that evaluate complex, multi-step interactions in real-world environments. In line with this necessity, our benchmark design explicitly targets these challenges.

Accurately detecting hand-object interactions is particularly important in industrial contexts, where precise characterization of contact patterns, manipulation strategies, and tool usage is fundamental for monitoring operator performance, diagnosing execution errors, and enabling reliable modeling of physically grounded interactions.

\subsection{Ego-Exo Cross-View VQA}
While recent works have explored cross-view correspondences, learning view-invariant representations~\cite{luo2025viewpoint} or introducing multi-view benchmarks for memory reasoning like Ego-Exo4D~\cite{Grauman_2024_CVPR} and EgoExoMem~\cite{egoexomem}, they generally evaluate unconstrained activities relying on standard matching formulations. Concurrently, Video Question Answering (VQA) has become a standard tool to probe Multimodal Large Language Models (MLLMs)~\cite{egoschema, pov}. Closer to our industrial setting, EgoCross~\cite{Li2025EgoCrossBM} recently benchmarked MLLMs for egocentric VQA using ENIGMA-51~\cite{ragusa2023enigma51}. However, EgoCross operates exclusively on pre-segmented, single-view clips. Overall, existing literature leaves a critical gap: evaluating whether models can perform genuine cross-view temporal grounding in highly structured environments without relying on dataset biases or semantic shortcuts.

To bridge this gap, we introduce the Video-Grounded Cross-View VQA benchmark. To our knowledge, it is the first to combine (i) synchronized ego-exo retrieval, (ii) a controlled, tiered distractor design that strictly isolates action and subject identities to test genuine temporal grounding, and (iii) a realistic, procedurally structured industrial domain. By requiring VLMs to resolve fine-grained procedural actions across viewpoints, differing only in subtle execution details or specific tools, ENIGMA-360 directly exposes the limitations of current architectures. As our results demonstrate, even strong contemporary VLMs collapse to chance-level accuracy once simple semantic shortcuts are systematically removed.

\section{The ENIGMA-360 dataset}\label{dataset}

In this Section, we present ENIGMA-360, a multimodal ego-exo dataset composed of 360 videos acquired in a real industrial laboratory (see Figure~\ref{fig:labels}).

Building upon the preliminary data collection presented in ENIGMA-51~\cite{ragusa2023enigma51}, ENIGMA-360 represents a massive scale-up not just in dataset size and annotation density, but in its expanded research utility. Specifically, this extension uniquely enables several new research questions and benchmark settings that were previously unexplored:

\begin{itemize}
    \item \textbf{Cross-View Representation Learning and Domain Gap:} The strictly synchronized Ego-Exo setup of ENIGMA-360 enables researchers to study the severe domain gap between fixed cameras and wearable devices. This opens the question of how to successfully transfer knowledge from exocentric to egocentric perspectives for industrial tasks.
    
    \item \textbf{Multi-View Multimodal Fusion for Assistive AI:} ENIGMA-360 allows researchers to investigate how fusing fine-grained, localized cues (e.g., hand-tool interaction in the egocentric view) with macroscopic context (e.g., body posture and spatial awareness in the exocentric view) can improve procedural understanding.
    
    \item \textbf{3D Spatial Grounding:} The public release of the laboratory's 3D models in ENIGMA-360 allows for novel benchmark settings focusing on 3D human-object interaction and spatial grounding. These are essential for developing Augmented Reality guidance systems on devices like the HoloLens.
    
    \item \textbf{Novel Benchmark Settings:} Thanks to its extensive and detailed label set, ENIGMA-360 enables the definition of rigorous evaluation protocols for tasks such as Temporal Action Segmentation and Keystep Recognition. Furthermore, to practically demonstrate the unique capabilities of our synchronized streams, we introduce an entirely new task: \textit{Video-Grounded Cross-View VQA}. This zero-shot VQA task challenges state-of-the-art models to identify temporally synchronized action matches across egocentric and exocentric viewpoints, using systematically designed distractors to isolate genuine temporal reasoning from semantic shortcuts.
\end{itemize}

\subsection{The ENIGMA laboratory}
\begin{figure}[ht]
    \centering
    \includegraphics[width=\linewidth]{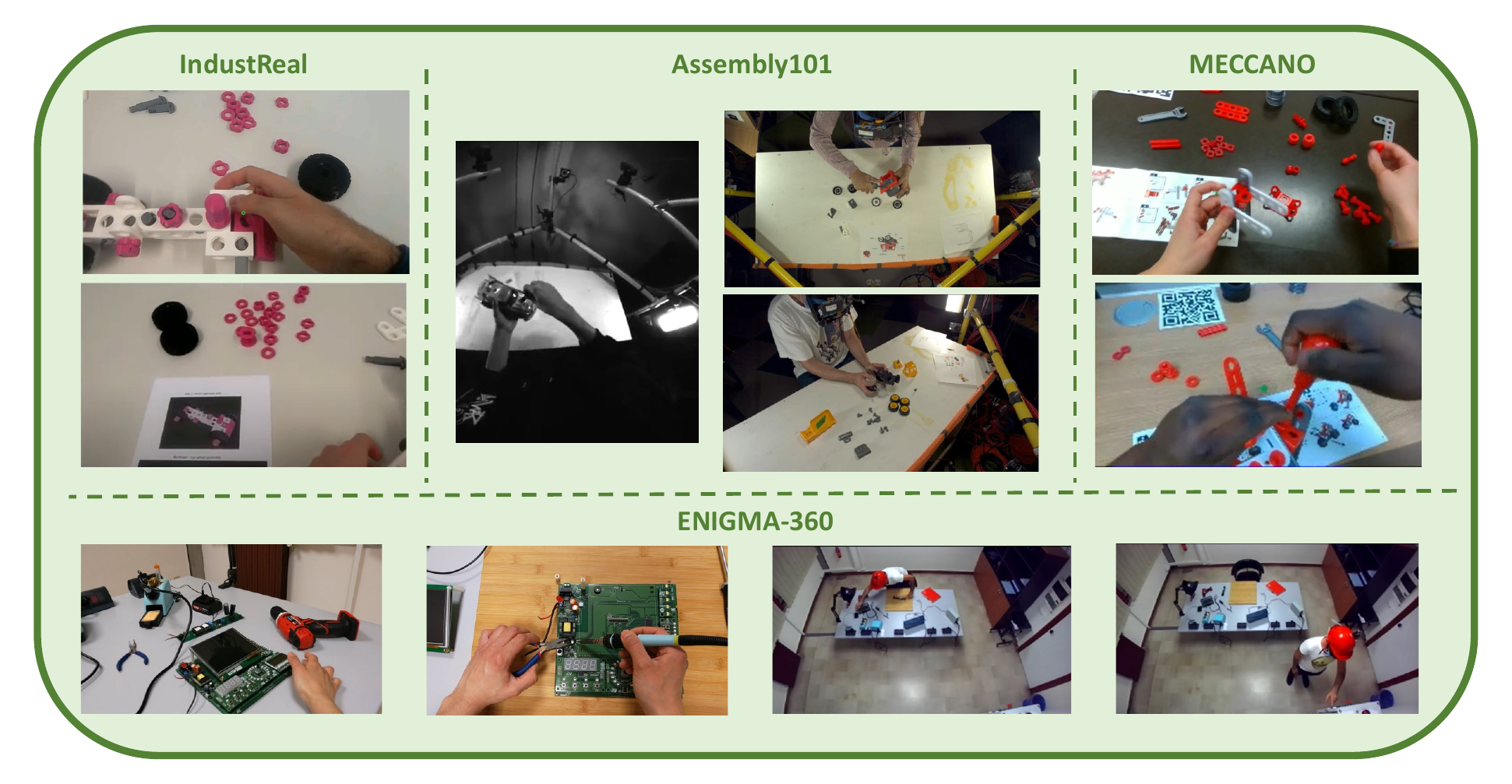}
    \caption{Comparison between state-of-the-art datasets comprising texture-less industrial-like objects (top) and the proposed ENIGMA-360 dataset, which includes realistic industrial objects (bottom).}
    \label{fig:objects}
\end{figure}

In our ENIGMA laboratory there are 25 different objects that can be grouped into fixed objects (such as an \textit{electric panel}) and movable objects (such as a \textit{screwdriver}). Unlike other egocentric datasets~\cite{ragusa2023meccano, sener2022assembly101, schoonbeek2024industreal} that contain industrial-like objects without textures, ENIGMA-360 includes real industrial objects as shown in Figure~\ref{fig:objects}. The complete list of the objects present in the ENIGMA laboratory is as follows: \textit{oscilloscope, 4 sockets, welder station, power supply cables, oscilloscope probe tip, oscilloscope ground clip, low and high voltage boards, pliers, welder base, screwdriver, low voltage board screen, power supply, welder probe tip, electric screwdriver, electric screwdriver battery, register, battery connector, left red button, left green button, right red button and right green button.}

\subsection{Data Acquisition}
To collect data suitable to study human behavior in industrial domain, we designed two procedures consisting of instructions that involve humans interacting with the objects present in the laboratory to achieve the goal of repairing two electrical boards (see Figure~\ref{fig:labels} for visual examples).
These procedures were developed in collaboration with industrial experts to capture realistic human-object interactions in a real industrial domain. Specifically, we designed a procedure for each electrical board---\textit{High Voltage Repair} and \textit{Low Voltage Repair}---that varies in the use of a \textit{screwdriver} or \textit{electric screwdriver} and the electrical component to be soldered (\textit{resistor, capacitor, or transformer}), resulting in four different versions for each procedure. With the provided instructions, we expected users to interact with movable objects (e.g., \textit{``Take the soldering iron’s probe''} or \textit{``Place the electric board on the working area''}) and with fixed machinery (e.g., \textit{``Press the two green buttons on panel A''} or \textit{``Adjust the voltage knob of the power supply to set a voltage of 5 Volts''}). 

\begin{figure}[t!]
    \centering
    \includegraphics[width=0.8\linewidth]{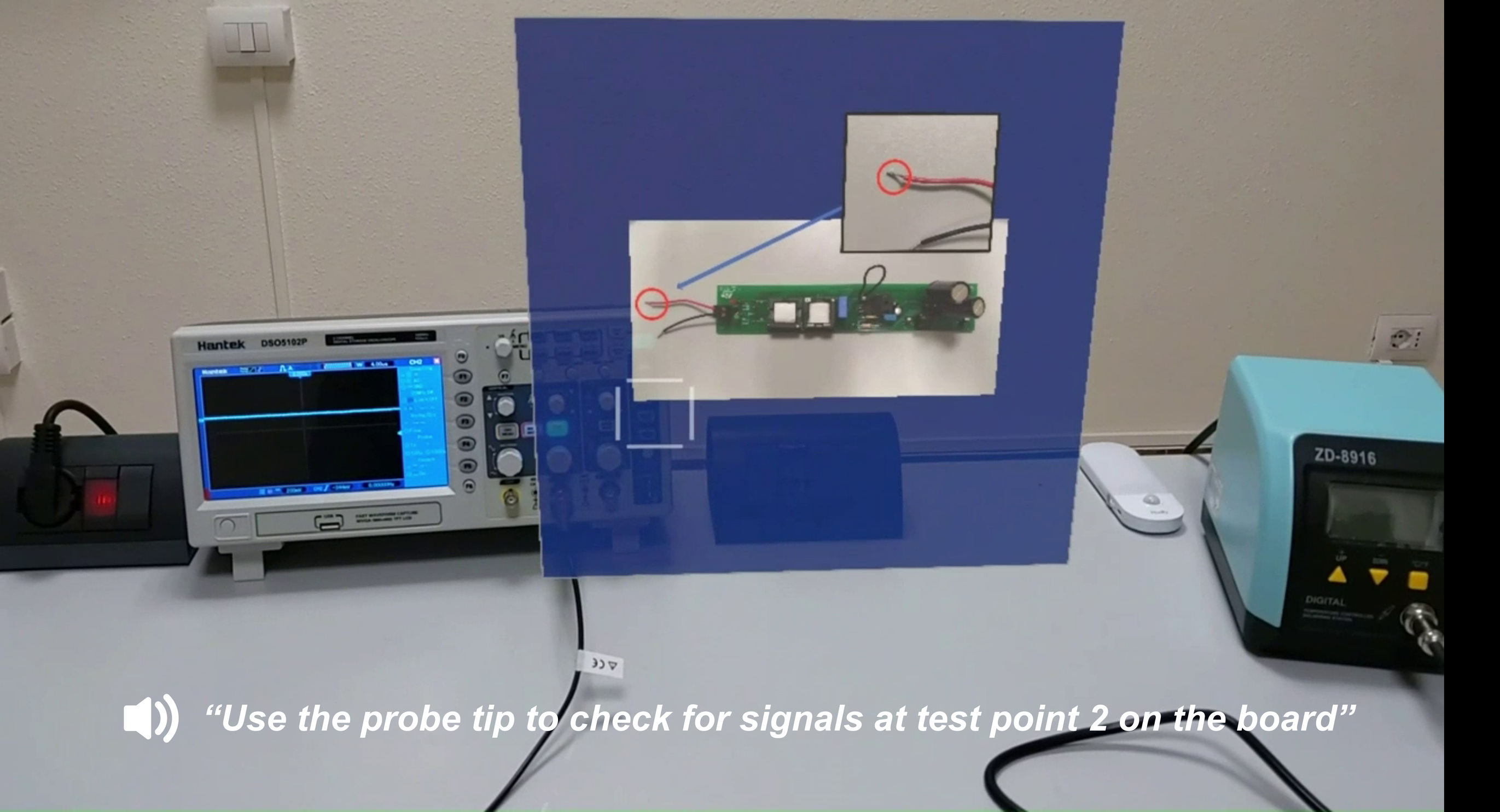}
    \caption{A screenshot captured from the developed application, during the acquisition phase.}
    \label{fig:hololens_app}
\end{figure}

Unlike other acquisition pipelines ~\cite{schoonbeek2024industreal, ragusa2023meccano}, we provide instructions to users without relying on physical manuals (see Figure~\ref{fig:objects}). To achieve this, we developed an application for Microsoft HoloLens 2\footnote{\url{https://www.microsoft.com/en-us/hololens}} using the Unity 3D game engine, which provides instructions to participants during the acquisition process. Specifically, the application aids operators during the acquisition phase by providing audio instructions and displaying images via mixed reality, facilitating complex tasks such as connecting the oscilloscope ground clip (see Figure~\ref{fig:hololens_app}).

\begin{figure}[ht!]
    \centering
    \includegraphics[width=0.98\linewidth]{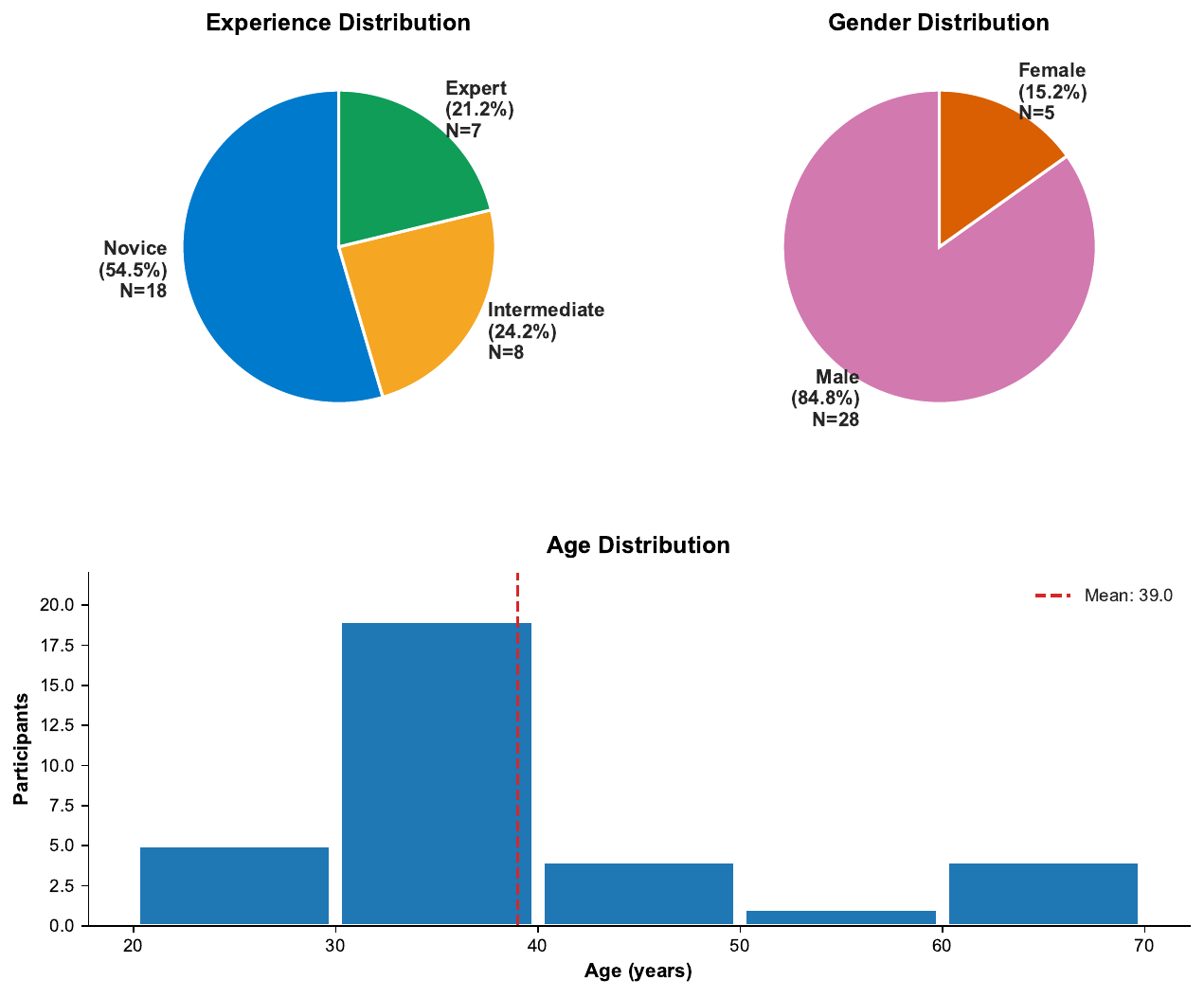}
    \caption{Demographic profile of the ENIGMA-360 dataset. The layout has been restructured into a balanced two-row format to eliminate text cropping, with updated percentages directly matching the $N=33$ participants' raw data.}
    \label{fig:stats_combined}
\end{figure}

An integrated audio guide provides step-by-step instructions during the acquisition process. To generate these audio prompts, a Python script was developed that adopts the gTTS (Google Text-to-Speech)\footnote{\url{https://github.com/pndurette/gTTS}} library, which interfaces with the Google Translate API. The script processes a structured input text file divided into discrete textual blocks and converts each block into an MP3 audio file. Once the device is worn, the operator can interact with the application using predefined voice commands such as \textit{Forward} for playing the audio track of the next set of instructions and \textit{Repeat} for replaying the audio track of the current set of instructions.
We involved 33 participants, aged between 20 and 70, to acquire 360 videos in our laboratory. They had different levels of experience in repairing electrical boards and using industrial tools (see Figure~\ref{fig:stats_combined}). For each recording, we collected the RGB video stream from the Microsoft HoloLens 2, with a resolution of 2272$\times$1278 pixels at 30 frames per second (fps) and the RGB video stream from the ZED camera, with a resolution of 672$\times$376 and a frame rate of 15 fps.

\begin{figure}[t]
  \centering
  \begin{subfigure}[t]{0.38\linewidth}
    \centering
    \includegraphics[width=\linewidth]{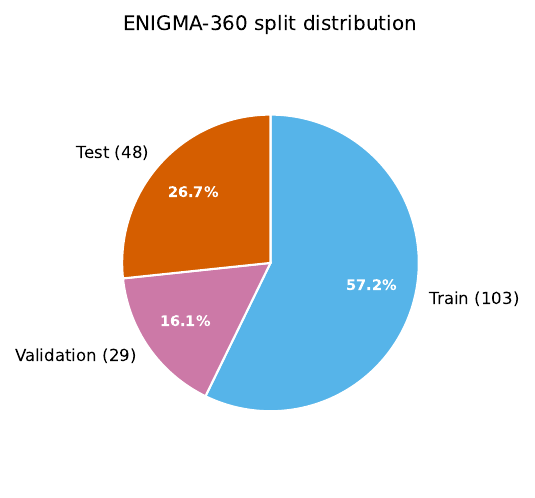}
    \caption{ }
    \label{fig:split}
  \end{subfigure}\hfill
  \begin{subfigure}[t]{0.6\linewidth}
    \centering
    \includegraphics[width=\linewidth]{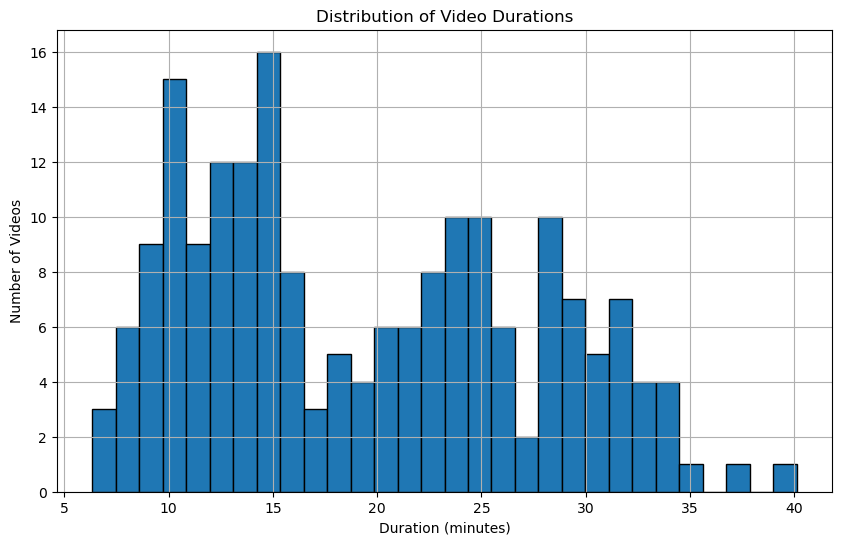}
    \caption{ }
    \label{fig:duration}
  \end{subfigure}
  \caption{ Distribution of data split (a) and number of videos of ENIGMA-360 dataset by duration. (b).}
  \label{fig:participant_demographics_two_plots}
\end{figure}

\begin{figure}[ht!]
    \centering
    \includegraphics[width=0.98\linewidth]{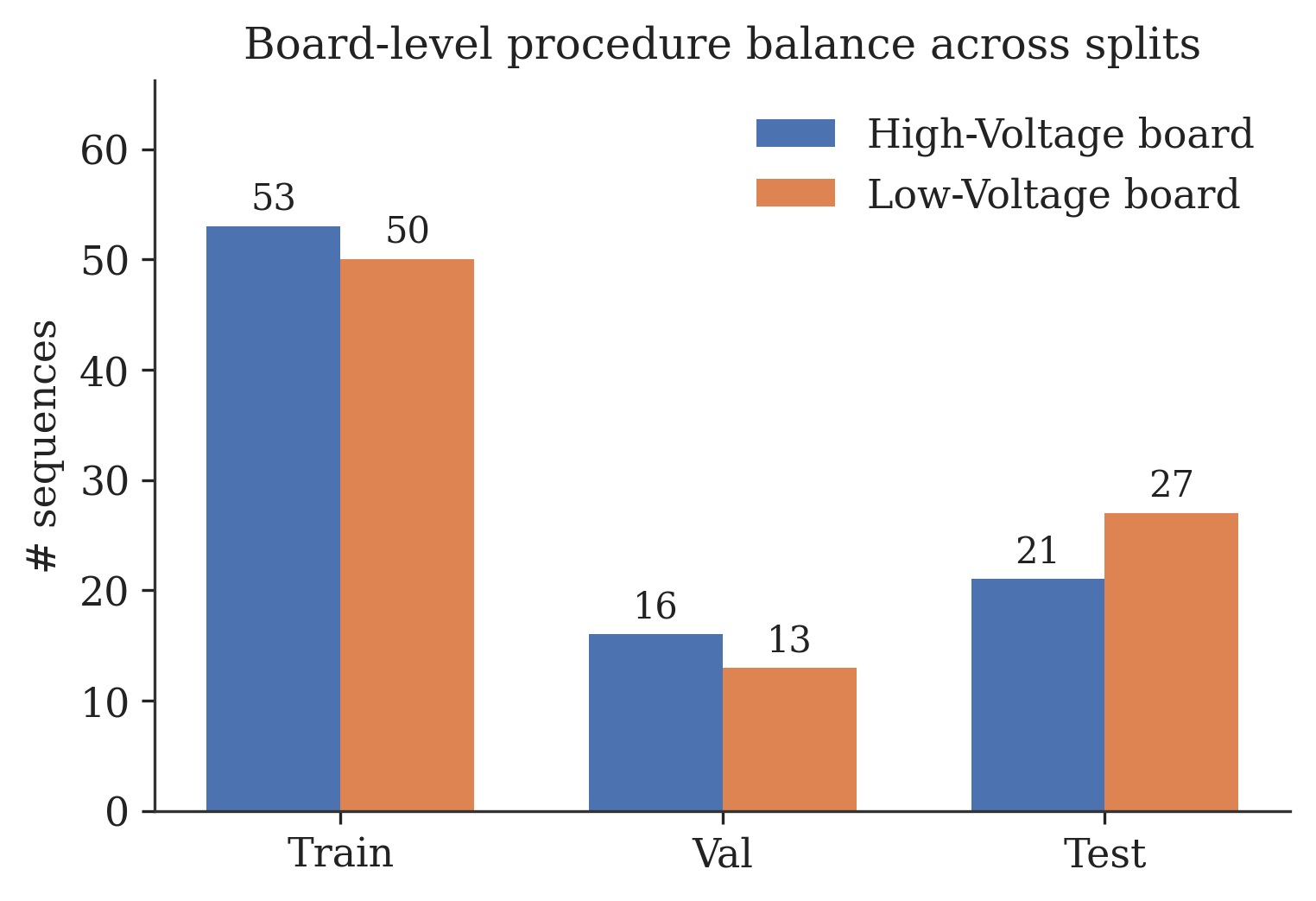}
    \caption{Distribution of the High-Voltage and Low-Voltage board procedures across the training, validation, and test splits. The dataset maintains a close balance between the two main procedures within each subject-disjoint split.}
    \label{fig:board_balance}
\end{figure}

\begin{figure}[ht!]
    \centering
    \includegraphics[width=0.98\linewidth]{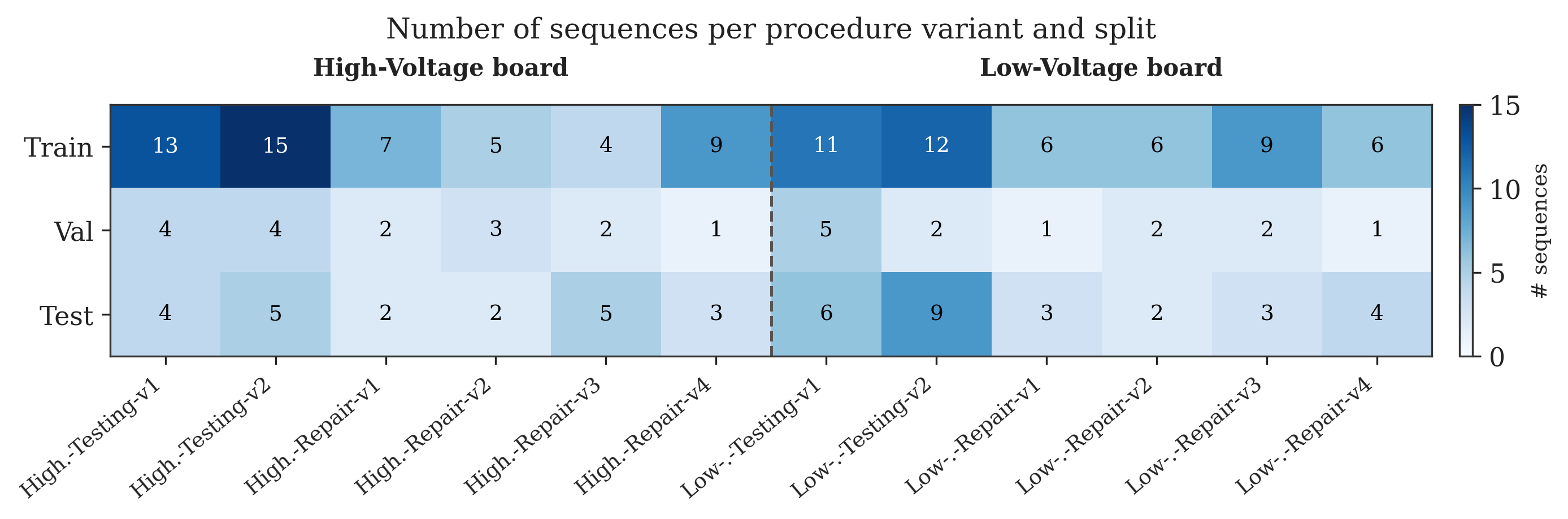}
    \caption{Heatmap detailing the number of sequences per procedure variant across the dataset splits. All 12 scenarios (6 for High-Voltage and 6 for Low-Voltage) are systematically represented across the training, validation, and test sets, ensuring a comprehensive evaluation.}
    \label{fig:heatmap_procedures}
\end{figure}

To ensure a robust evaluation, we adopted a strict partitioning protocol for the dataset. First, the splits are defined over synchronized ego-exo pairs rather than individual videos. Thus, we split ENIGMA-360 into training, validation, and test sets comprising 103, 29, and 48 video pairs, respectively (i.e., a total of 206, 58, and 96 individual videos) (see Figure \ref{fig:participant_demographics_two_plots}-a). Because this pairing is preserved by construction, no ego and exo recording of the same execution can appear in different splits, preventing any information leakage during the cross-view generalization experiments.
Second, the splitting was performed at the participant level to ensure subject-disjoint sets. Each of the 33 participants who contributed to the annotated sequences was assigned to exactly one split (19 in training, 5 in validation, and 9 in test), confirming that no participant contributes data to more than one split. Finally, we ensured a balanced distribution of the procedures and their variants. The dataset is organized around the two main board-level repair procedures (High-Voltage and Low-Voltage), each comprising 6 recorded variants: 2 variants isolating the initial testing phase, and 4 variants (v1–v4) covering the full repair procedure, for a total of 12 scenarios. Figure \ref{fig:board_balance} illustrates the overall balance between the two board-level procedures across the splits, demonstrating that they are evenly distributed. Furthermore, Figure \ref{fig:heatmap_procedures} details the number of sequences per specific variant, confirming that every scenario is adequately represented across the training, validation, and test sets.
The average duration of the recorded videos is 18.59 minutes, resulting in a total of approximately 111.54 hours of videos. Figure \ref{fig:participant_demographics_two_plots}-b reports the number of videos over their duration.

\begin{figure}[t]
	\centering
	\includegraphics[width=0.99\linewidth]{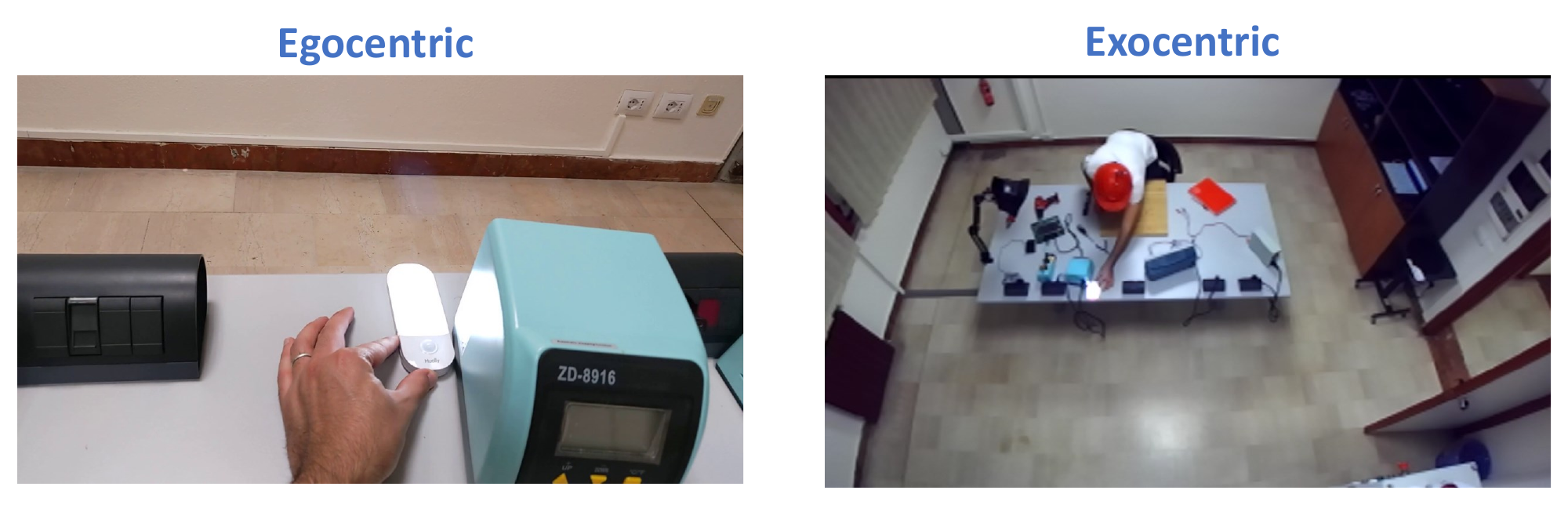}
	\caption{Temporal synchronization of ego and exo videos using the lamp light as a reference.}
	\label{fig:lamp_sync}
\end{figure}

To ensure temporal alignment between the audio instructions and the egocentric video content, we synchronized the two by assigning timestamps each time the user advanced to the next instruction. To temporally synchronize the two RGB streams, egocentric and exocentric, we asked each participant to turn on a lamp placed on the laboratory workbench. The lamp's light served as a reference point for temporal alignment, as shown in Figure~\ref{fig:lamp_sync}. In this way it is possible to use the set of annotations independently from the different chosen point of view.

\begin{table}[ht!]
	\centering
	\scriptsize 
	\begin{tabular}{|c|p{0.38\textwidth}|c|p{0.38\textwidth}|}
		\hline
		\textbf{ID} & \textbf{Keystep}                       & \textbf{ID} & \textbf{Keystep}                   \\
		\hline
		001         & Press the two green buttons on panel A & 035         & Randomly rotates the position knob \\
		002         & Moves to the workbench and sits down   & 036         & Uses the probe tip at test point 3 \\
		003         & Takes the high-voltage card            & 037         & Uses the probe tip at test point 4 \\
		004         & Places the card on the work area       & 038         & Uses the probe tip at test point 5 \\
		005         & Connects battery to screwdriver        & 039         & Uses the probe tip at test point 7 \\
		006         & Removes screws with screwdriver        & 040         & Disconnects ground clip and probe  \\
		007         & Fixes card to workbench                & 041         & Deactivates CH2 on oscilloscope    \\
		008         & Turns on power via tower switch        & 042         & Turns off oscilloscope             \\
		009         & Adjusts current knob until LED on      & 043         & Uses probe tip at test point 6     \\
		010         & Adjusts voltage knob to 5V             & 044         & Reinserts screws on work area      \\
		011         & Connects power cables to card          & 045         & Disconnects screwdriver battery    \\
		012         & Turns off card power supply            & 046         & Turns on battery charger           \\
		013         & Resets current and voltage to 0        & 047         & Notes results in the register      \\
		014         & Disconnects power supply cables        & 048         & Presses the two red buttons        \\
		015         & Removes card using screwdriver         & 049         & Touches transformer pin back side  \\
		016         & Turns on soldering iron                & 050         & Turns off battery charger          \\
		017         & Sets iron to 200° with UP button      & 051         & Removes workbench screws           \\
		018         & Grasps capacitor with pliers           & 052         & Fixes card with screwdriver        \\
		019         & Takes the soldering iron tip           & 053         & Removes card with screwdriver      \\
		020         & Touches capacitor pin 1                & 054         & Reinserts screws with screwdriver  \\
		021         & Touches capacitor pin 2                & 055         & Takes low-voltage card             \\
		022         & Places pliers on workbench             & 056         & Grasps resistor with pliers        \\
		023         & Places soldering tip                   & 057         & Touches resistor pin 1             \\
		024         & Puts card vertically                   & 058         & Touches resistor pin 2             \\
		025         & Touches cap. pin 1 on back             & 059         & Touches resistor pin 1 back        \\
		026         & Touches cap. pin 2 on back             & 060         & Touches resistor pin 2 back        \\
		027         & Touches transf. pin 2 on back          & 061         & Unscrews 4 screws back card        \\
		028         & Sets iron temp to 160°                & 062         & Removes display from card          \\
		029         & Turns off soldering iron               & 063         & Connects display with screwdriver  \\
		030         & Turns on oscilloscope                  & 064         & Unscrews 4 screws back card        \\
		031         & Activates CH2 on oscilloscope          & 065         & Connects display with screwdriver  \\
		032         & Clips probe to test point 1            & 066         & Grasps transformer with pliers     \\
		033         & Tests signal at test point 2           & 067         & Touches transformer pin 1          \\
		034         & Presses Auto Set button                & 068         & Touches transformer pin 2          \\
		\hline
	\end{tabular}
    \caption{Keystep Taxonomy of ENIGMA-360.}\label{tab:keysteps}
\end{table}

\subsection{Data Annotation}
The ENIGMA-360 dataset was collected and annotated to facilitate the study of human behavior in real industrial environments. Similarly to recent datasets~\cite{sener2022assembly101, Grauman_2024_CVPR}, ENIGMA-360 is a multi-view dataset, where different perspectives can be used independently or in combination to address different tasks. 
We provide temporal annotations for keysteps understanding that are useful to solve tasks which take into account the temporal dimension, such as temporal action segmentation (Section~\ref{sec:tas}) and keystep recognition (Section~\ref{sec:keysteps}). We additionally labeled keyframes where human-object interactions occur to study tasks such as Egocentric Human-Object Interaction Detection (Section~\ref{sec:ehoi}). A key distinguishing feature of ENIGMA-360 is the multi-task co-occurrence of annotations. To the best of our knowledge, our dataset is the first in the industrial domain to simultaneously offer co-located annotations for temporal action segmentation, Hand-Object Interaction (HOI) detection on real professional components, and keystep recognition on the exact same video streams. This unified taxonomy allows for cross-task reasoning (e.g., leveraging HOI spatial context to improve temporal boundary segmentation), a paradigm that cannot be explored on existing single-task datasets.

\begin{figure}[ht!]
    \centering
    \includegraphics[width=\linewidth]{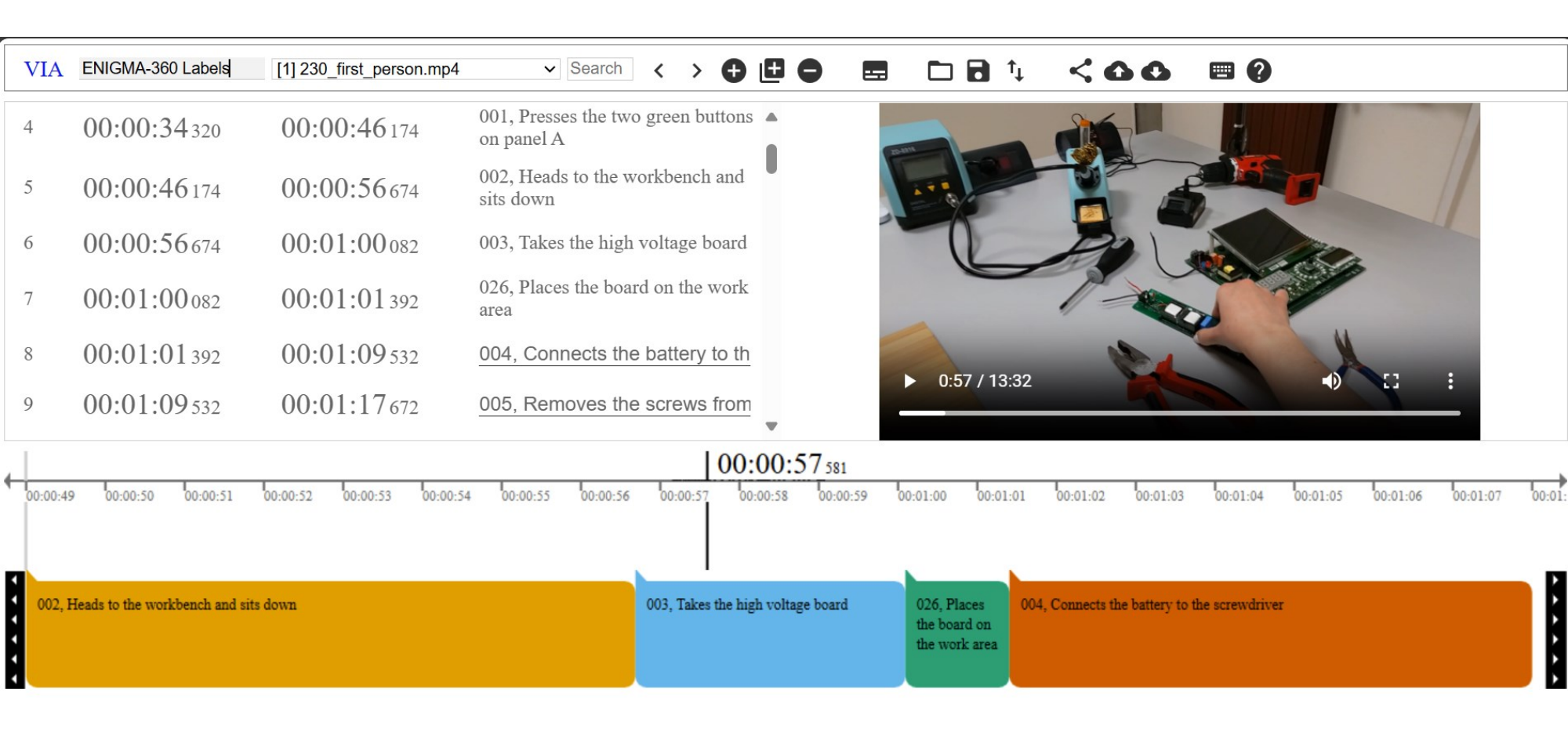}
    \caption{Example of Temporal Segmentation of Keysteps using our customized VIA interface.}
    \label{fig:VIA}
\end{figure}

\begin{figure}[ht!]
    \centering
    \includegraphics[width=0.98\linewidth]{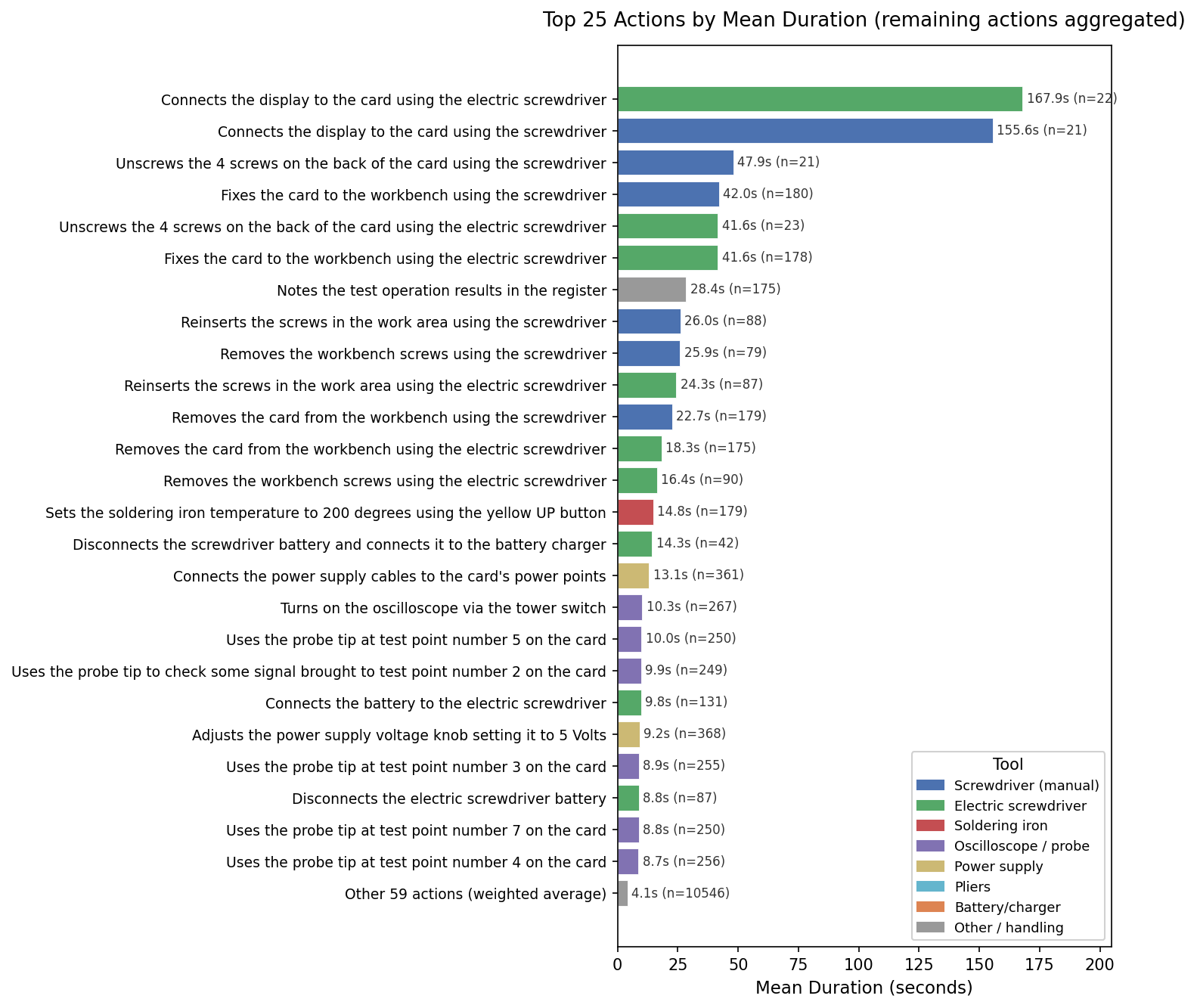}
    \caption{Distribution of keystep annotations of ENIGMA-360.}
    \label{fig:label_dist}
\end{figure}

\textbf{Keystep Temporal Annotations}. We identified 68 distinct keysteps distributed across the two designed procedures (the complete list of keysteps is provided in Table~\ref{tab:keysteps}). Starting from the logs generated by the developed application, where each step is associated with a timestamp, we engaged multiple annotators to refine the start times and determine the corresponding end times for each keystep. In particular, we customized VIA Subtitle Annotator Tool\footnote{\url{https://www.robots.ox.ac.uk/~vgg/software/via/demo/via_subtitle_annotator.html}} to allow the loading of these logs and to allow annotators to facilitate the labeling process. At the end of this process, for each keystep we obtained its start and end timestamps (see Figure~\ref{fig:VIA}). 
With this procedure, we annotated a total of 14,556 keysteps throughout the dataset, with an overall average duration of 7.92 seconds. The most frequent keystep is \textit{“Presses the oscilloscope's Auto Set button”} with 1,542 instances, followed closely by \textit{“Randomly rotates the position knob”} with 1,515 instances, and \textit{“Adjusts the power supply voltage knob setting”} with 368 instances. Conversely, the least frequent keysteps include \textit{“Connects the display to the card using the screwdrivers”} with 22 instances (mean length 167.9 s), \textit{“Unscrews the 4 screws on the back of the card”} with 21 instances (mean length 47.9 s), and another \textit{“Connects the display to the card using the screwdrivers”} with 21 instances (mean length 155.6 s). Figure~\ref{fig:label_dist} shows the distribution of temporal annotations in ENIGMA-360, considering our taxonomy.

\textbf{Interaction Keyframes.} We identified the interaction key frames within the egocentric subset of the ENIGMA-360 dataset, where the point of view facilitated more accurate detection of human-object interactions by the annotators. In addition, we sampled frames without interactions to create a set of negative examples representing the absence of human-object interactions. Each identified key frame was assigned a timestamp, resulting in a total of 14,036 annotated interactions. Figure~\ref{fig:labels} shows examples of these annotated interaction key frames.

\begin{figure}[ht]
    \centering
\includegraphics[width=\linewidth]{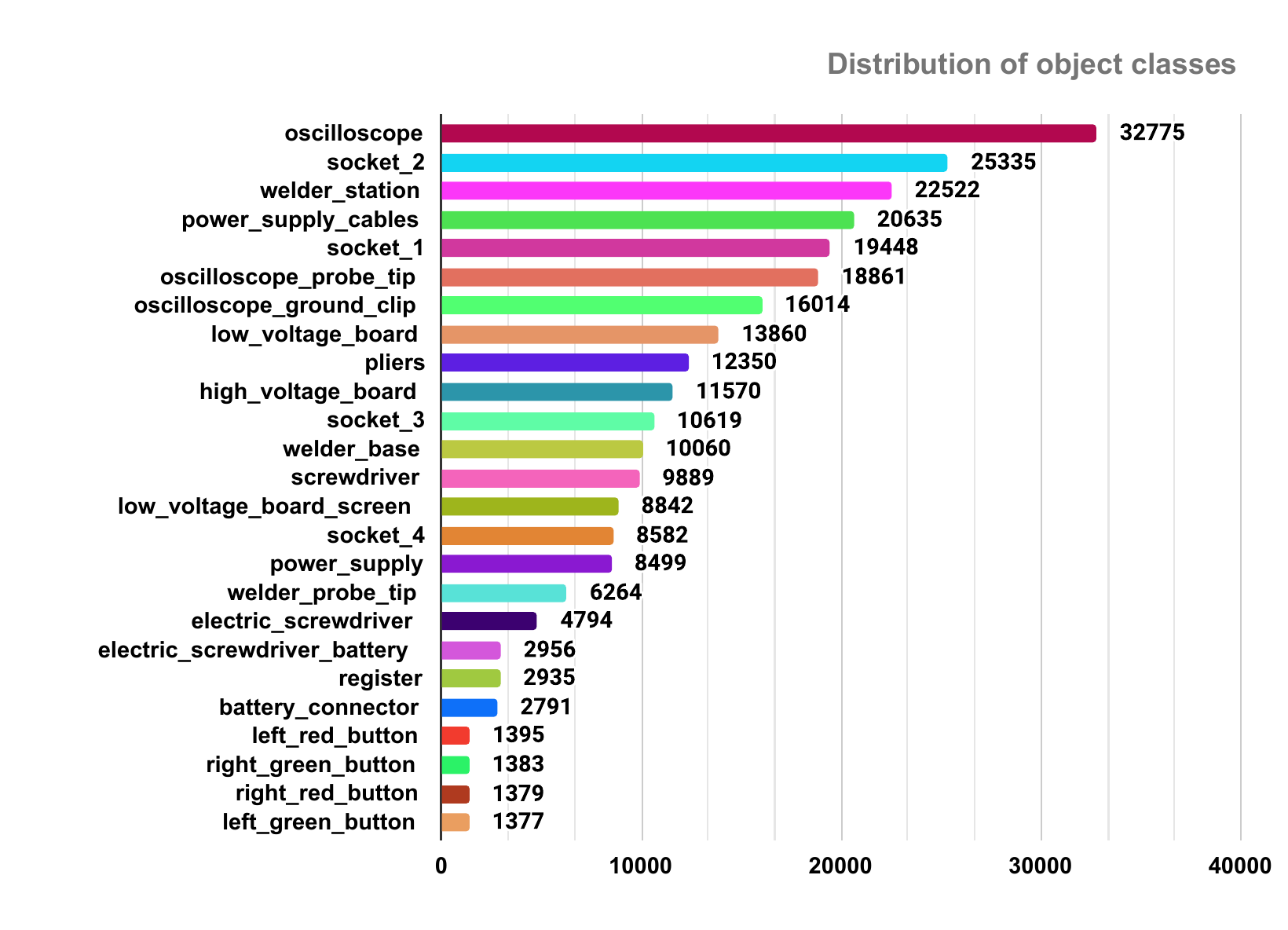}
    \caption{Object distribution in the annotated interaction keyframes of ENIGMA-360.}
    \label{fig:objects_distribution}
\end{figure}

\textbf{Object Annotations.} We considered 25 object classes which include both fixed (e.g., electric panel, oscilloscope) and movable objects (e.g., screwdriver, pliers) to assign a class to the objects present in the interaction key frames. Each object annotation consists in a tuple $(class, x, y, w, h, state)$, where $class$ represents the class of the object, $(x, y, w, h)$ are the
2D coordinates which define the bounding box around the object in the frame, and the $state$ indicates if the object is involved in an interaction or not (active object vs. passive object). With this annotation procedure, we annotated 275,135 objects. Figure~\ref{fig:objects_distribution} reports the distribution of the objects over the annotated interaction keyframes.

\textbf{Hands Annotations.} To annotate hand bounding boxes within the interaction key frames, we adopted a semi-automatic approach to streamline the process. Initially, we generated pseudo-labels using a hand-object detection model~\cite{Shan2020UnderstandingHH}, focusing exclusively on hand-related outputs. These preliminary annotations were then manually refined by annotators, who corrected hand side (left/right) and linked each hand to its corresponding previously labeled active object. This strategy enabled the efficient annotation of a total of 56,473 hands.

\begin{figure}[ht!]
    \centering
\includegraphics[width=0.7\linewidth]{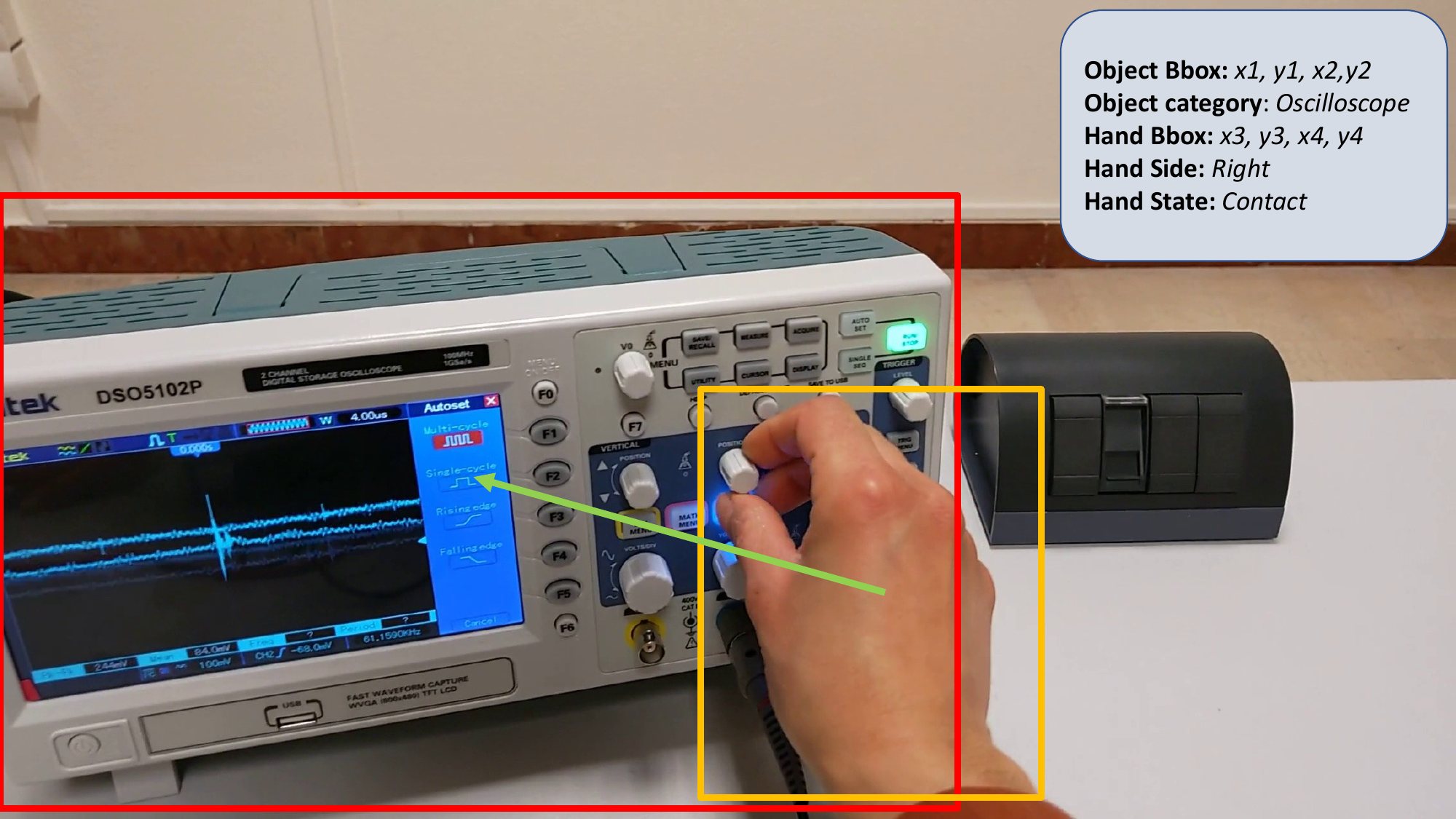}
    \caption{Sample of an interaction key frame with the correspondent annotations: hands and active object bounding boxes, 2) hands side, 3) hand contact state, 4) hand-object relationships, and 5) object category.}
    \label{fig:ehoi}
\end{figure}

\textbf{Egocentric Hand-Object Interaction Annotations.} For each of the interaction key frames, we considered: 1) hands and active object bounding boxes, 2) hand side (left and right), 3) hand contact state (contact and no contact), 4) hand-object relationships, and 5) object categories (see Figure~\ref{fig:ehoi}).  Following this procedure, we annotated 12,597 interaction frames, 17,363 hands of which 10,043 were in contact, and 9.342 active objects.

\subsubsection{Additional Pseudo-Labels and Features}
To further enrich the ENIGMA-360 dataset, we release a set of pseudo-labels designed to enhance its impact and support the study of multiple tasks through additional signal modalities.

\begin{figure}[t!]
    \centering
\includegraphics[width=\linewidth]{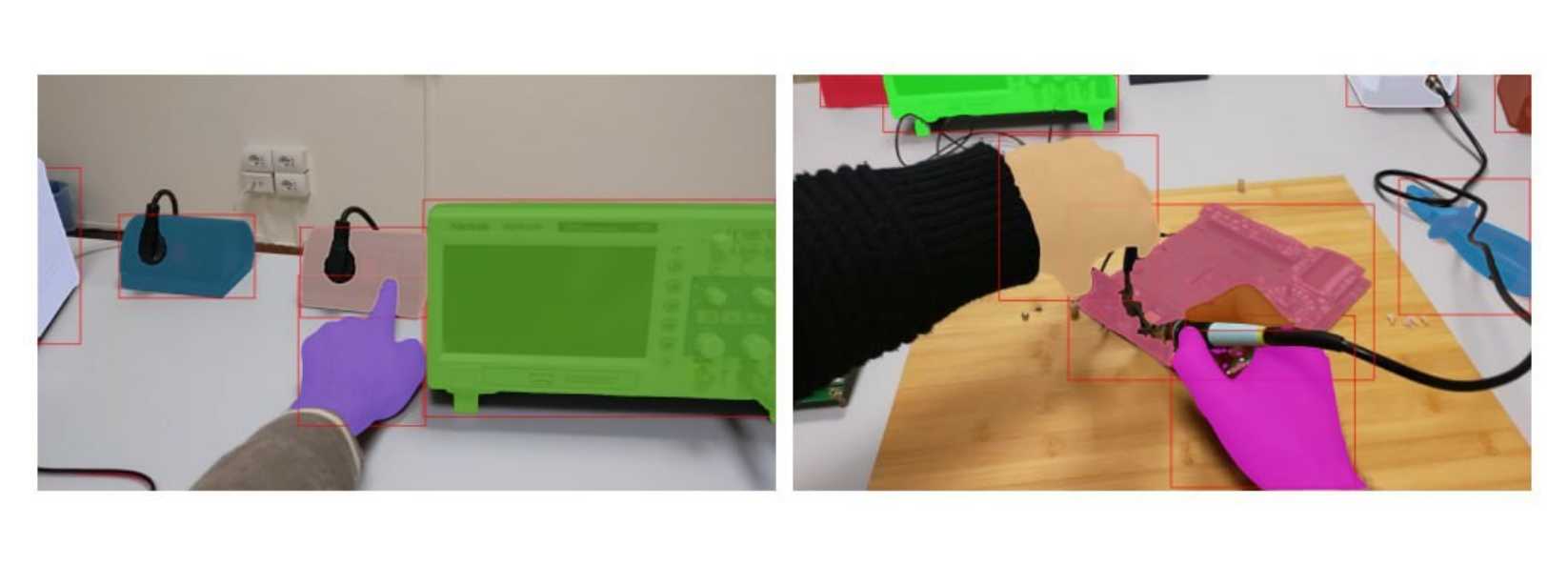}
    \caption{Samples of semantic segmentation masks obtained with SAM-HQ.}
    \label{fig:sam}
\end{figure}

\textbf{Semantic Segmentation Masks.} We provide segmentation masks for the hands and the objects using SAM-HQ~\cite{sam_hq}, an advanced extension of the Segment Anything Model (SAM), designed to enhance the segmentation of complex objects. It retains SAM's promptable design, efficiency, and zero-shot generalizability while accurately segmenting any object. Considering the challenging nature of objects like small clips and wires in our industrial scenario, we opted to use SAM-HQ as it proves to be a suitable solution for precise segmentation. For the mask extraction, we used the SAM code provided in the official repository\footnote{\url{https://github.com/facebookresearch/segment-anything}}. With SAM, we obtained a total of 197.814 hand masks and 1.435.006 object masks. Figure~\ref{fig:sam} reports some examples of segmentation masks obtained with SAM-HQ.

\textbf{DINOv2 Features.} We used DINOv2~\cite{oquab2023dinov2} to extract features from video frames. This model excels at learning high-level visual representations without the need for labeled data. By processing each frame independently, DINOv2 captures rich spatial features that can be aggregated to model temporal dynamics across sequences. Its ability to produce robust, general-purpose embeddings makes it particularly well-suited for downstream tasks such as temporal action segmentation, action recognition, and scene understanding in video analysis. We used the \textit{large} variant of DINOv2 to extract 1024-dimensional per-frame features, using the official implementation\footnote{\url{https://github.com/facebookresearch/dinov2}}. The extracted features were organized into an LMDB structure for efficient storage and retrieval.

\textbf{3D models.} To support the generation of synthetic data for training scalable domain adaptation and generalization methods, we reconstructed the 3D environment of the laboratory, including all industrial objects, using high-resolution scans obtained with the Matterport\footnote{\url{https://matterport.com/}} and ARTEC EVA\footnote{\url{https://www.artec3d.com/portable-3d-scanners/artec-eva}} devices.

\subsection{Quantitative Execution Variability}

While the high-level procedures in ENIGMA-360 are guided by standard TTS prompts to mirror strict industrial Standard Operating Procedures (SOPs), the low-level execution remains highly unconstrained. To quantitatively demonstrate that our dataset captures natural human variability, we analyzed the execution times of identical keysteps across different participants.

As shown in Table~\ref{tab:execution_variability}, the execution time for the same operation exhibits massive variance. For instance, the keystep \textit{"Connects display with electric screwdriver"} has a global average duration of 172.62 seconds, but ranges significantly from 72.84 seconds for experts to 213.00 seconds for novices, with a standard deviation of 87.36 seconds. 

Furthermore, by analyzing the overall trial durations, we observed that novice users naturally introduce micro-variations that differ entirely from the smooth executions of expert users. As detailed in Table~\ref{tab:overall_duration}, a novice user’s complete trial takes on average $22.05\%$ longer than an expert's trial. This delay is entirely due to natural hesitations, repeated micro-actions, and implicit exploratory behaviors (e.g., repeatedly checking the equipment or re-adjusting the grip). These statistics demonstrate that despite the scripted macroscopic instructions, ENIGMA-360 inherently preserves the rich, unconstrained behavioral diversity typical of "in-the-wild" human operators.

\begin{table*}[t]
\centering
\caption{Quantitative Execution Variability: Top 5 keysteps exhibiting the highest temporal variance across users. The massive gap between Expert and Novice execution times demonstrates the unconstrained behavioral diversity captured by ENIGMA-360. Times are in seconds.}
\label{tab:execution_variability}
\resizebox{\textwidth}{!}{
\begin{tabular}{lcccccc}
\toprule
\textbf{Keystep Description} & \textbf{Global Mean} & \textbf{Min Time} & \textbf{Max Time} & \textbf{Std Dev} & \textbf{Expert Mean} & \textbf{Novice Mean} \\ 
\midrule
Connects display with electric screwdriver & 172.62 & 65.92 & 369.34 & \textbf{87.36} & 72.84 & 213.00 \\ 
Connects display with screwdriver & 146.90 & 50.03 & 223.81 & \textbf{49.72} & 170.76 & 136.63 \\ 
Fixes card to workbench & 60.07 & 0.00 & 174.17 & \textbf{44.04} & 39.02 & 57.72 \\ 
Disconnects screwdriver battery & 23.28 & 9.89 & 107.95 & \textbf{32.02} & 10.07 & 35.91 \\ 
Fixes card with screwdriver & 45.99 & 20.25 & 167.57 & \textbf{30.79} & 37.86 & 50.07 \\ 
\bottomrule
\end{tabular}
}
\end{table*}

\begin{table}[h]
\centering
\caption{Overall Procedure Duration: Experts vs. Novices. The aggregated execution time highlights the impact of micro-improvisations and hesitations, resulting in a significant temporal delay for novice users.}
\label{tab:overall_duration}
\begin{tabular}{lccc}
\toprule
\textbf{User Expertise} & \textbf{Total Average Time (s)} & \textbf{Difference (s)} & \textbf{Increase (\%)} \\ 
\midrule
Expert & 4085.85 & - & - \\ 
Novice & 4986.81 & +900.96 & +22.05\% \\ 
\bottomrule
\end{tabular}
\end{table}

\subsection{Annotation Quality}

\begin{figure}[t]
    \centering
    \begin{subfigure}[b]{0.48\textwidth}
        \centering
        \includegraphics[width=\textwidth]{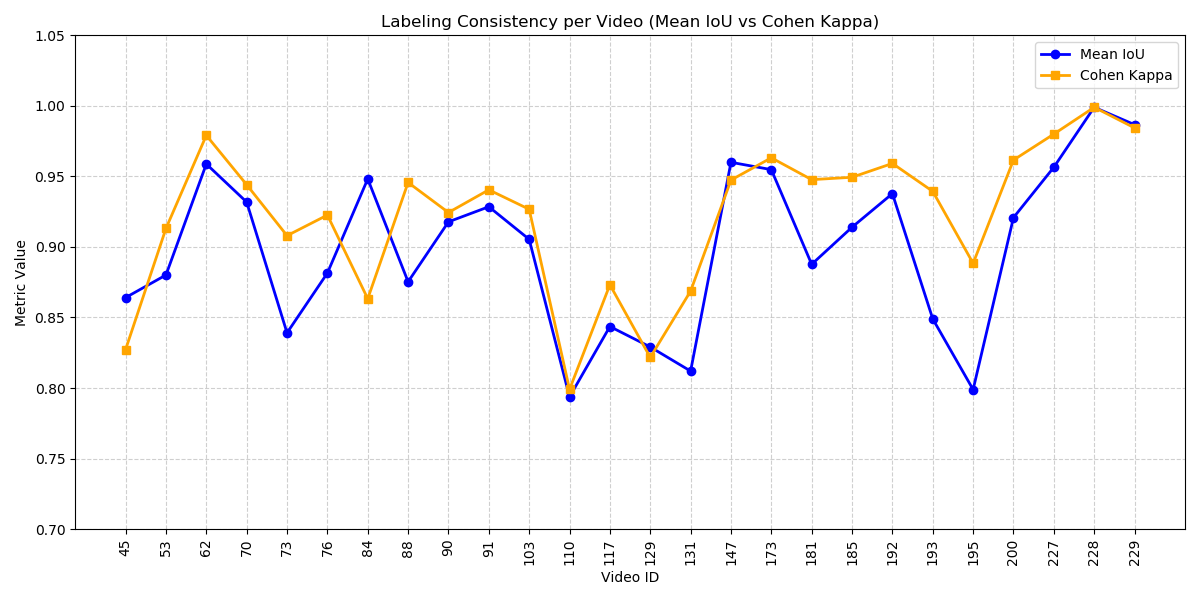} 
        \caption{Temporal Annotation Consistency}
        \label{fig:temporal_agreement}
    \end{subfigure}
    \hfill
    \begin{subfigure}[b]{0.48\textwidth}
        \centering
        \includegraphics[width=\textwidth]{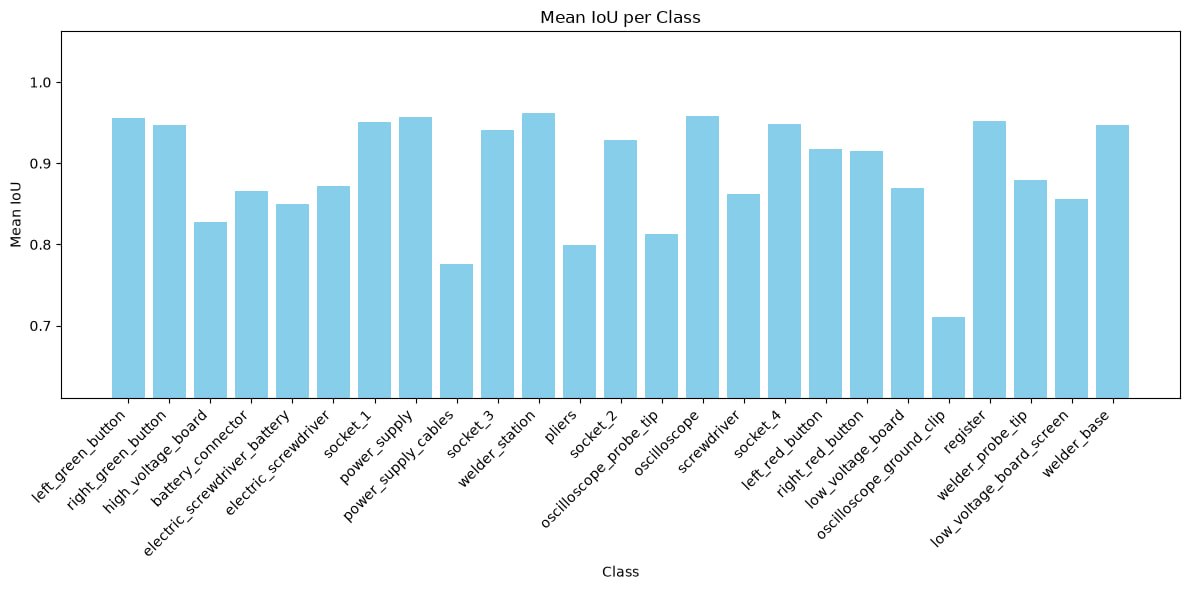} %
        \caption{Spatial Localization Consistency (Mean IoU per Class)}
        \label{fig:spatial_iou_class}
    \end{subfigure}
    \caption{Inter-annotator agreement results validating the robustness of ENIGMA-360's annotation pipelines. (a) Temporal consistency across a subset of 26 videos. (b) Spatial localization consistency, over the 25 object classes.}
    \label{fig:annotation_quality}
\end{figure}

To quantitatively assess the reliability of our annotation pipelines, we conducted rigorous inter-annotator agreement studies for both temporal and spatial labels.For \textbf{temporal annotations}, a randomly selected subset of 26 videos was independently processed by a second annotator following identical guidelines. We measured Cohen's Kappa for keystep class assignment ($\kappa = 0.9222$) and mean temporal Intersection over Union (IoU = 0.8990) for boundary positions. These results confirm the high consistency of our temporal annotation process, which we primarily attribute to the objective pre-labeling provided by the HoloLens logs. As illustrated in Figure~\ref{fig:annotation_quality}-a, the distribution of Mean IoU and Cohen's Kappa across the evaluated subset demonstrates highly consistent agreement performance with minimal variance across all videos.
For \textbf{spatial annotations}, we evaluated both localization precision and semantic classification consistency. Regarding object localization, a randomly selected subset of keyframes was independently re-annotated by a second expert, yielding a strong mean object bounding box IoU of 0.8799. Figure~\ref{fig:annotation_quality}-b details the Mean IoU per class, confirming robust localization performance across the vast majority of our 25 object categories. Furthermore, to evaluate the semantic consistency of hand-object interaction states (Contact/Take vs. Decontact/Release), a dedicated pool of 318 contact keyframes was blindly re-evaluated by 6 independent annotators. Even when factoring in highly ambiguous and severely occluded edge cases, the independent annotators achieved an $88.68\%$ agreement rate with the original dataset labels. These comprehensive metrics conclusively demonstrate a robust shared understanding of the interaction guidelines and the overall reliability of the ENIGMA-360 spatial annotations.

To properly contextualize these figures, we compare our inter-annotator agreement metrics directly with those reported in similar procedural and interaction-focused datasets. First, regarding temporal boundaries, the mean temporal IoU in ENIGMA-360 is 0.8990, compared to typical human agreement ranges of 0.60 to 0.75 in standard action datasets like ActivityNet\cite{Heilbron_2015_CVPR}. Furthermore, we achieve a temporal Cohen's Kappa of 0.9222, which is significantly higher than the mean inter-annotator agreement of $\kappa = 0.572$ reported in the large-scale multi-view procedural dataset Assembly101\cite{sener2022assembly101} Second, for spatial localization, our mean bounding box IoU is 0.8799, which is highly competitive when compared to the typical human agreement of 0.80 to 0.85 reported in dense object datasets such as Visual Genome\cite{krishna2017visual}. Finally, regarding HOI semantic classification, our mean label agreement is 88.68\%. This number is highly comparable, and indeed superior, to existing representative datasets such as the egocentric benchmark EPIC-Contact\cite{bansal2026hopformer}, which reports a moderate inter-annotator agreement of $\kappa \approx 0.61$ for contact state estimation. These comparisons demonstrate that ENIGMA-360 successfully overcomes the inherent difficulty of annotating fine-grained hand-object interactions under severe motion blur and multi-view self-occlusions.

To provide full transparency on the remaining annotator inconsistencies across all reported metrics, we conducted a qualitative breakdown of the disagreements, explicitly distinguishing between inherent task ambiguity and standard annotator error:
\begin{itemize}
    \item \textbf{Temporal Boundaries ($\sim 10\%$ inconsistency):} The majority of these discrepancies stem from inherent task ambiguity related to trailing actions (e.g., defining the exact frame an action concludes when a hand slowly retracts after tightening a screw). Standard annotator error in this modality is minimal, mostly consisting of slight frame-level misalignments when manually refining the exact start or end boundaries, leading to minor temporal shifts between independent annotators.
    \item \textbf{Spatial Bounding Boxes ($\sim 12\%$ inconsistency):} Inherent visual ambiguity is the primary factor here, driven by severe motion blur during rapid egocentric head movements and heavy hand-tool self-occlusions, which force annotators to estimate the occluded physical extent of an object. Standard annotator error is primarily limited to ``loose" drawing, where the bounding box is drawn slightly larger than the strict object margins.
    \item \textbf{Semantic HOI Classification ($\sim 11\%$ inconsistency):} We estimate that roughly $75\%$ of semantic disagreements stem from inherent visual ambiguities. The main causes are micro-occlusions and sub-millimeter boundary definitions, particularly when a fine-grained tool (e.g., a soldering iron or oscilloscope probe) hovers just above the board, making the exact transition between \textit{Contact} and \textit{Decontact} highly subjective. The remaining $25\%$ is attributable to standard annotator error, such as briefly misclassifying visually identical peripheral objects (e.g., confusing two adjacent disconnected cables) or overlooking a secondary, passive interaction while focusing on the primary active tool.
\end{itemize}

\section{Benchmark and baselines results}\label{benchmark}

We evaluate and provide baseline results for 4 tasks focusing on human behavior understanding in the industrial domain: 1) Temporal Action Segmentation, 2) Keystep Recognition, 3) Egocentric Human-Object Interaction Detection, and 4) Video-Grounded Cross-View VQA. Since the dataset has a rich set of annotations, we hope that the research community will explore different applications and tasks. All models are trained and evaluated completely independently with separate weights. Furthermore, a complete computational cost comparison, detailing the GPU memory usage and training times across all these evaluated baselines, is provided in the Supplementary Material.

\subsection{Temporal Action Segmentation}
\label{sec:tas}
Temporal Action Segmentation involves recognizing and segmenting distinct actions within long, untrimmed video sequences. The aim is to assign an action label, belonging to our taxonomy, to each frame or temporal segment of the video, accurately identifying the start and end boundaries of each action. This task requires not only precise classification but also a coherent temporal understanding of the sequence and duration of actions.

\subsubsection{Baselines and Evaluation Measures}
We address the Temporal Action Segmentation (TAS) task by leveraging several state-of-the-art methodologies~\cite{singhania2021coarse}. In particular, we adopted C2F-TCN~\cite{singhania2021coarse}, ASFormer~\cite{chinayi_ASformer}, LTContext~\cite{ltc2023bahrami}, MSTCN++~\cite{li2020ms},  DiffAct~\cite{liu2023diffusion}, and FACT~\cite{lu2024fact}.
Each baseline was trained on the egocentric subset and evaluated on both the egocentric and exocentric test sets. Similarly, each baseline was also trained on the exocentric subset and tested on both the egocentric and exocentric test sets.

Baselines have been evaluated using standard metrics following previous works~\cite{sener2022assembly101,lea2017temporal,ding2022temporal}. In particular, we report two segment-based evaluation measures, namely edit score and $F1\{\beta\}$ score, and a frame-based evaluation measure, namely Mean over Frames (MoF). Following the protocol adopted in~\cite{lu2024fact}, we consider two variants of MoF: the standard version, which excludes background frames (Acc), and another that includes background frames (AccB).

\subsubsection{Results}
\begin{figure}[t]
    \centering
    \includegraphics[width=\linewidth]{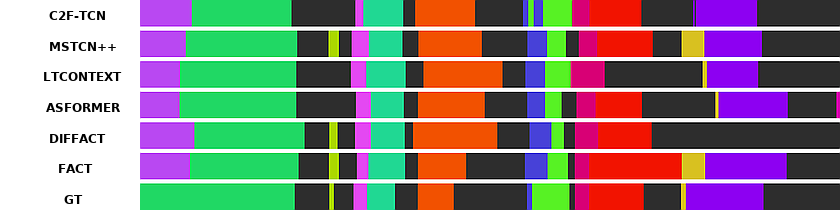}
    \caption{Qualitative results of temporal action segmentation on the egocentric subset of ENIGMA-360.}
    \label{fig:tas_ego}
\end{figure}

\begin{figure}[t]
    \centering
    \includegraphics[width=\linewidth]{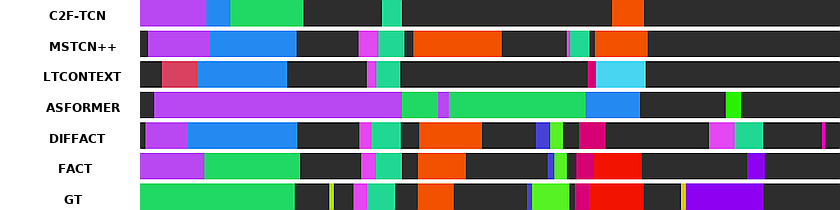}
    \caption{Qualitative results of temporal action segmentation on the exocentric subset of ENIGMA-360.}
    \label{fig:tas_exo}
\end{figure}

We trained each baseline independently on the egocentric and exocentric training subsets, and then evaluated each trained model on both test sets to assess cross-view generalization. Table~\ref{tab:tas} reports the results on both the egocentric and exocentric test sets of ENIGMA-360.  When the source and target domains are the same (Table~\ref{tab:tas_part1}), the FACT model achieves the best performance across all metrics, including $F1\{\beta\}$, Edit score, and accuracy. The only exception occurs in the EGO$\rightarrow$EGO setting, where FACT ranks second and third in accuracy, with scores of 70.8 and 72.7, respectively. In cross-domain scenarios (Table~\ref{tab:tas_part2}), the generalization capabilities of the baseline models become evident. In the EGO$\rightarrow$EXO setting, the C2F-TCN model consistently achieves the best and second-best results across all metrics. Conversely, in the EXO$\rightarrow$EGO setting, the ASFormer baseline outperforms others on all metrics except the Edit score, where it achieves the second-best result (22.8).
Notably, the DiffAct model demonstrates strong robustness across all experimental settings, frequently achieving second-best performance, highlighting its generalization capabilities. Qualitative results are reported in Figures~\ref{fig:tas_ego} and \ref{fig:tas_exo} where we report color-coded temporal segmentations of a sample video. Each row corresponds to a different baseline, with the ground truth provided in the last row. 

Despite following the same training protocol, all models show a substantial performance drop in cross-view evaluations. This indicates that egocentric and exocentric views, although temporally synchronized, exhibit strong appearance and motion discrepancies that are not trivial for current models to bridge.

\begin{table*}[t]
    \centering

    \begin{subtable}{\textwidth}
        \centering
        \resizebox{\linewidth}{!}{
        \begin{tabular}{|l|ccc|c|cc|ccc|c|cc|}
            \hline
            & \multicolumn{6}{c|}{\textbf{EGO$\rightarrow$EGO}} 
            & \multicolumn{6}{c|}{\textbf{EXO$\rightarrow$EXO}} \\
            \hhline{|~|------|------|}
            & \multicolumn{3}{c|}{F1@\{10,25,50\}} & Edit & Acc & AccB
            & \multicolumn{3}{c|}{F1@\{10,25,50\}} & Edit & Acc & AccB \\
            \hline
            C2F-TCN 
            & 66.9 & 59.9 & 44.0 & 65.0 & 58.2 & 69.2
            & 57.2 & 51.0 & 37.2 & 57.2 & 47.6 & 63.6 \\
            MSTCN++ 
            & 76.2 & 69.2 & 53.1 & 78.0 & 70.5 & 72.2
            & 65.0 & 56.0 & 37.4 & \underline{73.9} & 56.3 & 63.8 \\
            LTContext 
            & 74.3 & 70.4 & \underline{58.0} & 73.5 & 65.5 & 71.3
            & 64.7 & 59.5 & 45.7 & 62.7 & 48.5 & 60.0 \\
            ASFormer 
            & 76.0 & 69.7 & 55.6 & 78.0 & 70.0 & \textbf{73.6}
            & 58.3 & 54.3 & 38.8 & 58.8 & 55.7 & 62.8 \\
            DiffAct 
            & \underline{77.2} & \underline{72.7} & 57.1 & \underline{82.3} & \textbf{73.3} & \underline{73.0}
            & \underline{70.3} & \underline{64.8} & \underline{48.4} & 73.5 & \underline{59.7} & \underline{65.4} \\
            FACT 
            & \textbf{78.7} & \textbf{73.2} & \textbf{59.5} & \textbf{84.6} & \underline{70.8} & 72.7
            & \textbf{72.8} & \textbf{66.5} & \textbf{50.4} & \textbf{76.6} & \textbf{63.3} & \textbf{68.2} \\
            \hline
        \end{tabular}}
        \caption{EGO→EGO and EXO→EXO}
        \label{tab:tas_part1}
    \end{subtable}

    \vspace{0.5cm}

    \begin{subtable}{\textwidth}
        \centering
        \resizebox{\linewidth}{!}{
        \begin{tabular}{|l|ccc|c|cc|ccc|c|cc|}
            \hline
            & \multicolumn{6}{c|}{\textbf{EGO$\rightarrow$EXO}} 
            & \multicolumn{6}{c|}{\textbf{EXO$\rightarrow$EGO}} \\
            \hhline{|~|------|------|}
            & \multicolumn{3}{c|}{F1@\{10,25,50\}} & Edit & Acc & AccB
            & \multicolumn{3}{c|}{F1@\{10,25,50\}} & Edit & Acc & AccB \\
            \hline
            C2F-TCN 
            & \textbf{7.2} & \textbf{5.5} & \textbf{3.5} & \underline{13.1} & \textbf{5.6} & \underline{29.7}
            & 8.4 & 7.0 & \underline{4.8} & 19.0 & \underline{9.6} & 34.2 \\
            MSTCN++ 
            & 3.2 & 2.6 & 1.3 & 3.3 & 1.9 & 3.7
            & 3.8 & 2.6 & 2.1 & 4.6 & 7.0 & 17.7 \\
            LTContext 
            & 2.0 & 1.5 & 0.8 & 2.2 & 2.2 & 22.8
            & 2.8 & 1.5 & 1.1 & 3.2 & 1.0 & \underline{41.7} \\
            ASFormer 
            & 2.9 & 2.2 & 1.3 & 4.3 & 2.3 & \textbf{38.9}
            & \textbf{20.9} & \textbf{19.0} & \textbf{13.6} & \underline{22.8} & \textbf{22.7} & \textbf{49.0} \\
            DiffAct 
            & \underline{6.2} & \underline{5.1} & \underline{2.6} & \textbf{16.1} & \underline{4.1} & 23.6
            & \underline{10.31} & \underline{7.7} & 3.1 & \textbf{23.4} & 7.3 & 28.5 \\
            FACT 
            & 1.7 & 1.1 & 0.3 & 1.8 & 0.9 & 29.5
            & 9.4 & 6.4 & 2.9 & 15.9 & 6.3 & 35.0 \\
            \hline
        \end{tabular}}
        \caption{EGO→EXO and EXO→EGO}
        \label{tab:tas_part2}
    \end{subtable}

    \caption{Temporal Action Segmentation results on the ENIGMA-360 dataset. The best results are highlighted in bold, while the second-best results are underlined.}
    \label{tab:tas}
\end{table*}

\subsection{Keystep Recognition}
\label{sec:keysteps}

In industrial settings, the ability to identify key procedural steps is crucial for accurately modeling task structure, ensuring that mandatory operations are performed correctly and in the prescribed order, and supporting downstream systems in reasoning about task progression at a fine-grained level.
Keystep Recognition focuses on identifying key procedural steps within a video sequence. Specifically, the goal is to predict the keystep label $k_t$ for each temporal window $t$ in a given video $V$. Each keystep corresponds to a semantically meaningful and temporally localized action that is critical to the overall activity being performed. Unlike coarse action segmentation, keystep recognition emphasizes the detection of essential sub-events that define the structure of complex tasks, such as those found in ENIGMA-360 videos.

\subsubsection{Baselines and Evaluation Measures}
To benchmark the Keystep Recognition task, we evaluated four models: TimeSformer~\cite{gberta_2021_ICML}, GLEVR~\cite{romero2025keystep}, Video-Mined Task Graphs~\cite{ashutosh2024videomined}, and ProVideLLM~\cite{chatterjee2025memory}. For all the evaluated approaches, we adopted the default settings and architectures provided by the respective authors. We performed training and testing independently on both the egocentric (EGO) and exocentric (EXO) domains.
We evaluate our baselines using Precision, Recall, and F1-score. Following \cite{dvornik2021drop,dvornik2022flow} we also evaluate with Frame-wise Accuracy (Acc) and Intersection over Union (IoU). For each keystep \( k_i \), frame-wise Accuracy is calculated as the fraction of frames with ground truth \( k_i \) that are correctly classified. The overall Frame-wise Accuracy is the mean accuracy across all keysteps.

\subsubsection{Results}

The results in Table \ref{tab:timesformer_evaluation} highlight a clear performance gap between the egocentric and exocentric perspectives across all evaluated baselines. The egocentric view consistently outperforms the exocentric view across all evaluation metrics, suggesting that it provides richer and more detailed information about the user's actions.

Among the evaluated models, GLEVR emerges as the strongest baseline in the egocentric domain, achieving the highest Recall (0.82), F1-score (0.82), and Accuracy (0.94), while tying for the best IoU (0.74). Video-Mined Task Graphs also demonstrates highly competitive performance, securing the highest Precision (0.84) in the egocentric view and dominating the exocentric view with the best Precision (0.71), Recall (0.60), F1-score (0.63), and IoU (0.50). Conversely, ProVideLLM struggles significantly in our industrial setting, particularly in the exocentric domain, where it yields the lowest performance across all metrics.

\begin{figure}[ht]
    \centering
    \includegraphics[width=\linewidth]{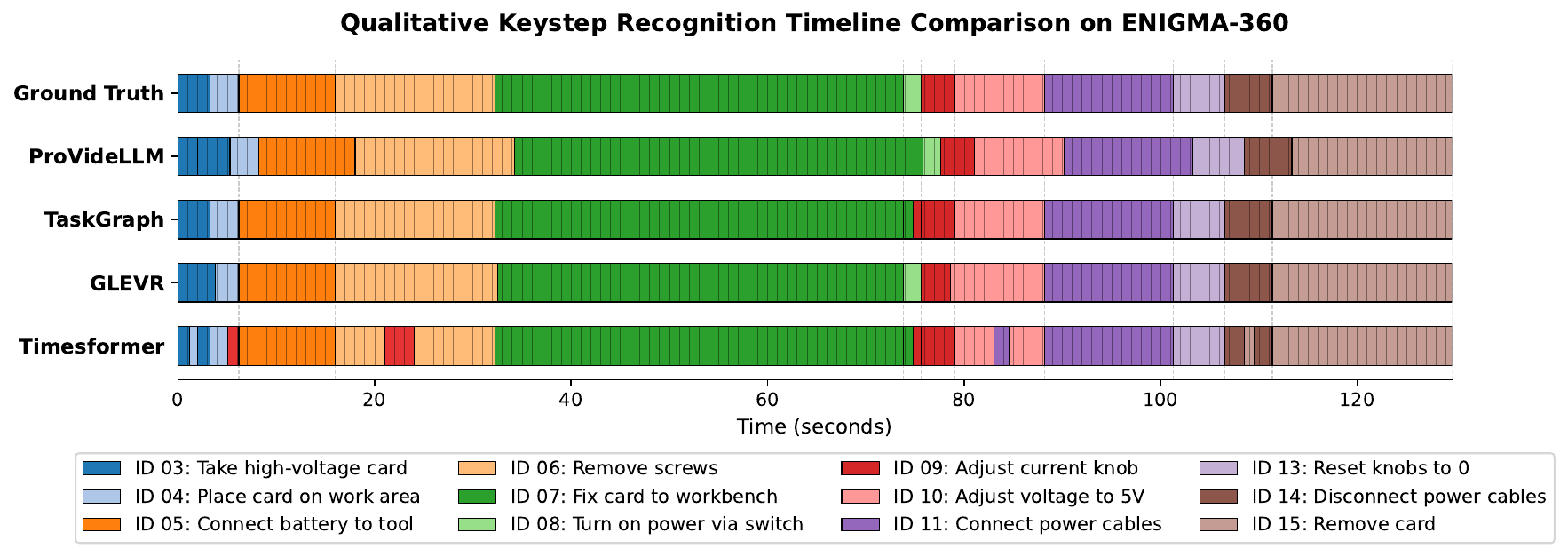}
    \caption{Qualitative results of keystep recognition baselines.}
    \label{fig:keystep_qual}
\end{figure}

Overall, the exocentric perspective likely suffers from occlusions of hand movements or manipulated objects, which limits its effectiveness in capturing fine-grained details. Many keysteps in ENIGMA-360 are defined by subtle hand-object manipulations that are inherently more visible from the egocentric viewpoint. In the exocentric videos, these cues often appear small, partially hidden, or visually ambiguous. As a result, all models struggle to reliably recognize each keystep from the third-person view, leading to the lower performance observed in the exocentric setting. 
Figure \ref{fig:keystep_qual} reports qualitative results for the evaluated keystep recognition baselines on a sample sequence. The timeline comparison effectively visualizes the temporal consistency of the predictions. Both GLEVR and Video-Mined Task Graphs (TaskGraph) produce smooth, continuous segments that closely align with the ground truth boundaries, accurately capturing long-duration steps such as \textit{fixing the card to the workbench (ID 07)}. ProVideLLM also maintains good temporal continuity, though with minor boundary shifts. In contrast, the TimeSformer baseline exhibits noticeable fragmentation and over-segmentation, frequently oscillating between classes. This visualizes the superior temporal modeling capabilities of the more recent architectures in managing complex, long-form procedural sequences.

\begin{table}[t]
    \centering
    \resizebox{\linewidth}{!}{
        \begin{tabular}{|l|ccccc|ccccc|}
            \hline
            \textbf{Approach} & \multicolumn{5}{c|}{\textbf{EGO}} & \multicolumn{5}{c|}{\textbf{EXO}} \\
            \hline
             & \textbf{Precision} & \textbf{Recall} & \textbf{F1} & \textbf{Acc} & \textbf{IoU} & \textbf{Precision} & \textbf{Recall} & \textbf{F1} & \textbf{Acc} & \textbf{IoU} \\
            \hline
            TimeSformer~\cite{gberta_2021_ICML}
            & 0.70 & 0.68 & 0.66 & 0.68 & 0.54
            & 0.59 & 0.57 & 0.57 & 0.58 & 0.43 \\

            GLEVR~\cite{romero2025keystep}
            & 0.83 & \textbf{0.82} & \textbf{0.82} & \textbf{0.94} & \textbf{0.74}
            & 0.62 & 0.58 & 0.58 & \textbf{0.78} & 0.45 \\

            Video-Mined Task Graphs~\cite{ashutosh2024videomined}
            & \textbf{0.84} & 0.60 & 0.63 & 0.88 & \textbf{0.74}
            & \textbf{0.71} & \textbf{0.60} & \textbf{0.63} & 0.63 & \textbf{0.50} \\

            ProVideLLM~\cite{chatterjee2025memory}
            & 0.68 & 0.66 & 0.66 & 0.77 & 0.56
            & 0.39 & 0.38 & 0.38 & 0.47 & 0.27 \\
            \hline
        \end{tabular}
    }
    \caption{Keystep Recognition results. The best results for each metric and modality are in bold.}
    \label{tab:timesformer_evaluation}
\end{table}

\subsection{Egocentric Hand-Object Interaction Detection}
\label{sec:ehoi}
We address the task of detecting Egocentric Hand-Object Interactions from egocentric RGB images as defined in \cite{Shan2020UnderstandingHH, VISOR2022}. For each input image, the goal is to predict the triplet \textit{hand, hand contact state, active object}. Further details on the task are provided in \cite{Shan2020UnderstandingHH, VISOR2022}.

\subsubsection{Baselines and Evaluation Measures}
We present results for three baselines~\cite{VISOR2022, leonardi2024synthetic, leonardi2024synthdata} that address this problem using different types of annotations: two baselines utilize segmentation masks~\cite{VISOR2022, leonardi2024synthetic}, while the third focuses exclusively on bounding boxes~\cite{leonardi2024synthdata}. 
\\ \textbf{Segmentation Masks:} We use the VISOR HOS method~\cite{VISOR2022} as one baseline for our experiments. This approach builds upon the PointRend~\cite{kirillov2020pointrend} instance segmentation network, further enhanced with three modules specifically designed to detect the hand side, the contact state (either contact or no contact), and an offset vector that links the hand to the object being interacted with. Additionally, we employ the method from~\cite{leonardi2024synthetic}, which adapts the VISOR HOS architecture within the Adaptive Teacher (AT) framework~\cite{li2022cross} for unsupervised domain adaptation and self-supervised learning. In this work, we disabled the domain adaptation module and focused solely on the self-supervised aspect, applying it to both a labeled and an unlabeled dataset, as detailed in Section~\ref{dataset}. 
\\ \textbf{Bounding Boxes:} For Hand-Object Interaction detection using bounding boxes, we adopt the method proposed in~\cite{leonardi2024synthdata} as our baseline. This approach is based on the method of~\cite{Shan2020UnderstandingHH}, which extends a two-stage object detector by incorporating additional modules that leverage hand features to predict the hand contact state, hand side, and an offset vector that indicates the object the hand is interacting with.

Following~\cite{VISOR2022}, we assess performance using the \textit{COCO Mask AP} metric~\cite{coco_dataset}. Specifically, we employ the Hand + Object (Overall) AP, which evaluates the accuracy of predicted hands and objects, the hand-state (contact vs. no contact), and the offset vector that describes the relationship between the hand and the active object. Additionally, we analyze performance through Mask AP metrics that focus on different aspects of the predictions: Hand (H), Hand + Side (H+S), Hand + Contact (H+C), and Object (O). For the baseline~\cite{leonardi2024synthdata}, we evaluate the performance of bounding box predictions accordingly.

\begin{table}[t]
	\centering
	\resizebox{\linewidth}{!}{
		\begin{tabular}{|lcccccc|}
			\hline
			\textbf{Approach}                                         & \textbf{Annotation Type} & \textbf{Overall} & \textbf{Hand}  & \textbf{Hand+Side} & \textbf{Hand+Contact} & \textbf{Object} \\
			\hline
			Baseline~\cite{VISOR2022}                                 & Mask                     & 63.84            & 85.01          & 81.05              & 52.32                 & 51.35           \\
			Baseline w Label Propagation~\cite{leonardi2024synthetic} & Mask                     & \textbf{64.55}   & \textbf{85.75} & \textbf{82.51}     & \textbf{52.59}        & \textbf{53.24}  \\
			\hline
			Baseline~\cite{leonardi2024synthdata}                     & Bounding box             & \textbf{46.24}   & \textbf{90.81} & \textbf{90.35}     & \textbf{46.51}        & \textbf{46.24}  \\
			\hline
		\end{tabular}}
    \caption{Egocentric Hand-Object Interaction Detection results. The best results for each group are in bold.}\label{tab:ehoi_results}
\end{table}

\subsubsection{Results}
The results in Table~\ref{tab:ehoi_results} show the performance of the three baselines for both segmentation masks and bounding boxes, with various metrics for the different aspects of hand-object interactions. 
\\ \textbf{Segmentation Masks:} The two baselines focusing on segmentation masks demonstrate similar performance. The VISOR HOS baseline~\cite{VISOR2022} achieves an Overall AP of 63.84\%, with high results in the Hand (85.01\%) and Hand+Side (81.05\%) metrics. However, the use of Label Propagation~\cite{leonardi2024synthetic} further enhances these results, increasing the Overall AP to 64.55\% and improving scores across all the other evaluated metrics. This suggests that incorporating self-supervised learning with label propagation can enhance prediction quality, especially in more complex tasks like Hand+Contact and Object recognition.
\\ \textbf{Bounding Boxes:} The baseline using bounding boxes~\cite{leonardi2024synthdata} shows different results compared to the segmentation-based methods. While it obtains high performance in the Hand and Hand+Side categories with APs of 90.81\% and 90.35\%, respectively, its Overall AP (46.24\%) is significantly lower than those obtained with segmentation masks. This disparity might be due to the fact that bounding boxes, although effective in localizing hands and objects, lack the fine-grained detail captured by segmentation masks, which is crucial for accurately evaluating complex tasks such as Hand+Contact classification and Object recognition.

\begin{figure}[ht!]
    \centering
\includegraphics[width=\linewidth]{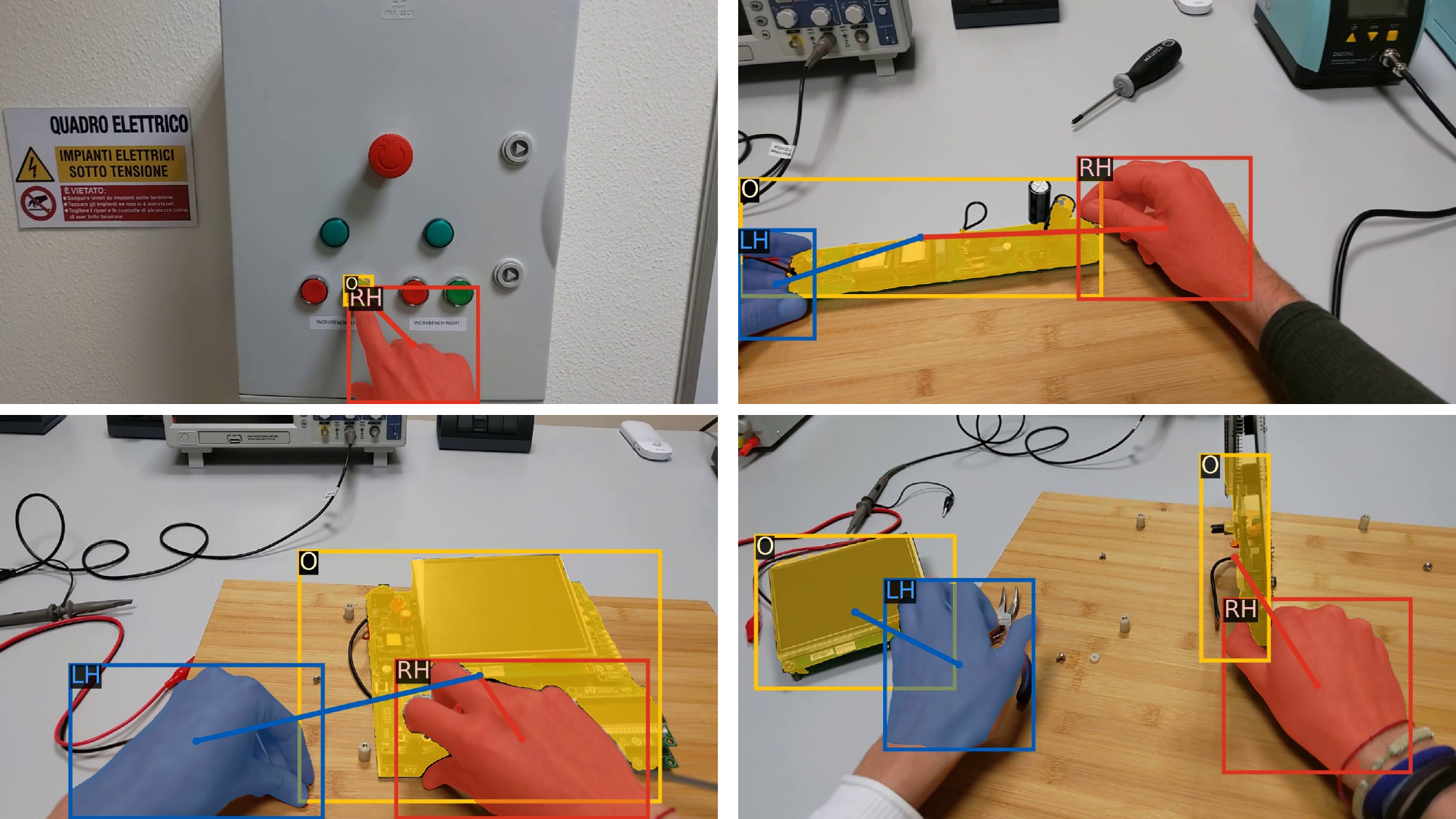}
    \caption{Qualitative results of the best hand-object interaction detection approach.}
    \label{fig:ehoi_qualitative}
\end{figure}

Qualitative results of the best approach are shown in Figure~\ref{fig:ehoi_qualitative}. The first row presents two correct predictions, where each hand is properly associated with the corresponding active object. The second row shows two failure cases: on the left, the approach incorrectly assigns an object to a hand that is not in contact; on the right, it fails to correctly identify the active object for each hand.

\subsection{Video-Grounded Cross-View VQA}
To address the need for video-grounded visual question answering (VQA) and multi-view alignment in assistive scenarios, we introduce a zero-shot cross-view retrieval benchmark formulated as a multiple-choice VQA task. Given an observation video clip from a query perspective (e.g., egocentric), a model must inspect four candidate clips captured from the opposite viewpoint (e.g., exocentric) and identify the temporally synchronized match depicting the same subject performing the same action instance at the same time.

\subsubsection{Baselines and Evaluation Measures}

We evaluate two state-of-the-art Vision-Language Models, Qwen3-VL-8B-Instruct and InternVL-3.5-8B, in a zero-shot setting without task-specific fine-tuning. For each instance, the query clip and four candidate option clips (A--D) are uniformly sampled to 16 frames and presented jointly in a single conversational turn. The model is asked a video-grounded multiple-choice question to select the correct synchronized candidate label (A--D) using greedy decoding.
To rigorously assess whether models perform genuine spatiotemporal visual reasoning or exploit superficial shortcuts, we construct a 3-tier benchmark evaluated across two query directions ($ego \rightarrow exo$ and $exo \rightarrow ego$):
\begin{itemize}
    \item \textbf{Easy (Different Action, Different Subject):} Distractors differ in both action type and subject identity from the query clip.
    \item \textbf{Medium (Same Action, Different Subject):} Distractors depict the same action type performed by different subjects, forcing the model to rely on subject appearance and fine-grained execution style.
    \item \textbf{Hard (Same Action, Same Subject):} Distractors show the \textit{exact same subject} performing the \textit{exact same action} at different, non-synchronized temporal instances. This setting eliminates all semantic shortcuts and forces true temporal synchronization reasoning.
\end{itemize}
To prevent duration from acting as a trivial shortcut, all distractors are temporally cropped to match the exact duration of the query clip. In total, the benchmark comprises 3,572 unique query instances evaluated in both query directions, yielding 7,144 evaluation runs across 64 action categories (60 for the Hard tier due to strict occurrence constraints). We report performance using top-1 retrieval Accuracy (\%).

\begin{table}[ht!]
	\centering
	\resizebox{\linewidth}{!}{
		\begin{tabular}{|llcccc|}
			\hline
			\textbf{Model} & \textbf{Query Direction} & \textbf{Easy} & \textbf{Medium} & \textbf{Hard} & \textbf{Overall} \\
			\hline
			Qwen3-VL-8B & Ego $\rightarrow$ Exo & \textbf{53.67\%} & \textbf{41.56\%} & \textbf{28.56\%} & \textbf{41.26\%} \\
			& Exo $\rightarrow$ Ego & 32.11\% & 32.27\% & 24.51\% & 29.30\% \\
			& \textit{Direction Gap} & \textit{21.56\%} & \textit{9.29\%} & \textit{4.05\%} & \textit{11.96\%} \\
			\hline
			InternVL-3.5-8B & Ego $\rightarrow$ Exo & 28.01\% & 25.04\% & 27.03\% & 26.67\% \\
			& Exo $\rightarrow$ Ego & 25.96\% & 23.94\% & 23.02\% & 24.41\% \\
			& \textit{Direction Gap} & \textit{2.05\%} & \textit{1.10\%} & \textit{4.01\%} & \textit{2.26\%} \\
			\hline
		\end{tabular}}
	\caption{Zero-shot Video-Grounded Cross-View VQA accuracy (\%) using Qwen3-VL-8B-Instruct and InternVL-3.5-8B across distractor difficulty tiers and query directions. Random chance is 25.0\%.}\label{tab:vqa_results}
\end{table}

\subsubsection{Results}
Table~\ref{tab:vqa_results} summarizes the performance across all six experimental settings for both Qwen3-VL-8B and InternVL-3.5-8B. Overall, Qwen's accuracy ranges from 53.67\% to 28.56\%, highlighting significant performance drops as distractors become progressively more challenging. In contrast, InternVL-3.5-8B struggles across all tiers, with its accuracy remaining tightly constrained between 23.02\% and 28.01\%, barely exceeding the random chance level of 25.0\%.

In the egocentric-to-exocentric ($ego \rightarrow exo$) direction, Qwen follows a clear downward trend: accuracy drops from 53.67\% in the \textbf{Easy} tier to 41.56\% in \textbf{Medium}, and collapses close to chance level (28.56\%) in \textbf{Hard}. Comparing the decrease between tiers reveals that removing subject identity shortcuts ($\text{Medium} \rightarrow \text{Hard}$) causes a significantly larger performance loss for Qwen than removing action identity shortcuts ($\text{Easy} \rightarrow \text{Medium}$). Conversely, InternVL-3.5-8B does not exhibit a monotonic decrease with distractor difficulty; instead, the differences between difficulty levels remain within three percentage points (28.01\% in \textbf{Easy}, 25.04\% in \textbf{Medium}, 27.03\% in \textbf{Hard}). This indicates that InternVL is unable to leverage the semantic shortcuts that inflate Qwen's scores in the easier settings.

Evaluating the reverse direction ($exo \rightarrow ego$) demonstrates a strong directional asymmetry for Qwen in the \textbf{Easy} setting, where exocentric queries yield only 32.11\% accuracy compared to 53.67\% for egocentric queries. This highlights that first-person views provide richer motion and visual grounding cues for cross-view matching than wider third-person observations. However, as distractor difficulty increases, Qwen's viewpoint gap progressively narrows (9.29 points in \textbf{Medium}) and consistently reduces in \textbf{Hard} (28.56\% vs 24.51\%). InternVL lacks this asymmetry entirely: its exocentric-to-egocentric accuracy peaks at 25.96\% in \textbf{Easy} and bottoms out at 23.02\% in \textbf{Hard}. Compared with Qwen, which still retained a measurable advantage in the easier settings, InternVL appears considerably less capable of leveraging either viewpoint-specific information or semantic cues for retrieval, performing at near-chance levels regardless of the query direction.

\begin{figure}[ht!]
    \centering
    \includegraphics[width=0.98\linewidth]{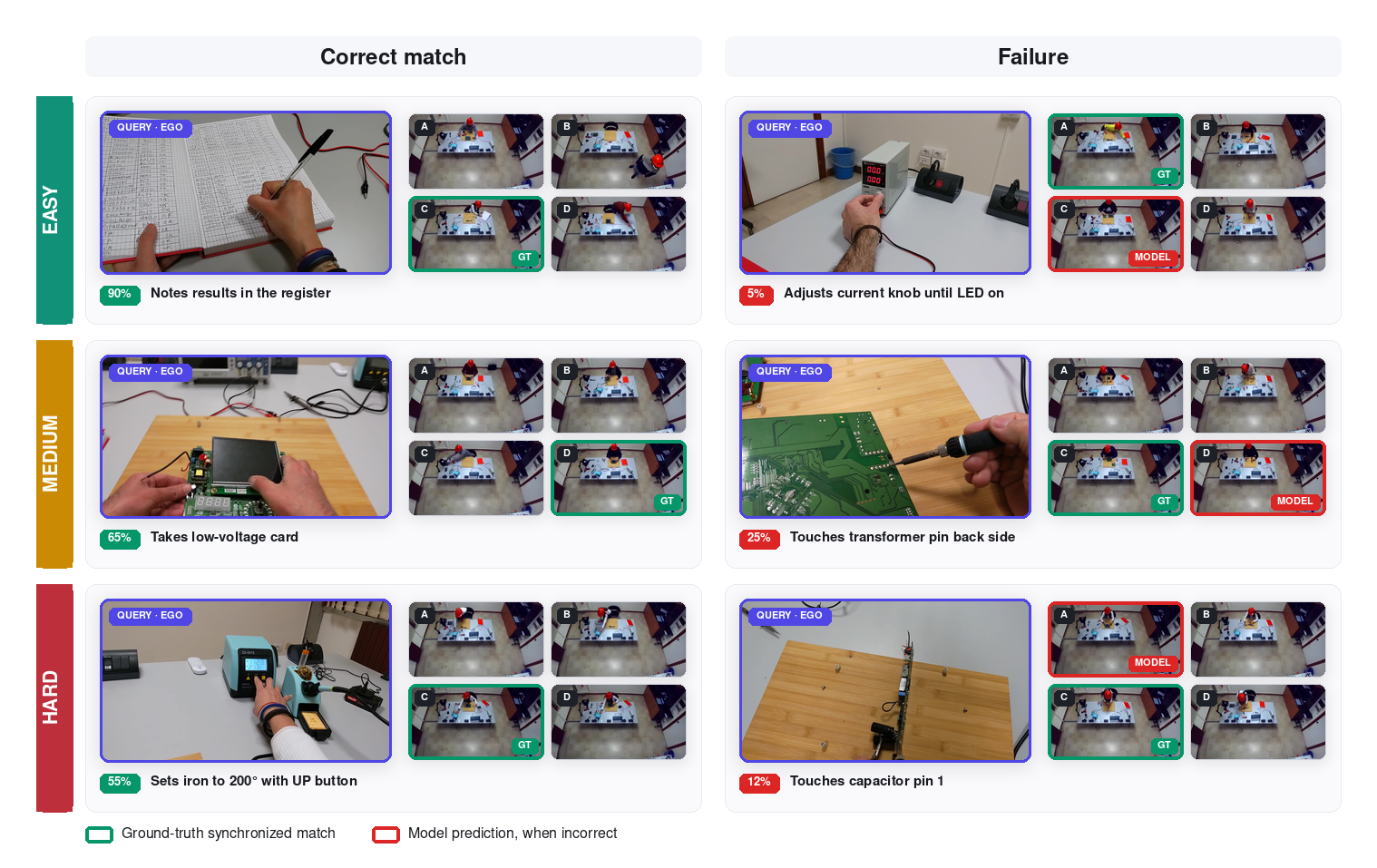}
 \caption{Qualitative examples $(ego\rightarrow exo)$ of the zero-shot Video-Grounded Cross-View VQA task across different difficulty tiers (Easy, Medium, Hard). The green border highlights the ground-truth synchronized match (GT), while the red border indicates an incorrect VLM prediction. In the presented failure cases, the model struggles due to: a lack of view-invariant visual anchors (Easy tier); a loss of discriminative pixel-level signals across views during close-contact pin manipulations (Medium tier); and the inability to distinguish between temporally distinct but visually repetitive instances of the identical fine-grained action (Hard tier).}
    \label{fig:ego_vqa_qualitative}
\end{figure}

\begin{figure}[ht!]
    \centering
    \includegraphics[width=0.98\linewidth]{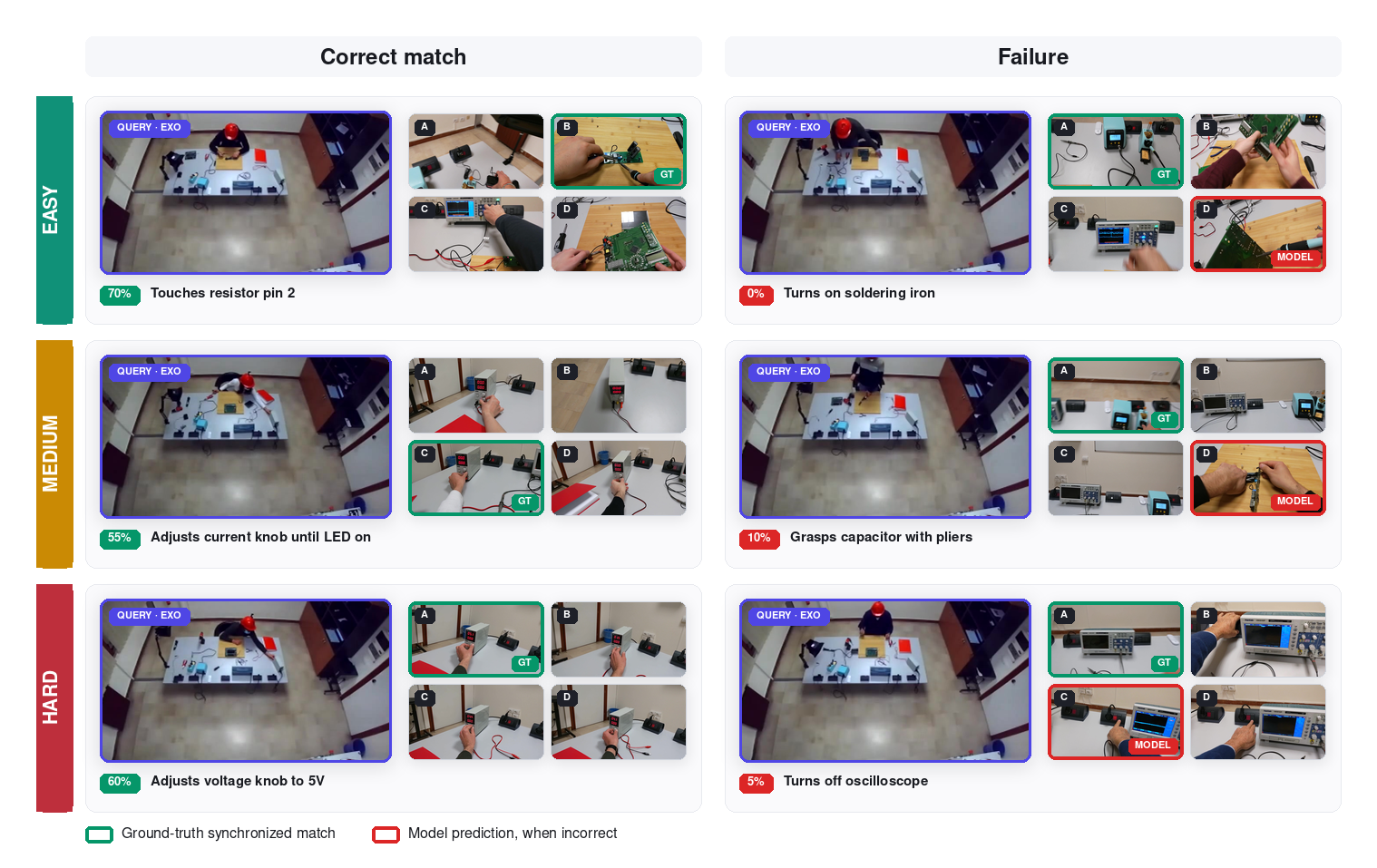}
    \caption{Qualitative examples $(exo\rightarrow ego)$ of the zero-shot Video-Grounded Cross-View VQA task across different difficulty tiers (Easy, Medium, Hard). The green border highlights the ground-truth synchronized match (GT), while the red border indicates an incorrect VLM prediction. In the presented failure cases, the model struggles due to: a lack of distinctive visual anchors across viewpoints (Easy tier); confusion in resolving the specific manipulated object or tool (Medium tier); and a spatial proximity bias, where the model hallucinates an interaction with a salient nearby instrument rather than identifying the true temporal match (Hard tier).}
    \label{fig:exo_vqa_qualitative}
\end{figure}

Given its superior performance across all experimental settings, we focus our qualitative analysis exclusively on Qwen3-VL-8B.  Figures~\ref{fig:ego_vqa_qualitative} and~\ref{fig:exo_vqa_qualitative} provide qualitative support for these trends, contrasting a correctly retrieved instance (\textit{Correct match}) with a failure case (\textit{Failure}) for each difficulty tier, in the $ego \rightarrow exo$ and $exo \rightarrow ego$ directions, respectively. The examples reveal a shift in the underlying matching strategy across tiers. In \textbf{Easy}, correct matches are anchored to actions with a distinctive, view-invariant visual signature (e.g., the handwritten register), while failures involve queries that carry little discriminative information in either view (e.g., \textit{``Adjusts current knob until LED on''}), so the model can succeed via simple scene-level correspondence but fails when no such anchor exists. In \textbf{Medium}, correct matches still rely on recognizable tool-object configurations, whereas failures concentrate on close-contact, pin-level manipulations (e.g., \textit{``Touches transformer pin back side''}), where the discriminative signal is confined to a few pixels and is easily lost across viewpoints. In \textbf{Hard}, correct matches appear largely incidental rather than systematic, even the best-performing instance is visually near-identical across the subject's repeated occurrences of that action, while failures are dominated by repetitive, fine-grained actions with no reliable appearance-based cue, confirming that genuine cross-view temporal synchronization remains largely out of reach for current VLMs.

\section{Discussion and Conclusion}
In this work, we presented ENIGMA-360, a comprehensive multi-view dataset acquired in a real industrial setting, designed to support advanced research on human behavior understanding. Alongside establishing rigorous benchmarks for Temporal Action Segmentation, Keystep Recognition, and Egocentric Human-Object Interaction Detection, we introduced a novel Video-Grounded Cross-View VQA task to evaluate temporal alignment across viewpoints.

Beyond establishing these benchmarks, our extensive evaluations yield critical insights. First, our baseline analysis highlights a persistent, asymmetric domain gap between viewpoints. While recent state-of-the-art architectures achieve competitive performance on egocentric streams by exploiting localized hand-tool interactions, generalizing these capabilities to exocentric views remains a severe challenge, heavily bottlenecked by occlusions and the loss of fine-grained visual cues. More crucially, our novel cross-view temporal alignment task exposes a fundamental limitation in current Vision-Language Models. The sharp degradation in performance demonstrates that today's models largely rely on coarse appearance matching rather than possessing genuine, video-grounded temporal reasoning capabilities. Ultimately, these findings suggest that scaling up generic multi-view data is insufficient; future progress necessitates specialized architectures capable of genuine multimodal fusion and fine-grained temporal synchronization.

To address these open challenges, ENIGMA-360 enables a broad range of new research directions. Thanks to its combination of temporally and spatially rich annotations, synchronized ego-exo videos, precomputed additional pseudo-labels and features, and the full 3D models of the laboratory and objects, these resources open avenues for investigating fine-grained procedural reasoning, multiview representation learning, advanced human-object interaction modeling, and 3D-aware synthetic-to-real approaches for human behavior understanding in industrial domains.

We acknowledge that ENIGMA-360 was collected in a single industrial laboratory, characterized by a fixed layout, controlled lighting conditions, and a predefined set of tools and components. Moreover, the dataset focuses on two specific maintenance procedures, performed by a limited group of participants to ensure annotation quality and procedural consistency. While these factors constrain environmental variability and behavioral diversity, which may limit generalization to other facilities featuring entirely different spatial configurations or workflows, they represent a necessary trade-off to provide strictly synchronized and densely annotated streams. Nevertheless, we believe that ENIGMA-360 will serve as a highly valuable and challenging resource to drive the next generation of multi-view AI in realistic industrial environments.

\section*{\uppercase{Acknowledgements}}
\small
This research is supported by Next Vision\footnote{Next Vision: https://www.nextvisionlab.it/} s.r.l., by MISE - PON I\&C 2014-2020 - Progetto ENIGMA  - CUP: B61B19000520008, and by the project FAIR – PNRR MUR Cod. PE0000013 - CUP: E63C22001940006.

\newpage
\appendix

\section{ENIGMA Interactions Labeling Guide}
\label{sec:appendix_labeling_guide}

\subsection{Introduction to the Labeling Activity}
The purpose of this activity is the temporal and spatial annotation of interactions between the user and specific objects, as well as between the objects themselves, starting from videos acquired via wearable devices. We define an interaction between the user and an object as the process of the user manipulating the object. In particular, each annotation will consist of a ``time of contact or release'' and one or more ``active objects''. The time of contact indicates the exact temporal instant in which an interaction occurs between the human and an object or between two objects. When the interaction occurs, the object involved in such interaction changes its state and becomes ``active''.

The goal of the activity is therefore to label all human-object and object-object interactions, providing for each of them the contact time, the type of interaction, the spatial position of the active objects, and the class of these objects. 

\begin{figure}[ht!]
    \centering
    \includegraphics[width=0.98\linewidth]{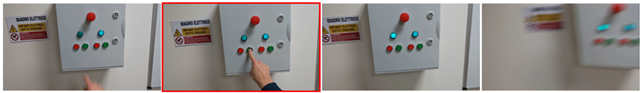}
    \caption{Example of human-object interaction.}
    \label{fig:labeling_guide_fig1}
\end{figure}

The videos to be labeled were acquired using Microsoft HoloLens 2, a wearable device capable of recording videos from the user's point of view. At the beginning of each video, the user performs a ``synchronization'' operation, which consists of turning on a lamp located on the workbench. This allows temporal synchronization between the video data stream acquired by the HoloLens and external sensors, such as another camera recording the video stream from a different point of view.

\begin{figure}[H]
    \centering
    \includegraphics[width=0.98\linewidth]{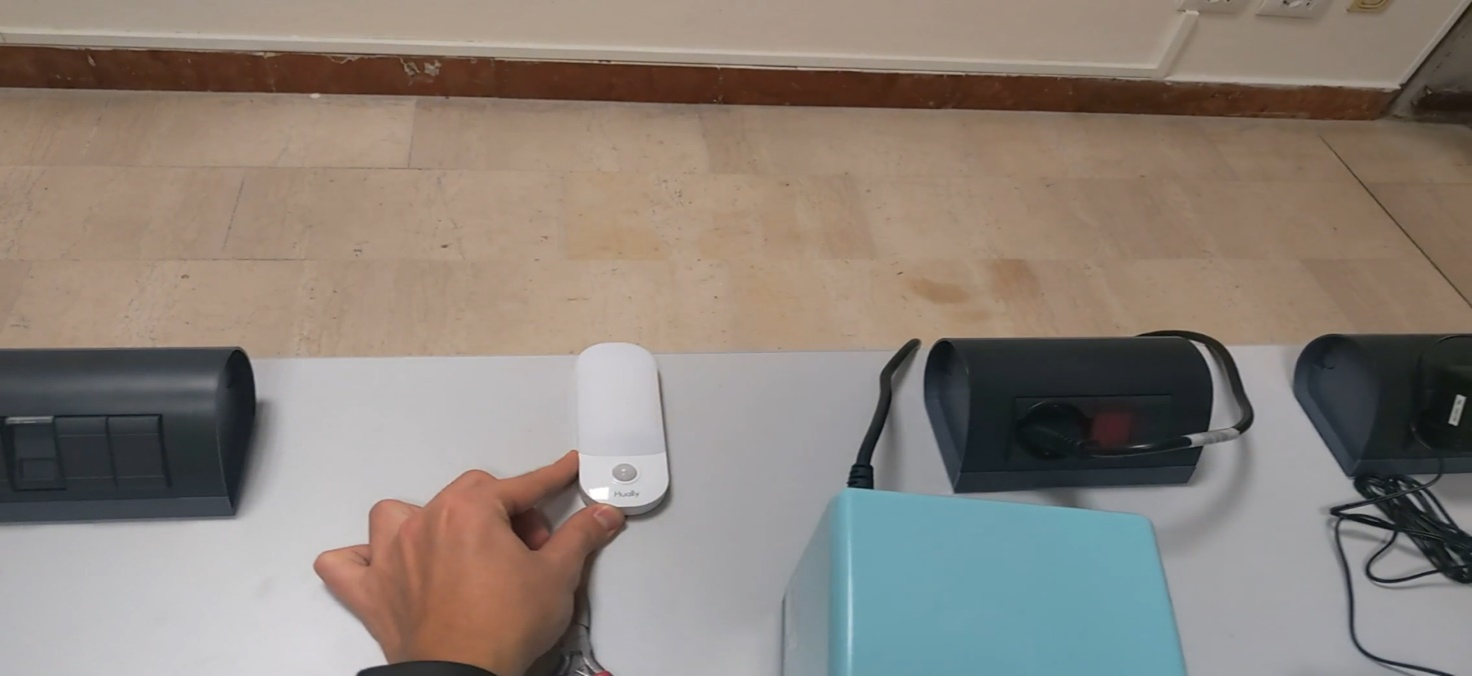}
    \caption{Sychronization operation thourgh the lamp.}
    \label{fig:labeling_guide_fig2}
\end{figure}

\subsection{Human-object interaction}
In every acquisition session, the user will interact with objects and the surrounding environment. Human-object interaction identifies an interaction occurring between the human and one or more objects in the environment. Since the video is acquired in first-person, the human will be represented by their hands; therefore, any interaction involving the operator's hand(s) and one or more objects will belong to this first macro-class.

The human-object interactions to be labeled are of four types:
\begin{itemize}
    \item \textbf{Hand\_Contact-Take};
    \item \textbf{Hand\_Decontact-Release}.
\end{itemize}

The \textbf{Hand\_Take} interaction identifies all those interactions where the user picks up an object from the workbench.

\begin{figure}[H]
    \centering
    \includegraphics[width=0.8\linewidth]{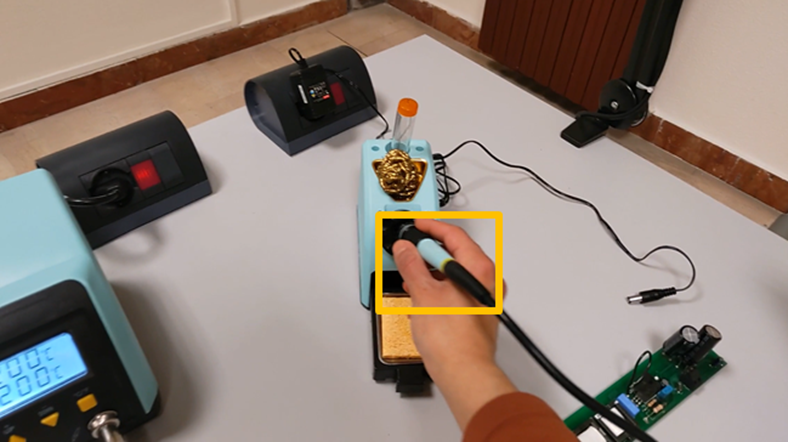}
    \caption{Example of Hand\_Take interaction.}
    \label{fig:labeling_guide_fig3}
\end{figure}

The \textbf{Hand\_Release} interaction identifies all those interactions in which the contact between the user's hand and the held object is interrupted (e.g., after placing an object on the workbench).

\begin{figure}[H]
    \centering
    \includegraphics[width=0.8\linewidth]{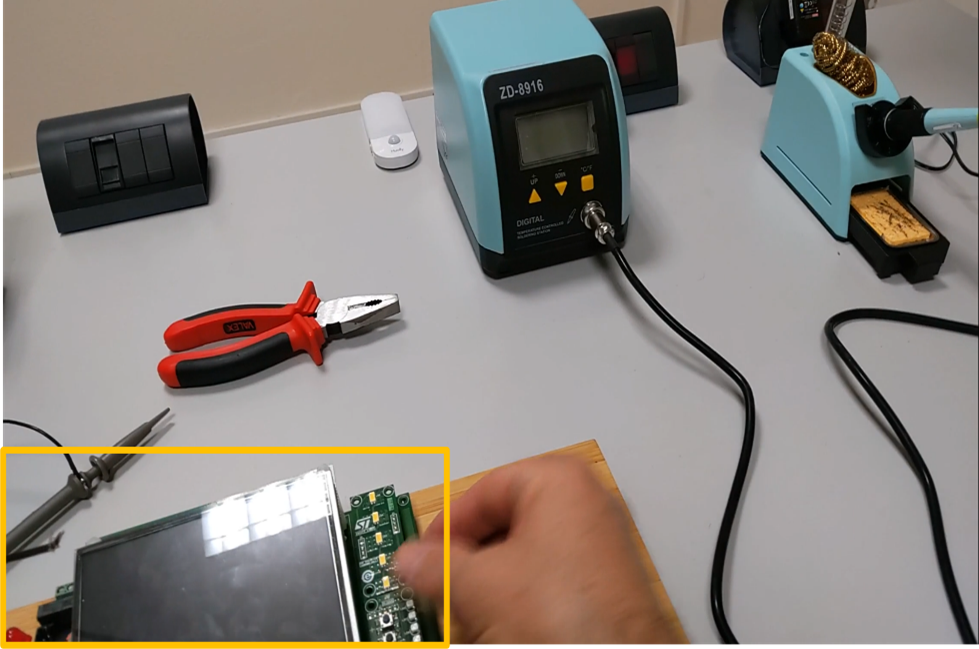}
    \caption{Example of Hand\_Release interaction.}
    \label{fig:labeling_guide_fig4}
\end{figure}

The \textbf{Hand\_Contact} interaction identifies all those interactions in which the user touches an object without picking it up or putting it down (e.g., touching the board, touching a button).

\begin{figure}[H]
    \centering
    \includegraphics[width=0.8\linewidth]{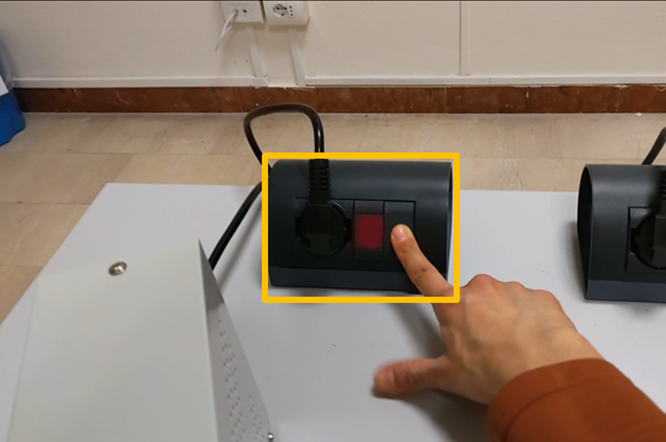}
    \caption{Example of Hand\_Contact interaction.}
    \label{fig:labeling_guide_fig5}
\end{figure}

The \textbf{Hand\_Decontact} interaction identifies all those interactions in which the user interrupts contact with an object that was not held in the hand.

\begin{figure}[H]
    \centering
    \includegraphics[width=0.8\linewidth]{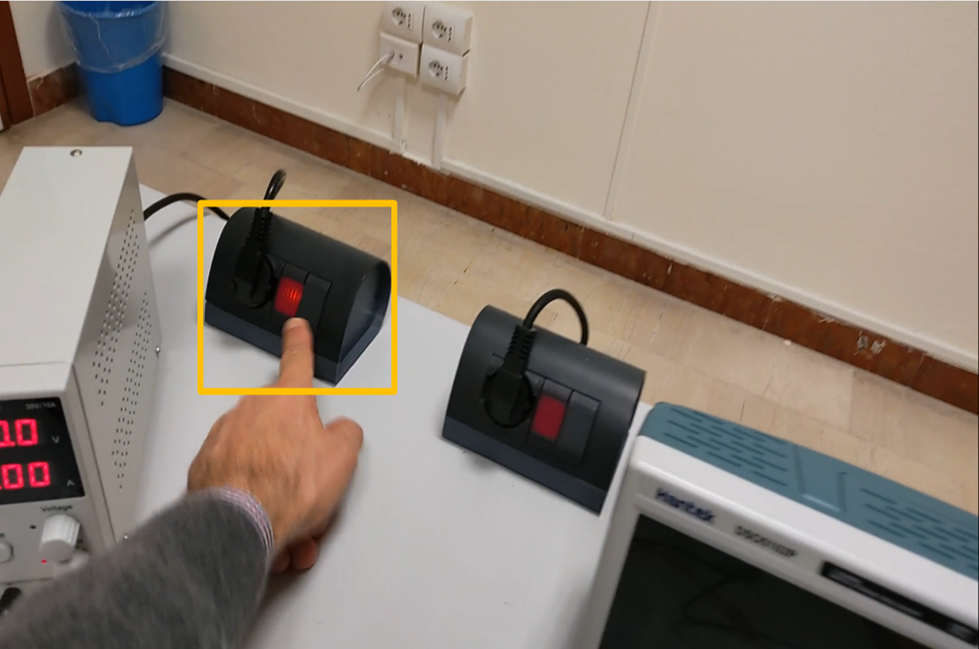}
    \caption{Example of Hand\_Decontact interaction.}
    \label{fig:labeling_guide_fig6}
\end{figure}

In order to effectively understand the type of interaction performed by the user, it is necessary to carefully check the video sequence, therefore examining the frames preceding and following the contact frame\footnote{}.

\subsection{Object Taxonomy}
The environment will feature objects fixed to the walls or movable objects placed on the workbench with which the user can interact. The complete list of objects is reported in Table \ref{tab:object_taxonomy}.

\begin{table}[h]
\centering
\resizebox{\textwidth}{!}{
\begin{tabular}{|l|l|l|l|}
\hline
\textbf{Object ID} & \textbf{Object Name} & \textbf{Type} & \textbf{Description} \\ \hline
B01 & Power Supply & Instrumentation & Used to generate electrical current \\ \hline
B01\_1 & Power Supply Cables & Tool & Used to power the electronic boards \\ \hline
B02 & Oscilloscope & Instrumentation & Used for testing the boards \\ \hline
B02\_P & Oscilloscope Probe & Tool & Used for testing the boards \\ \hline
B02\_C & Ground Clip & Tool & Used for testing the boards \\ \hline
B03 & Soldering Station & Instrumentation & Used to solder components onto the boards \\ \hline
B03\_B & Soldering Base & Instrumentation & Used to solder components onto the boards \\ \hline
B03\_P & Soldering Iron Tip & Tool & Used to solder components onto the boards \\ \hline
B04 & Electric Screwdriver & Tool & To fix the board to a support on the table using screws \\ \hline
B04\_B & Electric Screwdriver Battery & Instrumentation & To power the electric screwdriver \\ \hline
B04\_C & Electric Screwdriver Battery Connector & Instrumentation & To recharge the electric screwdriver's battery \\ \hline
B05 & Screwdriver & Tool & To fix the board to the table or a support using screws \\ \hline
B06 & Pliers & Tool & Used during the soldering phase \\ \hline
B07 & High Voltage Board & Device & Object for testing or repair \\ \hline
B08 & Low Voltage Board & Device & Object for testing or repair \\ \hline
B08\_S & Low Voltage Board Screen & Device & Part of the low voltage board \\ \hline
B09 & Register & Instrumentation & To record the results of tests and repairs \\ \hline
B10 & Workbench Area & Structural Element & Workbench area on which to perform operations \\ \hline
F01 & Panel A & Structural Element & Used to turn on and off the switches on the workbench \\ \hline
I\_A\_L\_R & Left Red Switch & Structural Element & Used to turn off the switches (left side) \\ \hline
I\_A\_L\_V & Left Green Switch & Structural Element & Used to turn on the switches (left side) \\ \hline
I\_A\_R\_R & Right Red Switch & Structural Element & Used to turn off the switches (right side) \\ \hline
I\_A\_R\_V & Right Green Switch & Structural Element & Used to turn on the switches (right side) \\ \hline
I\_F04\_L\_B01 & Tower 1 Left & Structural Element & Switch for the power supply socket tower \\ \hline
I\_F04\_L\_B02 & Tower 2 Left & Structural Element & Switch for the oscilloscope socket tower \\ \hline
I\_F04\_R\_B03 & Tower 1 Right & Structural Element & Switch for the soldering iron socket tower \\ \hline
I\_F04\_R\_B04 & Tower 2 Right & Structural Element & Switch for the electric screwdriver socket tower \\ \hline
\end{tabular}
}
\caption{Taxonomy of the objects present in the ENIGMA laboratory.}
\label{tab:object_taxonomy}
\end{table}

\subsection{Labeling Software}
VGG Image Annotator (VIA) will be used as the tool for the annotation process. VIA allows manual annotation of images, audio, and video. It is web-based, stand-alone software that can be used offline via a web browser.

\subsubsection{Downloading and Using the software}
To download VIA, go to its web page and download \url{https://www.robots.ox.ac.uk/~vgg/software/via/}.

\begin{figure}[H]
    \centering
    \includegraphics[width=0.8\linewidth]{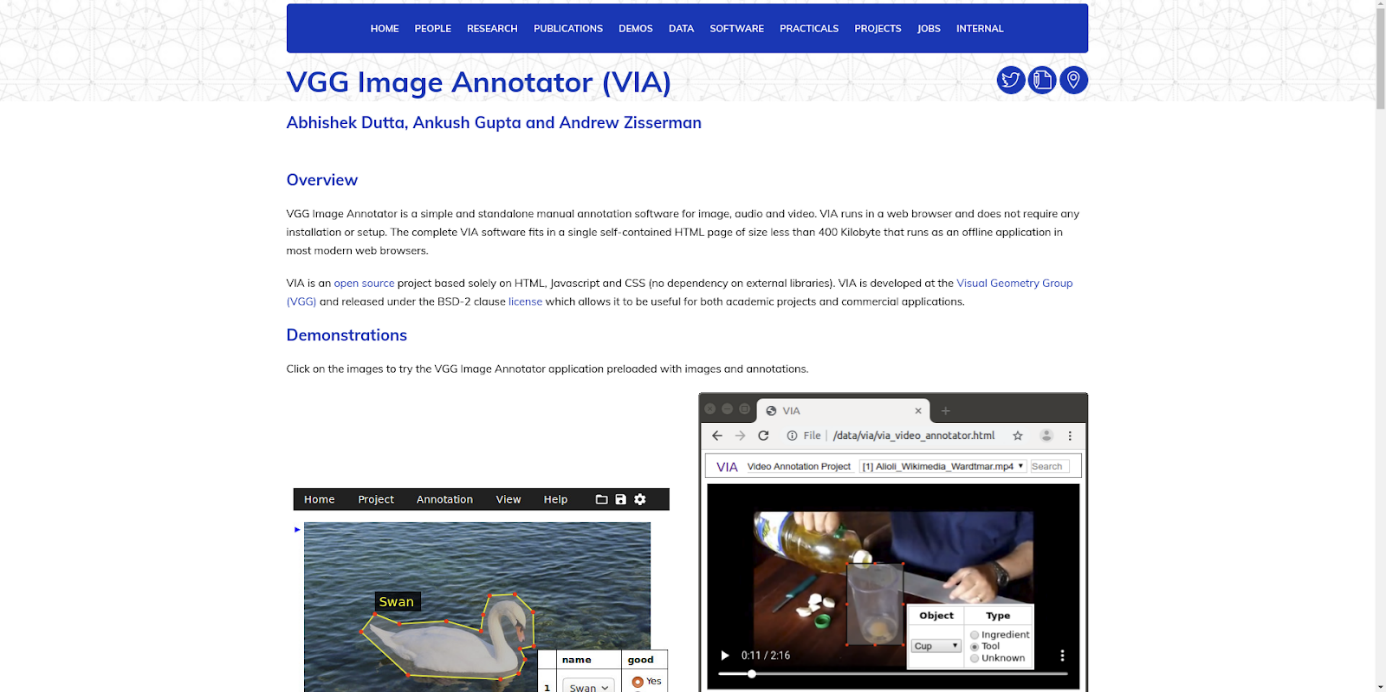}
    \caption{Home page of the labeling software.}
    \label{fig:labeling_guide_fig11}
\end{figure}

On the home page, you can find the version 3 download section, as shown in the figure.

\begin{figure}[H]
    \centering
    \includegraphics[width=0.8\linewidth]{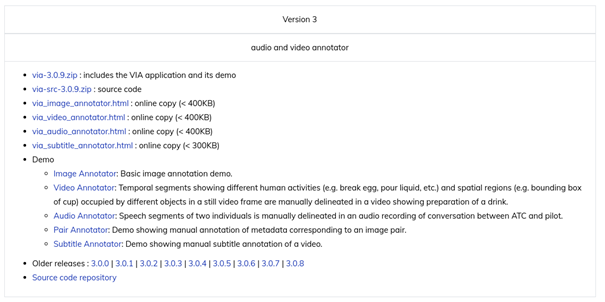}
    \caption{Download Section.}
    \label{fig:labeling_guide_fig12}
\end{figure}

Simply download "via\_video\_annotator.html". Once the downloaded file is opened, the interface shown in the figure will be displayed.

\subsection{Using the Software}
Open the software and load the \texttt{IPLAB\_VIA.json} project, which saves the annotator from having to set up the software manually. To load the JSON file containing the standard project, you need to click the button indicated in the figure.

\begin{figure}[H]
    \centering
    \includegraphics[width=0.8\linewidth]{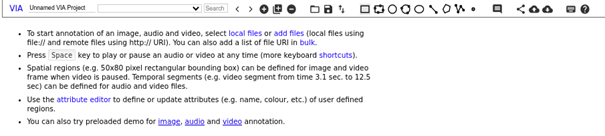}
    \caption{Button to click to load the project.}
    \label{fig:labeling_guide_fig13}
\end{figure}

To play or pause the video playback, use the spacebar. When paused, you can move frame by frame using the ``l'' (backward) and ``r'' (forward) keys. 

\begin{figure}[H]
    \centering
    \includegraphics[width=0.8\linewidth]{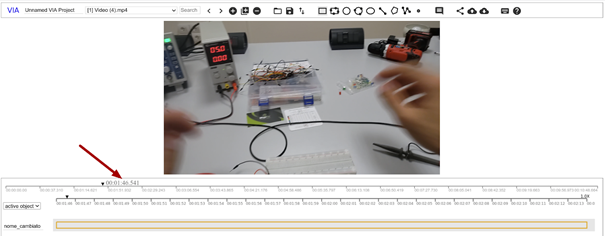}
    \caption{Timeline to move frames.}
    \label{fig:labeling_guide_fig16}
\end{figure}

Once an interaction has been identified, navigate to the frame to be labeled and then place the bounding box over the active object, as shown in the figure.

\begin{figure}[H]
    \centering
    \includegraphics[width=0.8\linewidth]{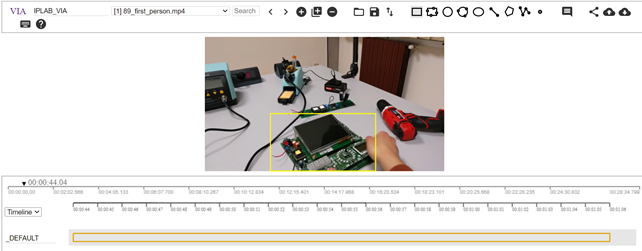}
    \caption{Annotation of the active object using a bounding box.}
    \label{fig:labeling_guide_fig17}
\end{figure}

After annotating the active object with the bounding box, clicking inside the bounding box will bring up a window where you can select attributes, as shown in the following figure.

\begin{figure}[H]
    \centering
    \includegraphics[width=0.8\linewidth]{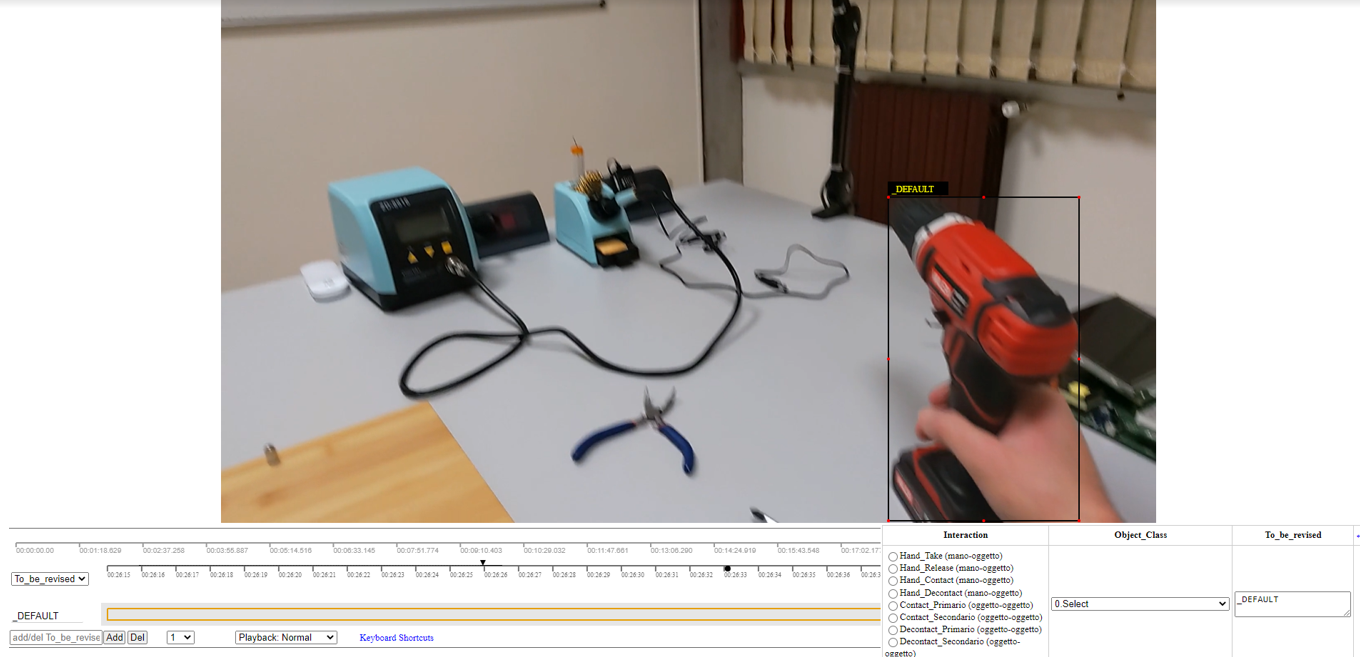}
    \caption{Selection of active object attributes.}
    \label{fig:labeling_guide_fig18}
\end{figure}

There are two types of attributes to select: Interaction and Object Class.

\begin{itemize}
    \item The Interaction attribute is used to specify the type of interaction taking place; for further details on interaction types. You can only select one type of interaction from those available.
    \item The Object Class attribute is used to define the class of the active object. The class of the active object must be chosen from the Object Class list. For further details on the various classes that can be associated with the active object, refer to the chapter on the Taxonomy of Objects.
\end{itemize}

If you are unsure about the type of interaction to associate with a specific frame, or if you have any other doubts, you should navigate to the frame where the interaction begins and press the "a" key. This will create a temporal segment on the timeline, which will help pinpoint the exact point in the video where the labeler has an indecision or doubt. The following figure shows an example of inserting a temporal segment (after pressing the "a" key) on the timeline.

\begin{figure}[H]
    \centering
    \includegraphics[width=0.8\linewidth]{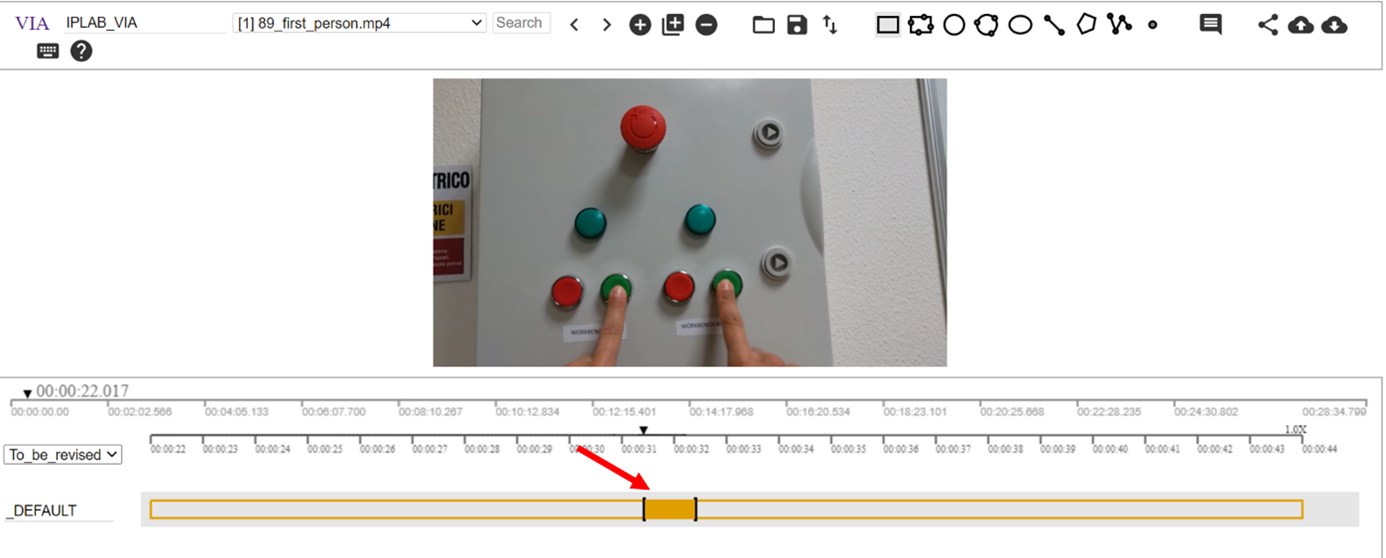}
    \caption{Insertion of a temporal segment due to an annotator's indecision. The segment is indicated by the red arrow.}
    \label{fig:labeling_guide_fig19}
\end{figure}

After inserting the temporal segment, you will still need to place the bounding box on the object you are unsure about, and in the attributes window, you can write your doubt in the dedicated "To\_be\_revised" text box, as shown in the figure.

\begin{figure}[H]
    \centering
    \includegraphics[width=0.8\linewidth]{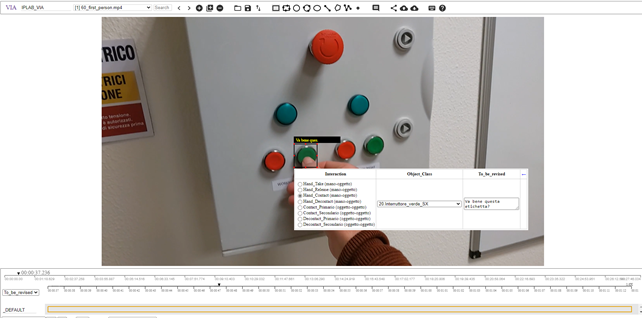}
    \caption{Inserting a comment in the dedicated "To\_be\_revised" box.}
    \label{fig:labeling_guide_fig20}
\end{figure}

Once the labeling phase is concluded, click the button indicated in the following figure to save the annotations. The program will save a JSON file. It is recommended to save the project frequently so as not to lose any changes made.
Thanks to this .json file, you will be able to resume the labeling phase at a later time. Therefore, instead of loading the IPLAB\_VIA.json file when the annotation tool is opened, you will need to load the latest saved .json file.

\begin{figure}[H]
    \centering
    \includegraphics[width=0.8\linewidth]{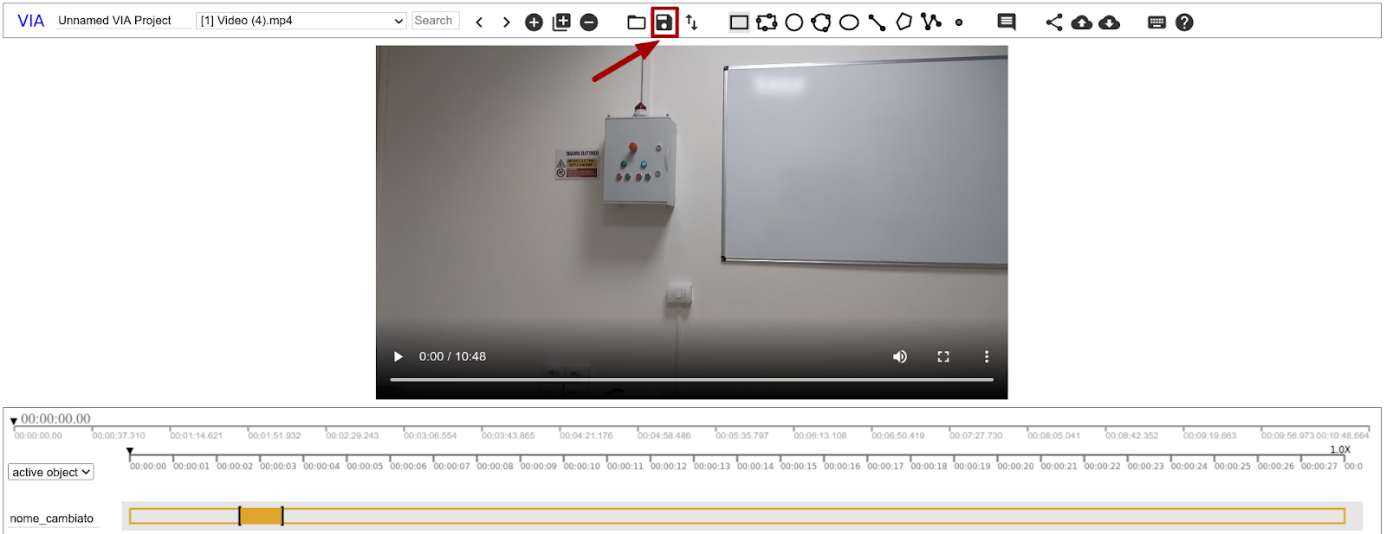}
    \caption{Saving annotations.}
    \label{fig:labeling_guide_fig21}
\end{figure}

\subsection{Folder Structure}
Each annotator will have a set of videos to label; for each video, there will be a folder containing the corresponding video. All \texttt{.json} and \texttt{.csv} files related to a specific video must be saved inside that corresponding folder and renamed with the exact name of the video.

\subsection{Guidelines and Special Cases}
Below are the guidelines for correctly labeling interactions, along with several sections on special cases that will be encountered during the labeling process.

\subsubsection{Guidelines}
If an interaction with an object occurs before the start of the synchronization action, it should not be labeled. The figure shows an example of the synchronization moment.

\begin{figure}[H]
    \centering
    \includegraphics[width=0.5\linewidth]{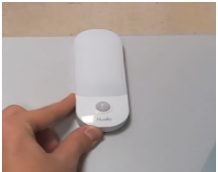}
    \caption{Synchronization operation.}
    \label{fig:labeling_guide_fig26}
\end{figure}

The moment when an object is touched, picked up, or put down (human-object interaction) is labeled. In particular, only interactions visible in the scene are labeled.

The following figure shows an example of a contact frame in a human-object interaction. Specifically, the interaction shown is of the Hand\_Contact-Take type. In the specific case of interaction with Panel A, both the Hand\_Contact-Take interaction with the button on Panel A and the Hand\_Contact-Take interaction with the entire Panel A must be labeled.

\begin{figure}[H]
    \centering
    \includegraphics[width=0.5\linewidth]{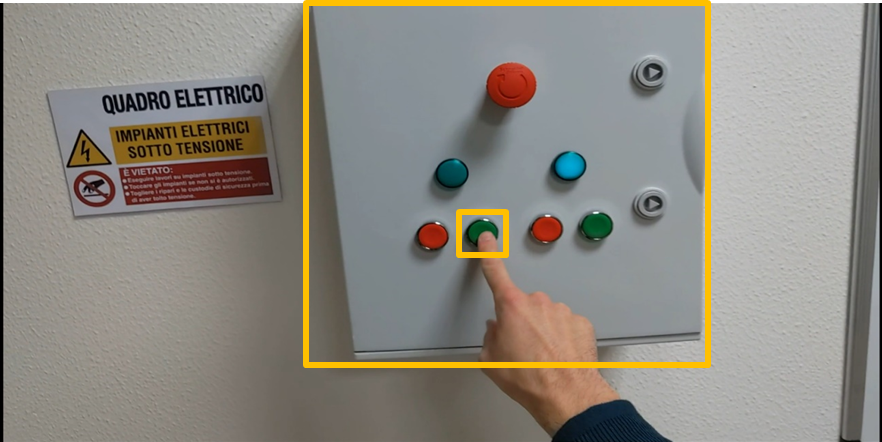}
    \caption{Example of a human-object Hand\_Contact (hand-object) interaction contact frame.}
    \label{fig:labeling_guide_fig27}
\end{figure}

An extension of a board is considered part of the board. For example, in the following figure, the cable connected to the board is considered an extension of the board, and therefore a human-object interaction with the board is taking place, specifically a Hand\_Contact-Take interaction.

\begin{figure}[H]
    \centering
    \includegraphics[width=0.8\linewidth]{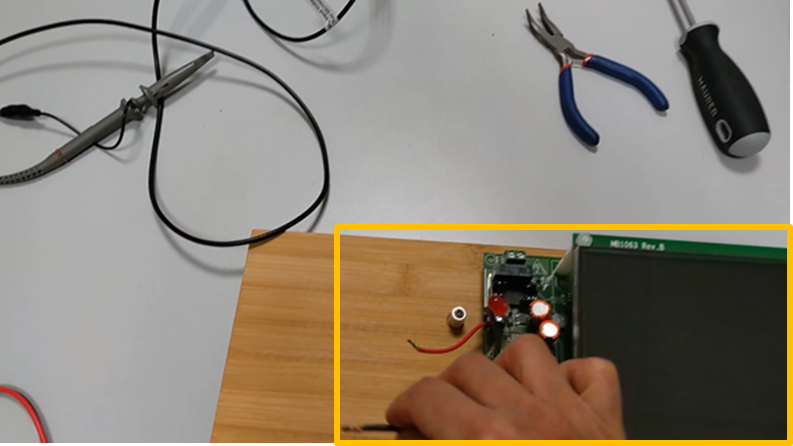}
    \caption{Example of a human-object interaction with a board.}
    \label{fig:labeling_guide_fig30}
\end{figure}

\paragraph{Lifting objects without detaching them from the workbench}
When the hand lifts the object without actually detaching it from the table, this human-object interaction should be labeled as a contact-take.
For example, sometimes the board is placed vertically on its side. The moment it is lifted to be positioned in that way, the take interaction must be labeled.

\begin{figure}[H]
    \centering
    \includegraphics[width=0.8\linewidth]{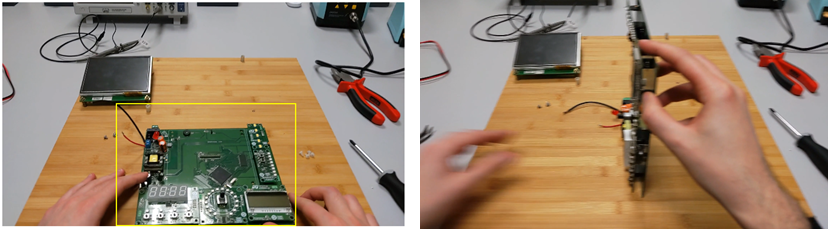}
    \caption{Example of lifting the board without detaching it from the workbench.}
    \label{fig:labeling_guide_fig32}
\end{figure}

\paragraph{"Take" interaction with hand already in contact}
A take does not always occur through a new contact between the hand and the object. In some cases, the user has already been in contact with the object in question for some time, and then lifts it without having any other interactions. Also in this case, the take must be labeled in the frame where the beginning of the movement to lift the object is noticed.
In the following example, it can be seen how the hand is already in contact with the high voltage board but only lifts it subsequently. In this case, the take must be labeled in the frame where the lifting is noticed.

\begin{figure}[H]
    \centering
    \includegraphics[width=0.8\linewidth]{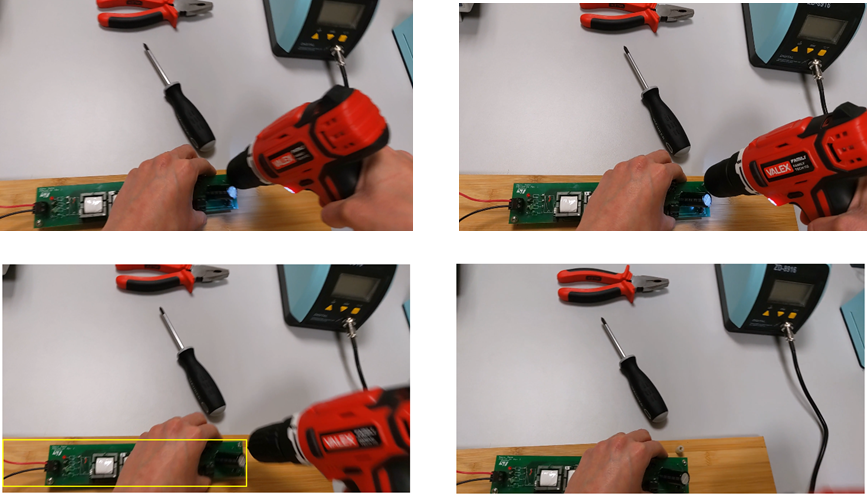}
    \caption{Example of a take with the hand already in contact with the object.}
    \label{fig:labeling_guide_fig33}
\end{figure}

\subsubsection{Special Cases regarding Object Contact}
If an object is touched simultaneously with both hands, the interaction is labeled only once. If the touches occur at two different times, even if they are temporally very close, two labels must be added for the interactions performed with the two hands. For example, in the following image sequence, it can be seen that the touch with the left hand and the oscilloscope occurs simultaneously with the insertion of the probe connector, so a single human-object interaction with the oscilloscope will be labeled, specifically Hand\_Contact-Take.

\begin{figure}[H]
    \centering
    \includegraphics[width=0.8\linewidth]{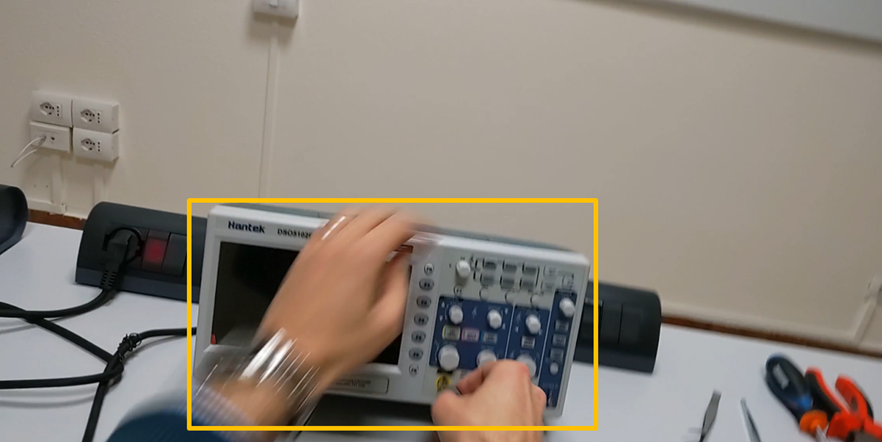}
    \caption{Example of an interaction where the hands simultaneously touch the Oscilloscope object.}
    \label{fig:labeling_guide_fig34}
\end{figure}

If one hand touches an object that the other hand was already operating on, the object is still labeled. The following figure shows that although the user already has their left hand on the board, the interaction with the right hand is still annotated, specifically as a Hand\_Contact-Take with the low voltage board.

\begin{figure}[H]
    \centering
    \includegraphics[width=0.8\linewidth]{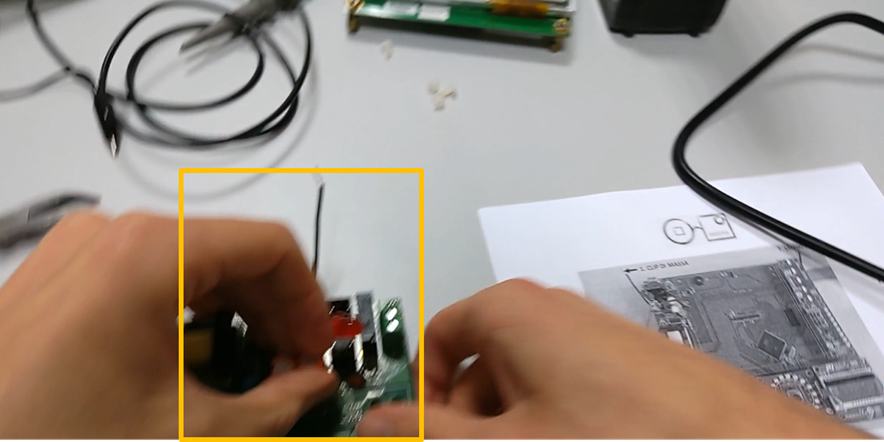}
    \caption{Example of an interaction where one hand touches an object already in contact with the operator's other hand.}
    \label{fig:labeling_guide_fig36}
\end{figure}

\subsubsection{Special Cases regarding Interaction Continuity}
Consecutive interactions on the same object are not labeled; it is sufficient to annotate only the initial interaction. For example, the following figure shows two consecutive frames. Since the interactions between the hand and the oscilloscope are consecutive and the continuity of the interaction is not interrupted (e.g., by moving the hand away, looking elsewhere, or touching another object), only the first frame where the interaction begins will be labeled, while subsequent operations will be ignored. In this specific case, there will be a Hand\_Contact-Take interaction with the oscilloscope.

\begin{figure}[H]
    \centering
    \includegraphics[width=0.8\linewidth]{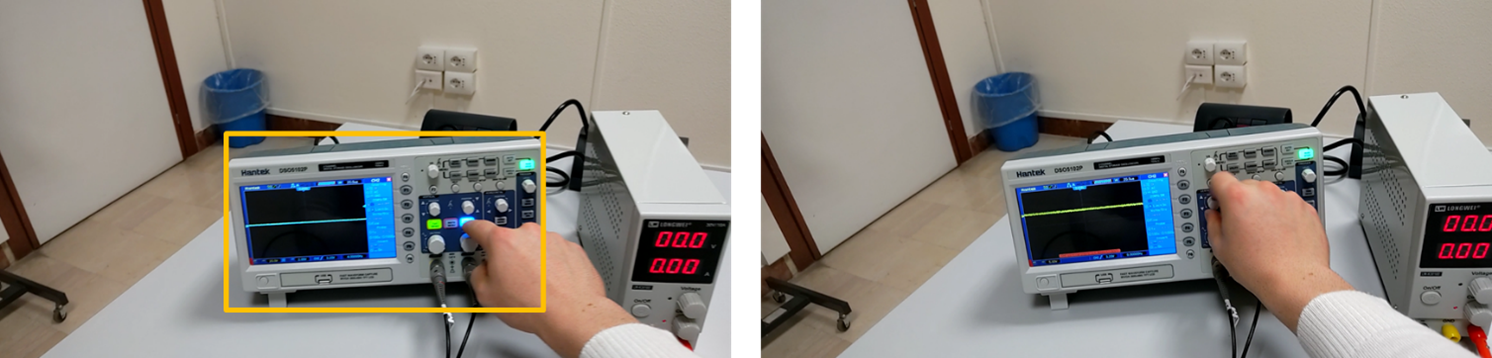}
    \caption{Example of consecutive interactions with the same object, where only the initial frame should be labeled.}
    \label{fig:labeling_guide_fig38}
\end{figure}

Furthermore, if you were continuously operating on an object and an event occurs that breaks the continuity, the next object touched must be labeled (again).
Passing an object from one hand to the other is not labeled. In the figure below, you can observe the passing of an "oscilloscope probe" from the left hand to the right hand.

\begin{figure}[H]
    \centering
    \includegraphics[width=0.8\linewidth]{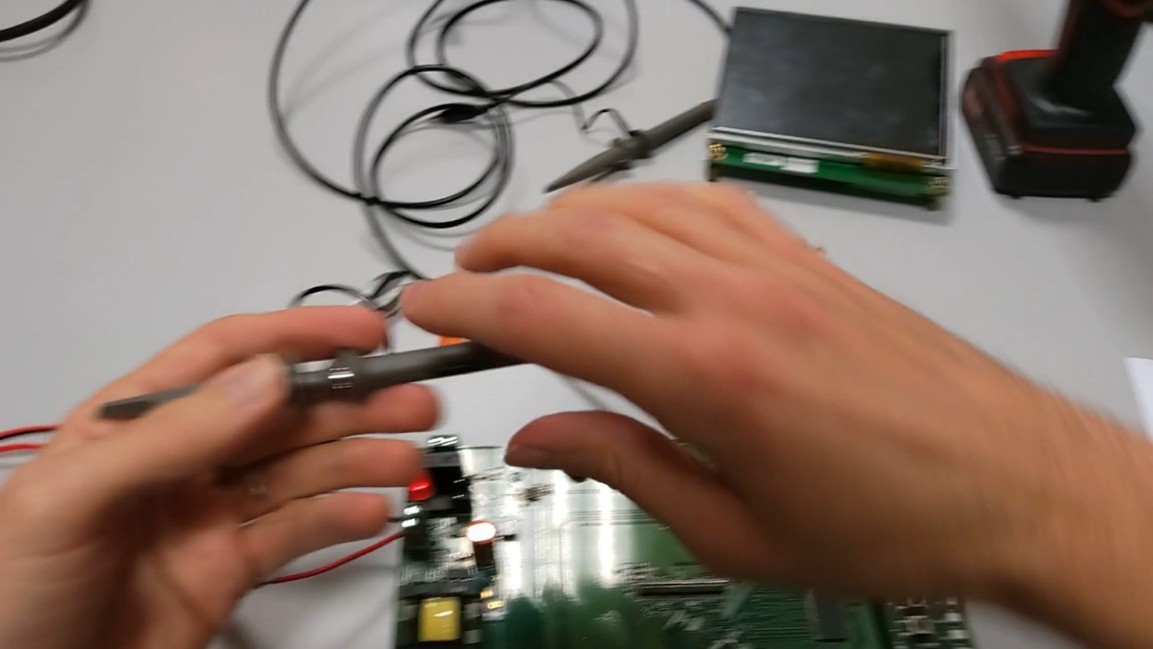}
    \caption{Example of passing an object from one hand to another, which should not be labeled.}
    \label{fig:labeling_guide_fig39}
\end{figure}

As long as an object remains in hand, it is not labeled again, even if the object goes out of frame.
For example, in the sequence below, the user picks up the soldering iron tip with their hands (to be labeled as a Hand\_Contact-Take interaction with the "soldering iron tip"), then shifts their gaze until the tip disappears from view. The moment the user looks back at the tip, it should not be relabeled because it has remained in hand.

\begin{figure}[H]
    \centering
    \includegraphics[width=0.8\linewidth]{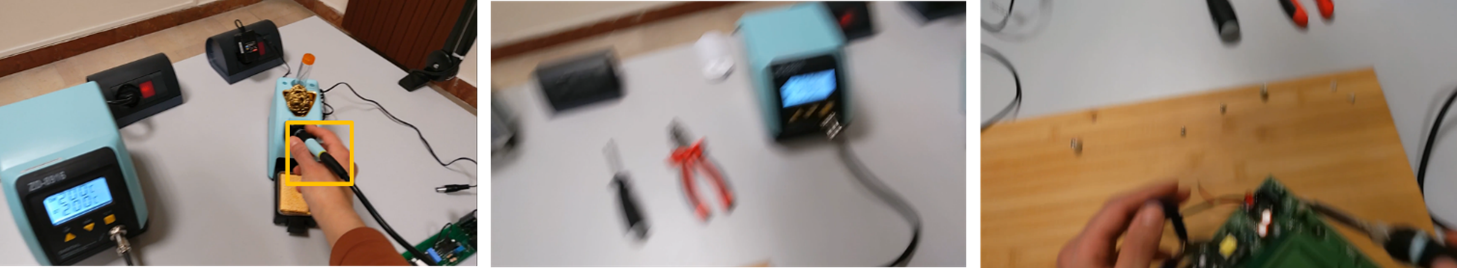}
    \caption{Example of what not to label due to a change in framing.}
    \label{fig:labeling_guide_fig40}
\end{figure}

\subsubsection{Special Cases regarding Cables on the Workbench}
If contact occurs with the cables present on the workbench (e.g., oscilloscope probe cable, power supply clamp cables, etc.), the interaction should not be labeled. The following figure shows a hand touching the oscilloscope probe cable.

\begin{figure}[H]
    \centering
    \includegraphics[width=0.8\linewidth]{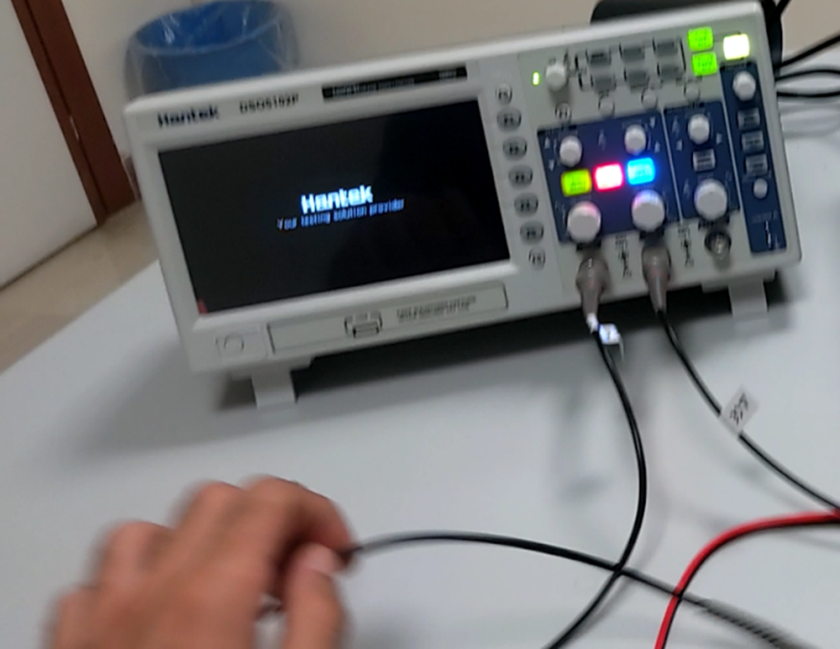}
    \caption{Example of a touch with the oscilloscope probe cable.}
    \label{fig:labeling_guide_fig41}
\end{figure}

\paragraph{``Take'' Human-Object Interaction when Cables are not on the Workbench}
When the user grabs only one of the two clips belonging to the power supply cables, but since they are connected by cables, the other one is also lifted, if the other clip is grabbed while it is in mid-air, the interaction is still considered a Take.

\paragraph{Hand\_Contact-Take on the Power Supply Cables}
In some cases, it is possible to see the user grab the power supply cables, which are already attached to the board's jumper wires, to bring the latter closer to another primary object to initiate an interaction between objects. In these cases, it is considered a Hand\_Contact-Take.
In the following image, you can notice how the power supply cable is grabbed to bring the board's wires closer to the oscilloscope probe.

\begin{figure}[H]
    \centering
    \includegraphics[width=0.8\linewidth]{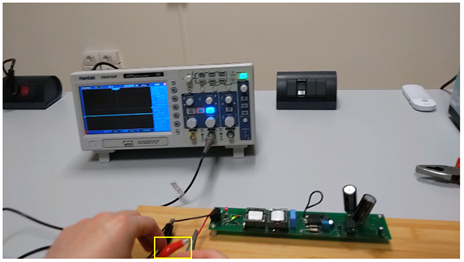}
    \caption{Hand\_Contact-Take interaction with the power supply cables.}
    \label{fig:labeling_guide_fig42}
\end{figure}

\subsubsection{Special Cases regarding the Low Voltage Board}
The low voltage board also includes a display that is removed from the board itself during certain procedures. When the board has the screen mounted and the hand touches the board and not the display, there is an interaction only with the board. For example, in the following figure, the user is touching the board, resulting in a human-object interaction with the low voltage board.

\begin{figure}[H]
    \centering
    \includegraphics[width=0.8\linewidth]{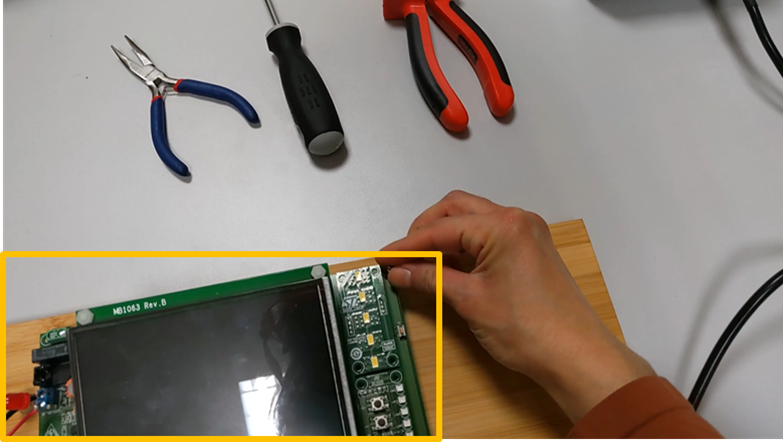}
    \caption{Example of a human-object interaction with the low voltage board.}
    \label{fig:labeling_guide_fig43}
\end{figure}

In the adjacent figure, there is an interaction with the low voltage board while the screen is absent because it has already been removed; thus, a Hand\_Contact-Take interaction with the low voltage board will be labeled. The same reasoning applies to Decontacts.

\begin{figure}[H]
    \centering
    \includegraphics[width=0.8\linewidth]{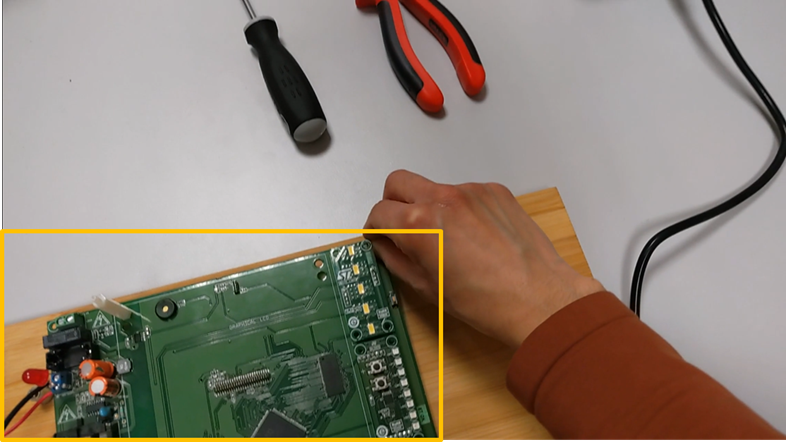}
    \caption{Example of a human-object interaction with the low voltage board without the screen.}
    \label{fig:labeling_guide_fig44}
\end{figure}

When interacting exclusively with the board's screen and it is disassembled, the interactions will be only with the screen. The following figure shows an interaction with the low voltage board's screen. Specifically, a Hand\_Decontact-Release interaction with the board's screen will be labeled. In this case, it applies the moment contact with the screen is broken.

\begin{figure}[H]
    \centering
    \includegraphics[width=0.8\linewidth]{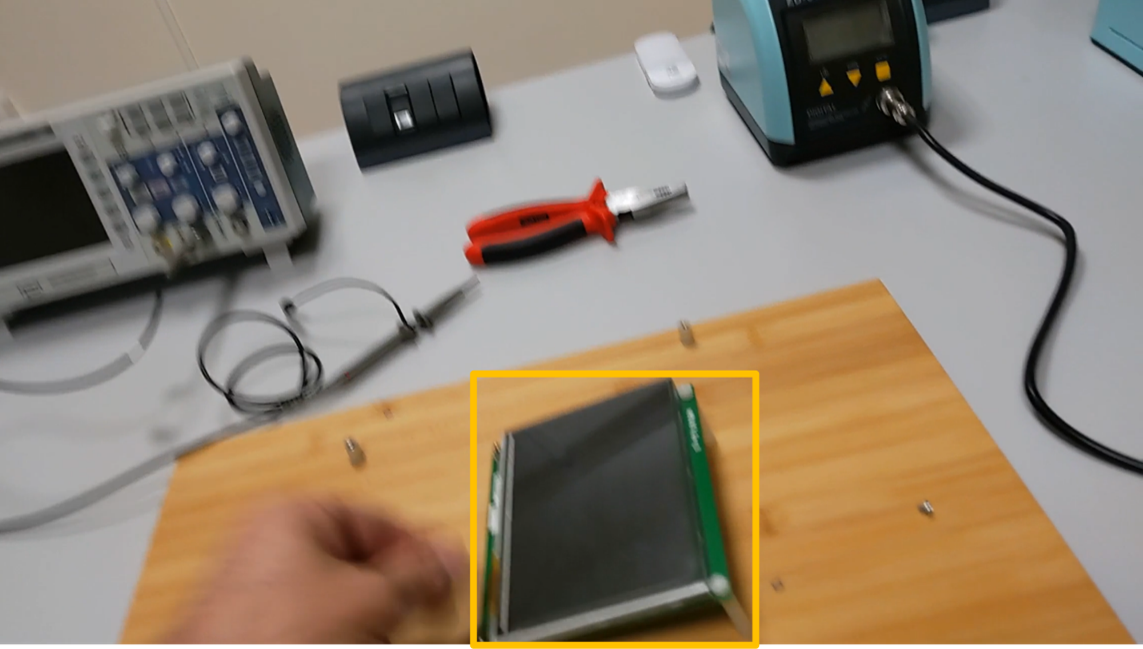}
    \caption{Example of a human-object interaction with the low voltage board's screen.}
    \label{fig:labeling_guide_fig46}
\end{figure}

\paragraph{Disassembling Board Parts}
It might happen that users disassemble parts of the board: these interactions should not be labeled. Moreover, these objects are not part of the defined taxonomy. Only a Hand\_Contact-Take with the board should be marked at the moment the hand touches it to detach/reattach the parts.
In the images, we can see an example: on the left, the Hand\_Contact-Take with the board is marked when the hand detaches the part, while on the right, no action is marked on the object placed on the table.

\begin{figure}[H]
    \centering
    \includegraphics[width=0.8\linewidth]{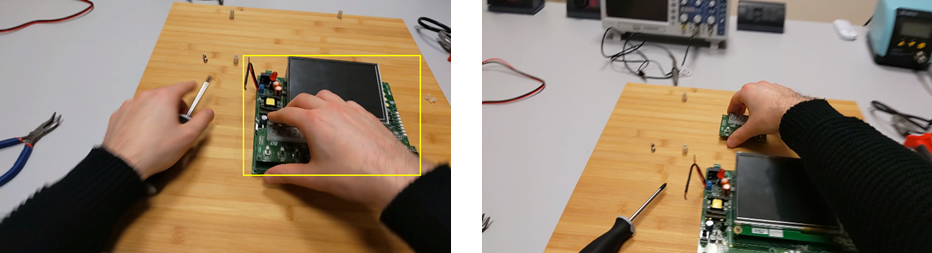}
    \caption{Example of disassembling parts of the low voltage board.}
    \label{fig:labeling_guide_fig47}
\end{figure}

\subsubsection{Special Cases regarding the Workspace}
The workspace should only be marked in interactions when the white pegs attached to it are involved in some way. The interaction must be indicated whether these pegs are touched by hands or in special cases where they align with the boards.
If an object is placed on the wooden part, or if the wood is simply touched, no interaction should be indicated.

\begin{figure}[H]
    \centering
    \includegraphics[width=0.8\linewidth]{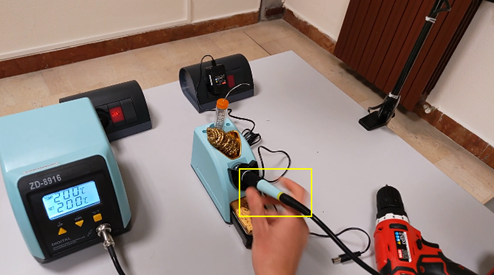}
    \caption{Example of contact between the hand and the workspace.}
    \label{fig:labeling_guide_fig54}
\end{figure}

Beyond the Hand\_Contact-Take action visible in the image above, you may find cases where the user: aligns one of the boards with the pegs before starting to screw, inserts/removes the screw connecting the board and pegs, or moves from one side of the board to the other (corner) and simultaneously touches the pegs. All these actions must be marked as a Hand\_Contact-Take on the workspace.
In this example, it is noticeable how the left hand aligns the board with the pegs, inevitably touching both objects.

\begin{figure}[H]
    \centering
    \includegraphics[width=0.8\linewidth]{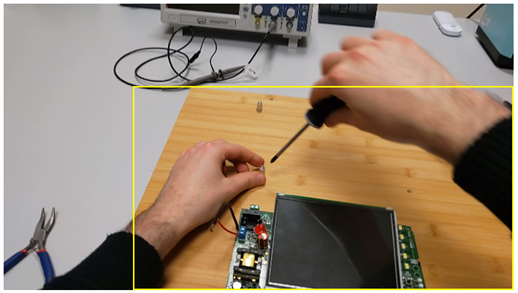}
    \caption{Contact between the hand, the workspace, and the board.}
    \label{fig:labeling_guide_fig55}
\end{figure}

\section{Computational Cost Analysis}

To provide a complete computational cost comparison and assist future research on ENIGMA-360, we report the hardware requirements and computational footprint of the evaluated baselines across all four benchmark tasks. As detailed in Table~\ref{tab:gpu_memory}, the analysis includes the specific GPU model utilized for each experiment, the training batch size, the total training time, and the computational cost during inference (measured as GPU memory in GB). As expected, Video-Grounded Cross-View VQA demands the highest memory footprint during inference due to the large parameter count of modern Vision-Language Models (which are evaluated in a zero-shot setting). Conversely, HOI Detection requires the most extensive training resources, whereas temporal segmentation and keystep recognition models remain significantly more lightweight to train and deploy.

\begin{table}[h]
\centering
\resizebox{\textwidth}{!}{
\begin{tabular}{llcccc}
\toprule
\textbf{Task} & \textbf{Baseline Model} & \textbf{GPU Model} & \textbf{Train Batch Size} & \textbf{Training Time} & \textbf{Inference Cost (GB)} \\
\midrule
\multirow{2}{*}{Temporal Action Segmentation} & FACT & 1x NVIDIA Tesla V100 & 2 & $\sim$18-24 hours & 12.5 \\
 & ASFormer & 1x NVIDIA Tesla V100 & 4 & $\sim$18-24 hours & 11.5 \\
\midrule
\multirow{2}{*}{Keystep Recognition} & GLEVR & 1x NVIDIA L40S & 16 & 0.08 hours & 0.5 \\
 & Video-Mined Task Graphs & 1x NVIDIA L40S & 16 & 0.33 hours & 0.5 \\
\midrule
\multirow{2}{*}{HOI Detection} & Baseline w/ Label Propagation & 4x NVIDIA L40S & 4 & $\sim$48 hours & 5.0 \\
 & Baseline w/ Bbox & 4x NVIDIA L40S & 4 & $\sim$12 hours & 5.0 \\
\midrule
\multirow{2}{*}{Video-Grounded Cross-View VQA} & Qwen3-VL-8B & 1x NVIDIA A30 & 1 & N/A (Zero-shot) & 22.6 \\
 & InternVL-3.5-8B & 1x NVIDIA A30 & 1 & N/A (Zero-shot) & 21.3 \\
\bottomrule
\end{tabular}
}

\caption{Detailed computational cost analysis across the baseline architectures evaluated on the ENIGMA-360 benchmark tasks. The table reports the specific GPU hardware used, the training batch size, the total training time, and the memory cost during inference (measured in GB).}
\label{tab:gpu_memory}
\end{table}

\bibliography{sn-bibliography}

\end{document}